\documentclass{article}

\usepackage{jfrExamplee}
\usepackage{apalike}

\usepackage[disable]{todonotes} % set [disable] option to remove all todo notes
\usepackage{amsmath}
\usepackage{amssymb}
\usepackage{mathtools}
\DeclarePairedDelimiter\ceil{\lceil}{\rceil}

\usepackage{graphicx}
\usepackage{tikz}
\usetikzlibrary{fit,positioning,shapes.geometric}

\usepackage{xcolor}

\usepackage[para,online,flushleft]{threeparttable}
\usepackage{setspace}
\usepackage{array}
\usepackage{booktabs}

\usepackage[]{subcaption}

\usepackage[binary-units=true,detect-weight=true,detect-family=true]{siunitx}
\DeclareSIUnit\pixel{px}
\DeclareSIUnit\ppm{ppm}
\usepackage[bottom]{footmisc} % Added to keep table above footnote
\usepackage{acro}           % Acronyms
\usepackage{url}

\usepackage{comment}
\newcommand{\xclk}{x_\mathrm{clock}}            % Command for Clock state.
\newcommand{\xclkdot}{\dot{x}_\mathrm{clock}}            % Command for Clock state differential eq.
\newcommand{\cclk}{k}            % Command for Clock constants.
\newcommand{\uclk}{u_\mathrm{clock}}            % Command for Clock control input
\newcommand{\zclk}{z_\mathrm{clock}}            % Command for Clock measurment input

\newcommand{\reffig}[1]{Fig.~\ref{#1}}
\newcommand{\reftab}[1]{Table~\ref{#1}}
\newcommand{\refsec}[1]{Section~\ref{#1}}

\newcommand{\refequ}[1]{Eq.~\ref{#1}}

\newcommand{\legendline}[1]{\textcolor[HTML]{#1}{\rule{15px}{2px}}}
\newcommand{\legenddot}[1]{\textcolor[HTML]{#1}{\rule{2px}{2px}}}

\newcommand{\reviewchanges}[1]{#1}
\DeclareAcronym{ADC}{
  short = ADC,
  long  = analog-to-digital converter,
  short-indefinite = an,
  long-indefinite = an,
}
\DeclareAcronym{AGL}{
  short = AGL,
  long  = height above ground level
}
\DeclareAcronym{CAD}{
  short = CAD,
  long  = computer-aided design
}
\DeclareAcronym{DAC}{
  short = DAC,
  long  = digital-to-analog converter
}
\DeclareAcronym{DSM}{
  short = DSM,
  long  = digital surface model
}
\DeclareAcronym{EIC}{
  short = EIC,
  long  = external interrupt controller,
  short-indefinite = an,
  long-indefinite = an,
}
\DeclareAcronym{EKF}{
  short = EKF,
  long  = extended Kalman filter,
  short-indefinite = an,
  long-indefinite = an,
}
\DeclareAcronym{EMI}{
  short = EMI,
  long  = electromagnetic induction
}
\DeclareAcronym{ENU}{
  short = ENU,
  long  = east-north-up
}
\DeclareAcronym{DFT}{
  short = DFT,
  long  = discrete Fourier transform
}
\DeclareAcronym{FLU}{
  short = FLU,
  long  = forward-left-up
}
\DeclareAcronym{FPGA}{
  short = FPGA,
  long  = field-programmable gate array,
  short-indefinite = an
}
\DeclareAcronym{FMCW}{
  short = FMCW,
  long  = frequency-modulated continuous-wave,
  short-indefinite = an
}
\DeclareAcronym{FSM}{
  short = FSM,
  long  = finite-state machine,
  short-indefinite = an
}
\DeclareAcronym{GCP}{
  short = GCP,
  long  = ground control point
}
\DeclareAcronym{GTSAM}{
  short = GTSAM,
  long  = Georgia Tech Smoothing and Mapping Library
}
\DeclareAcronym{GNSS}{
  short = GNSS,
  long  = global navigation satellite system
}
\DeclareAcronym{GPIO}{
  short = GPIO,
  long  = general-purpose input/output
}
\DeclareAcronym{GPR}{
  short = GPR,
  long  = ground-penetrating radar
}
\DeclareAcronym{GSD}{
  short = GSD,
  long  = ground sampling distance
}
\DeclareAcronym{IMU}{
  short = IMU,
  long  = inertial measurement unit,
  long-indefinite = an,
  short-indefinite = an
}
\DeclareAcronym{InSAR}{
  short = InSAR,
  long  = interferometric synthetic aperture radar
}
\DeclareAcronym{I2C}{
  short = I2C,
  long  = inter-integrated circuit,
  long-indefinite = an,
  short-indefinite = an
}
\DeclareAcronym{LQR}{
  short = LQR,
  long  = linear-quadratic regulator,
  long-indefinite = a,
  short-indefinite = an
}
\DeclareAcronym{MCU}{
  short = MCU,
  long  = microcontroller unit,
  long-indefinite = an,
  short-indefinite = an
}
\DeclareAcronym{RTK}{
  short = RTK,
  long  = real-time kinematic,
  short-indefinite = an,
  long-indefinite = a
}
\DeclareAcronym{LiDAR}{
  short = LiDAR,
  long  = light detection and ranging
}
\DeclareAcronym{MAP}{
  short = MAP,
  long  = maximum a posteriori,
  short-indefinite = an
}
\DeclareAcronym{MAV}{
  short=MAV,
  long = micro aerial vehicle,
  short-indefinite = an,
  long-indefinite = a
}
\DeclareAcronym{MEMS}{
  short = MEMS,
  long  = micro-electro-mechanical systems
}
\DeclareAcronym{NMEA}{
  short = NMEA,
  long  = National Marine Electronics Association,
  short-indefinite = an
}
\DeclareAcronym{PDF}{
  short = PDF,
  long  = probability density function
}
\DeclareAcronym{PPS}{
  short = PPS,
  long  = pulse per second
}
\DeclareAcronym{PWM}{
  short = PWM,
  long  = pulse width modulation
}
\DeclareAcronym{RMSE}{
  short = RMSE,
  long  = root mean square error,
  short-indefinite = an,
  long-indefinite = a
}
\DeclareAcronym{RMS}{
  short = RMS,
  long  = root mean square
}
\DeclareAcronym{RC}{
  short = RC,
  long  = remote control,
  short-indefinite = an,
  long-indefinite = a
}
\DeclareAcronym{ROS}{
  short = ROS,
  long  = robot operating system
}
\DeclareAcronym{RTC}{
  short = RTC,
  long  = real time counter
}
\DeclareAcronym{SBAS}{
  short = SBAS,
  long  = satellite-based augmentation system,
  short-indefinite = an,
  long-indefinite = a
}
\DeclareAcronym{SPI}{
  short = SPI,
  long  = serial peripheral interface,
  short-indefinite = an,
  long-indefinite = a
}
\DeclareAcronym{SVD}{
  short = SVD,
  long  = singular value decomposition
}
\DeclareAcronym{GPSAR}{
  short = GPSAR,
  long  = ground-penetrating synthetic aperture radar
}
\DeclareAcronym{PCB}{
  short = PCB,
  long  = printed circuit board
}
\DeclareAcronym{SHA}{
  short = SHA,
  long  = suspected hazardous area,
  short-indefinite = an,
  long-indefinite = a
}
\DeclareAcronym{TC}{
  short = TC,
  long  = timer/counter
}
\DeclareAcronym{TCC}{
  short = TCC,
  long  = timer/counter for control
}
\DeclareAcronym{UART}{
  short=UART,
  long = universal asynchronous receiver-transmitter,
}
\DeclareAcronym{UAV}{
  short = UAV,
  long  = uncrewed aerial vehicle,
  short-indefinite = a,
  long-indefinite = an
}
\DeclareAcronym{USB}{
  short = USB,
  long  = universal serial bus
}
\DeclareAcronym{UTC}{
  short = UTC,
  long  = coordinated universal time
}

\title{Under the Sand: Navigation and Localization of a \reviewchanges{Micro Aerial Vehicle} for Landmine Detection with Ground Penetrating Synthetic Aperture Radar} 

\author{
Rik B\"{a}hnemann\\
Autonomous Systems Lab\\
ETH Zurich\\
Switzerland, 8092 \\
\texttt{brik@ethz.ch} \\
\And
Nicholas Lawrance\\
Autonomous Systems Lab\\
ETH Zurich\\
Switzerland, 8092 \\
\texttt{lawrancn@ethz.ch} \\
\And
Lucas Streichenberg\\
Autonomous Systems Lab\\
ETH Zurich\\
Switzerland, 8092 \\
\texttt{stlucas@student.ethz.ch} \\
\And
Jen Jen Chung\\
Autonomous Systems Lab\\
ETH Zurich\\
Switzerland, 8092 \\
\texttt{chungj@ethz.ch} \\
\And
Michael Pantic\\
Autonomous Systems Lab\\
ETH Zurich\\
Switzerland, 8092 \\
\texttt{mpantic@ethz.ch} \\
\And
Alexander Grathwohl\\
Microwave Engineering\\
Ulm University\\
Germany, 89081 \\ 
\texttt{alexander.grathwohl@uni-ulm.de} \\
\And
Christian Waldschmidt\\
Microwave Engineering\\
Ulm University\\
Germany, 89081 \\ 
\texttt{christian.waldschmidt@uni-ulm.de} \\
\And
Roland Siegwart\\
Autonomous Systems Lab\\
ETH Zurich\\
Switzerland, 8092 \\
\texttt{rsiegwart@ethz.ch}
}

\begin{document}

\maketitle

\begin{abstract}
Ground penetrating radar mounted on \iac{MAV} is a promising tool to assist humanitarian landmine clearance.
However, the quality of synthetic aperture radar images depends on accurate and precise motion estimation of the radar antennas as well as generating informative viewpoints with the \ac{MAV}.
This paper presents a complete and automatic airborne \ac{GPSAR} system.
The system consists of a spatially calibrated and temporally synchronized industrial grade sensor suite that enables navigation above ground level, radar imaging, and optical imaging.
A custom mission planning framework allows generation and automatic execution of stripmap and circular \ac{GPSAR} trajectories controlled above ground level as well as aerial imaging survey flights.
A factor graph based state estimator fuses measurements from dual receiver \ac{RTK} \ac{GNSS} and \iac{IMU} to obtain precise, high rate platform positions and orientations.
Ground truth experiments showed sensor timing as accurate as \SI{0.8}{\micro\second} and as precise as \SI{0.1} {\micro\second} with localization rates of \SI{1}{\kilo\Hz}.
The dual position factor formulation improves online localization accuracy up to \SI{40}{\percent} and batch localization accuracy up to \SI{59}{\percent} compared to a single position factor with uncertain heading initialization.
Our field trials validated a localization accuracy and precision that enables coherent radar measurement addition and detection of radar targets buried in sand.
This validates the potential as an aerial landmine detection system.
\end{abstract}

\section{Introduction}
\begin{figure}
    \centering
    \includegraphics[width=\linewidth]{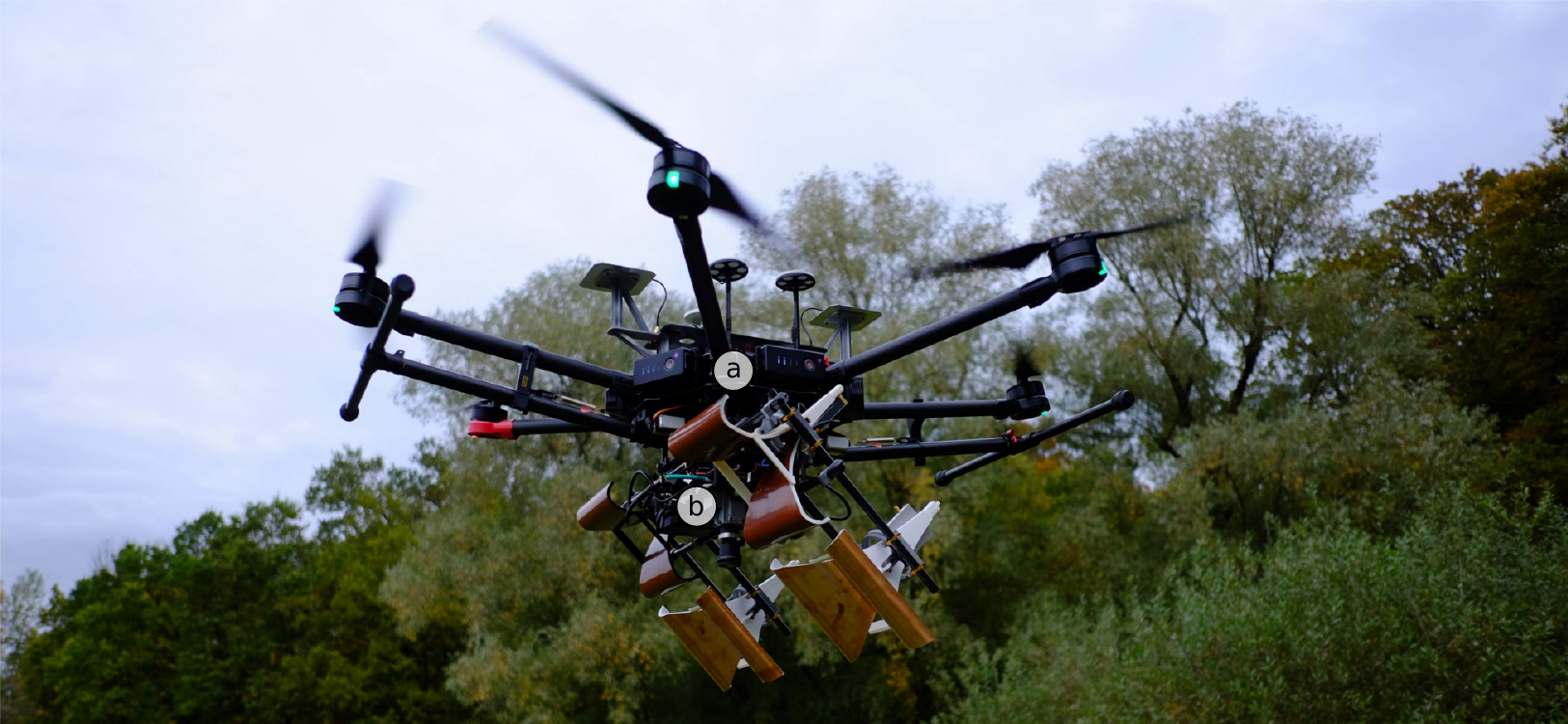}
    \caption{The DJI M600 Pro (a) equipped with a custom \ac{GPSAR} sensor payload (b) for landmine detection.}
    \label{fig:dji}
\end{figure}

Anti-personnel landmines are a massive obstacle in the pursuit of well-being in more than \si{55} affected states and regions.
Not only do they cause more than \si{4000} civilian casualties per year but also leave thousands of hectares of land uninhabitable.
The financial support for clearance has declined in the last two years, even though ongoing conflicts have continued to cause additional contamination \cite{landminemonitor2020}.
Recently, a novel detection method is being investigated to survey \acp{SHA} and eventually assist and accelerate demining.
This technology combines the mobility and ubiquity of \acp{MAV}, such as the \ac{MAV} in \reffig{fig:dji}, with the ground penetrating capability of radar imaging~\reviewchanges{\cite{garcia2020airborne}}.
In comparison to ground-based surveying and demining methods, such as metal detectors, dogs, and mine plows, \acp{MAV} can access any terrain from a safe distance and non-destructively search the area.
Our research shows that \iac{MAV} with a side-looking radar operating \SIrange{2}{4}{\metre} above ground with an along-track velocity of \SI{1}{\metre\per\second} could survey without further modification at least \SI{500}{\metre\squared\per\hour}.
This high area throughput would allow demining operations to quickly narrow down larger \acp{SHA} so that limited \reviewchanges{clearance} resources can be deployed more effectively~\reviewchanges{\cite{gichd2021imas0820}}. 

While previous publications \reviewchanges{develop the radar imaging process} on \acp{MAV}, this paper focuses on the autonomous navigation and localization of such a system.
Both capabilities are prerequisite to generating high-quality \ac{GPSAR} images.
This research describes a completely integrated system where the user is able to specify a georeferenced \ac{GPSAR} mission that the \ac{MAV} then autonomously executes.
After landing, the \ac{MAV} automatically computes precise radar antenna positions from synchronized \ac{GNSS} and \ac{IMU} measurements.
These positions are input to the radar imaging process.
The accompanying video\footnote{Watch the system summary video: \url{https://youtu.be/EfDhFKg5GEk}} summarizes the system capabilities.
In the short term, this system allows executing repeatable measurement campaigns to further develop \ac{GPSAR} imaging.
In the long term, automation will open the system to a broad range of users since complex system features are combined into simple operating elements.
Besides the overall functionality, the system includes several extensions to the state of the art in sensor interfacing, state estimation, and trajectory generation that are described in this paper:  
\begin{itemize}
    \item A low cost, lightweight, temporally synchronized and spatially calibrated sensor unit. The timing synchronization has an accuracy of \SI{0.8}{\micro\second} and precision of \SI{0.1}{\micro\second} with respect to \ac{GNSS} time and allows \ac{IMU} rates of \SI{1}{\kilo\Hz}.
    \item An extension to the \ac{GTSAM} framework to fuse \ac{IMU} with multiple \ac{GNSS} receivers in a fixed-baseline configuration. The estimator includes explicit \ac{GNSS} antenna phase center calibration. The modeling advances improve estimation accuracy by \SI{59}{\percent} over single \ac{GNSS} receiver fusion.
    \item \reviewchanges{Dynamically feasible l}inear and circular \ac{GPSAR} trajectory generation \reviewchanges{to ensure informative, uniform radar data sampling}.
    \item Providing all software components open-source listed in \reftab{tab:opensource} at the end of the document.
\end{itemize}

We organize the remainder of the paper as follows. 
\refsec{sec:related_work} sets our contributions in the context of related work.
\reviewchanges{\refsec{sec:notation} defines the notation used throughout the paper to describe the kinematic relations of the system's sensors.}
\refsec{sec:SAR} gives an overview of the \ac{GPSAR} image formation process, and the implications this sensor modality has on the system design.
\refsec{sec:system} describes the system operation overview as well as the hardware setup.
\refsec{sec:navigation} summarizes the autonomous mission generation, which includes the \ac{GPSAR} trajectory generation framework.
The novel sensor time synchronization framework is described in \refsec{sec:timing}.
In \refsec{sec:localization} we present our localization algorithm that fuses \ac{IMU} and \ac{GNSS} measurements.
\refsec{sec:results} presents the system results which demonstrate the effectiveness of the \reviewchanges{navigation and} localization pipeline and validate the detection of buried objects. 
\reviewchanges{Finally, we conclude the system development and discuss future research directions.}

\section{Related Work} \label{sec:related_work}
\reviewchanges{\Ac{GPR} is a complementary sensor to \ac{EMI} for landmine detection.
In particular its ability to discriminate low metal content makes it an attractive asset~\cite{daniels2009ground}.
Synthetic aperture radar processing further improves cross-range resolution of \ac{GPR} enabling 3D imaging of centimeter-sized objects.
\ac{GPSAR} in a side-looking or forward-looking geometry has been integrated onboard Unimogs and other military ground vehicles~\cite{wang2008image,peichl2014tirami}.}
Aerial vehicles offer considerable advantages over ground vehicles, particularly for demining where there is a need to cover large\reviewchanges{, inaccessible} areas and avoid interaction with the ground. 
The Minesweeper demonstrator project proposed the use of \ac{GPSAR} mounted on a crewed inflatable airship \cite{cristoforato2000feasibility}.
The project even conducted field tests in Kosovo but publications, in particular regarding radar results, have been sparse since then \cite{cramer2001}.
\reviewchanges{A first \ac{UAV} was later proposed by \cite{moussally2004ground}.
The authors integrate a \ac{GPSAR} payload on a Schiebel CAMCOPTER.
Their \ac{GPSAR} system provides \ac{GNSS}-inertial state estimation and stripmap and circular data collection capabilities to enable buried object detection on U.S. Army test sites.
This system was commercialized and is available today as part of \cite{miragesystems}.
However, no further developments have since been shared with the research community, and
we have no information that this system is actively used in humanitarian mine action.
We suspect that the platform they propose is still too large and expensive to be attractive for the humanitarian demining market.
As of today their system operates at \SI{150}{\metre} flight altitude, and a CAMCOPTER has more than \SI{3}{\metre} rotor diameter and weighs \SI{200}{\kilo\gram}~\cite{camcopter}. 
This is not only a logistical hurdle but also makes it unsafe to inspect cluttered unstructured terrain at low altitude.
%Additionally, their system is partially based on tactical grade equipment which may imply legal restrictions for import.
However, such a system may still be applicable in a civilian radar imaging context, where this flight limitation is not as critical~\cite{frey2021measurement}.
}

\reviewchanges{
In the humanitarian demining context, today's ambition is to integrate \ac{GPSAR} on even smaller, affordable, programmable, globally available, electrically powered \acp{MAV}.
These vehicles can operate in inaccessible, cluttered environments which increases their application range significantly over larger vehicles~\cite{fang2017robust}. 
Furthermore, they are small and in total weigh approximately \SI{10}{\kilo\gram} which simplifies transport and offers safer deployment.
Also the hardware cost of less than ten thousand USD accelerates research and makes application in humanitarian mine action more realistic.
Finally, open access research results offer the possibility of wider adoption and rapid, branched development of such systems around the world.
However, the miniaturization also raises new questions.
Smaller sensors have to be developed and integrated.
Localization and imaging processes have to be adapted to the new sensors and low-altitude flight characteristics.
And new navigation methods have to be developed to make full use of the precise, autonomous flight capabilities of \acp{MAV} for radar imaging.
Multiple research groups tackle these challenges simultaneously~\cite{fasano2017proof,colorado2017integrated,fernandez2018synthetic,schartel2018uav,schreiber2019advanced,esposito2020uav,vsipovs2020lightweight,bekar2021low,svedin2021small}.
}

\reviewchanges{\cite{fernandez2018synthetic} have been the first to present a complete \ac{GPSAR} system that can map buried objects with \iac{MAV}.
Their system has a downward-looking radar and uses \ac{RTK} \ac{GNSS} to localize in the horizontal plane as well as \ac{LiDAR} for ground distance measurement.
The second iteration of their system improves data collection, localization and radar imaging~\cite{garcia2019autonomous}.
In particular, they update their \ac{GNSS} and radar hardware, implement autonomous waypoint navigation, and introduce radar subsampling to cope with non-uniform and inaccurate in-flight data collection.
The third iteration further reduces ground surface reflection by applying \acl{SVD} filtering and deploys a dual-channel radar to reduce clutter~\cite{garcia2020airborne}.
To the best of our knowledge, the system presented in our work is the only other complete \ac{MAV}-\ac{GPSAR} system that has shown repeated buried object detection~\cite{heinzel2019comparison,schartel2020experimental,grathwohl2021experimental}.}

\reviewchanges{Our system has a different operation principle than the system presented by~\cite{garcia2020airborne}.
The \ac{GPR} is mounted in a side-looking geometry which reduces ground reflections but also has implications on the navigation.
First, the ground distance, necessary for terrain correction in the \ac{GPSAR} process, cannot be measured with a downward-facing \ac{LiDAR}.
Instead, we implement a two-step approach, where we first perform an optical survey flight to create a \ac{DSM} and then fly the georeferenced \ac{GPSAR} mission. 
Alternatively, the terrain model can be determined in the same flight with \ac{InSAR}~\cite{burr2021uav}.
Second, the side-looking view point geometry changes the data collection.
A downward-looking system has a similar field of view as nadir camera or \ac{LiDAR} setups.
It can thus utilize existing implementations of autonomous waypoint control and coverage path planning to collect radar measurements.
On the contrary, the flight path for a side-looking geometry is horizontally offset from the surface patch that is being investigated.
Thus special flight paths are necessary.
Our system implements circular and stripmap trajectories that can be combined in an arbitrary fashion to illuminate the desired surface, increase the synthetic aperture length, and provide different view points.
Despite these two key differences in the operation principle, we also approach localization and navigation differently.}

\reviewchanges{One main requirement for \ac{GPSAR} imaging is high-precision localization of the radar antennas in flight.}
In previous publications we demonstrated repeated detections of buried landmines when we fused total station theodolite position measurements with high rate \ac{IMU} measurements \cite{heinzel2019comparison,schartel2020experimental}.
In this paper, we replace the total station with \ac{RTK} \ac{GNSS} as it has better timing characteristics, operates without direct line-of-sight, delivers georeferenced positioning and is more cost-efficient.
Most importantly, it has proven to deliver similar quality radar imaging results over multiple measurement campaigns in open fields \cite{schartel2021signalverarbeitungskonzepte}.
The general understanding is that \ac{GNSS} is prone to errors due to multi-path effects, atmospheric interference and satellite occlusion in complex environments such as cities, forests, and mountains.
With the more recent introduction of alternative \acp{GNSS} such as GLONASS, BeiDou and Galileo, accuracy and coverage have increased notably \cite{li2015precise}.
Furthermore, various developments on antennas, receivers and navigation processors have improved multi-path rejection \cite{strode2016gnss}.
Commercial solutions are available that implement tight \ac{GNSS} \ac{IMU} fusion such as the APX-15 UAV by Applanix \cite{mian2015direct}.
While these systems are well integrated and would probably deliver comparable positioning quality to our proposed system, the proprietary structure does not offer the same flexibility and control as is available on our own hardware system.
In our system we have full control over the spatial and temporal calibration, the raw measurements and estimation algorithms.
This leaves room for future integration of additional sensor modalities such as \ac{LiDAR} or vision, and allows for direct extensions to our work from the wider robotics community.

Our previous estimation algorithm was based on a Kalman filter, followed by a full factor graph smoothing step \cite{schartel2020experimental}. 
Now, we combine the state estimation into a single \ac{GTSAM} framework, where an incremental smoothing algorithm runs online during the flight and full inference is computed after landing. 
It has been shown that incremental smoothing, in particular for systems with highly non-linear dynamics, such as rotary wing vehicles, performs significantly better than filtering \cite{indelman2013information}.
\reviewchanges{Furthermore, the factor graph formulation provides a favorable formalism to include additional localization sensors in the future.}
% Fusion GNSS IMU
The motion estimator proposed in this paper is loosely coupled, i.e., the \ac{GNSS} raw double difference measurements are prefiltered into a single position measurement using commercial software before being fused with \ac{IMU} linear acceleration and angular velocity measurements.
This design decision simplifies the development process significantly, because the software handles difficulties such as multi-path, shadowing, carrier phase ambiguity and switching between the \ac{RTK} and \ac{SBAS} solution.
Tightly coupled systems on the other hand would deliver positional constraints even in underconstrained cases, for example when observing only two satellites \cite{schneider2016fast,cao2021gvins}.
Furthermore, tight integration would allow more accurate uncertainty propagation of the raw distance measurements and modeling of the frequency-dependent antenna phase centers.
To partially account for this error, loosely coupled frameworks often model the offset between the \ac{IMU} and \ac{GNSS} antenna \cite{kaplan2005understanding}. 
We integrate this offset into the \ac{GTSAM} optimization framework which then permits the system to spatially calibrate the \ac{GNSS} antenna position relative to the \ac{IMU}.
Additionally, we extend \ac{GTSAM} to include \ac{GNSS} moving baseline measurements, i.e., the \ac{RTK} baseline vector between two \ac{GNSS} antennas mounted on the vehicle.
Finally, we analyse the influence of the two measurement factors.

One remaining issue when fusing multiple sensor modalities is time synchronization, as drifting or noisy time stamps usually break the modeling assumptions.
\reviewchanges{\cite{ding2008time} perform an error analysis of time delay on \ac{GNSS}-inertial navigation systems.
They come to the conclusion that time synchronization is particularly important in dynamically fast, self-calibrating systems with high-precision sensors, i.e., exactly the robotic system described in this work.}
Timing errors may occur due to varying clock speeds, protocol transfer delays, sensor exposure or computing capacities. 
Robotic systems often time stamp sensors on different clocks and then translate the stamp from one clock to the other by estimating the clock offset and skew.
The time offset estimation can either be based on comparing the same time stamps on two clocks \cite{eidson2002ieee,sommer2017low,osadcuks2020clock} or based on the correlation of sensor motions, such as camera and \ac{IMU} movement \reviewchanges{\cite{furgale2012continuous,li2014online,kelly2014general,qin2018online}}.
More specifically, \cite{skog2011time} propose \reviewchanges{\iac{GNSS}-inertial} fusion algorithm that self-calibrates the timing offset.
Time translation in software only, however, does not solve the whole sensor data acquisition problem.
\cite{kelly2021question} showed recently, that co-estimating time offset in causal filters introduces inconsistencies and potentially worsens the estimation result.
Furthermore, even though the sensor data may be time stamped accurately, the sensor triggering could still happen out of sync and indeterminately.
This issue then needs to be handled by interpolating sensor data as well as implementing \reviewchanges{stochastic cloning and} complex buffering schemes to insert delayed sensor data into the optimization framework~\reviewchanges{\cite{roumeliotis2002stochastic,lynen2013robust,cioffi2020tightly}}.
Instead of handling sensor timing compensation after acquisition, a dedicated hardware solution can trigger and time stamp sensor data from a single clock.
\reviewchanges{Different designs based on \acp{FPGA} or \acp{MCU} offer more or less freedom to be extended to a custom sensor setup and \ac{GNSS} synchronization~
\cite{kais2006multi,ding2008time,huck2011precise,nikolic2014synchronized,albrektsen2018user,tschopp2020versavis,faizullin2021open,faizullin2021synchronized}.}
\reviewchanges{We base our work on the open-source \textit{VersaVIS}~\cite{tschopp2020versavis} sensor synchronization board}.
Here a SAMD21 \ac{MCU} triggers an \ac{IMU}-camera setup and stores synchronized time stamps.
\reviewchanges{This} hardware setup offers a wide variety of serial interfaces such as \ac{SPI}, \ac{UART} and \ac{I2C}.
We extend \reviewchanges{the hardware} by adding a high resolution oscillator which we synchronize with respect to \ac{GNSS}.
Furthermore, our system implements hardware-generated \acp{PWM} that allow triggering multiple sensor sources in parallel at desired \ac{GNSS} time stamps.
And we utilize the SAMD21's hardware capture capability to improve timing precision of sensor strobe signals.

% Motion generation
\reviewchanges{Finally, \ac{GPSAR} imaging is heavily dependent on the data collection process.
While in principle \ac{GPSAR} can integrate measurements from anywhere along the trajectory, in practice over- and undersampling, e.g., due to alternating velocities, causes artefacts in the radar image.
This effect has been studied by~\cite{garcia20203d} who propose a post processing step to subsample informative radar measurements and normalize the image.
A complementary way to improve sampling is to enforce constant velocity along the \ac{GPSAR} measurement trajectory.
Available trajectory planners such as QGroundControl primarily focus on waypoint missions with a rudimentary option to constrain the flight speed~\cite{qgroundcontrol}.
These planners do not consider platform acceleration limitations.
This results in non-uniform sampling when changing directions because the position tracking controller does not generate a smooth transition.
Also geometric primitives, such as circular segments, are not necessarily implemented.
Our approach to mission planning is based on polynomial trajectory generation.
Rotary wing \acp{MAV} with a perpendicular thrust vector are differentially flat.
Their pose is fully defined by the position, heading and their derivatives.
Consequently, polynomial trajectories, defined in position, yaw, and time, are a suitable representation to generate fully-defined, smooth trajectories~\cite{richter2016polynomial}.
Because the \ac{MAV} state is defined at every step in time, dynamic feasibility can be enforced~\cite{mueller2015computationally}.
Our approach uses these principles to generate uniform sampling measurement trajectories.
The trajectories have a predefined start and end time to mask the radar measurements.
They enforce constant velocity along circular or linear trajectories.
And they are dynamically feasible with respect to maximum thrust and rotation rates to enable precise tracking.
In combination with robust altitude tracking and waypoint control this leads to a navigation system that allows precise measurement collection, even beyond visual line of sight.}

% ----------------------------------------------------------------------------- %
% ---------------------------------- SECTION ---------------------------------- %
% ----------------------------------------------------------------------------- %
\reviewchanges{
\section{Conventions and Notations} \label{sec:notation}
In this paper, coordinate systems are denoted by calligraphic capital letters, including $\mathcal{I}$ to represent the world-fixed, inertial frame and $\mathcal{B}$ to represent the body-fixed frame rigidly attached to the vehicle.
A 3D vector is represented by a lower case letter with a leading subscript describing the coordinate system the vector is represented in.
Vectors have one or two right-hand subscripts for further description.
For instance, ${}_{\mathcal{I}}r_{\mathcal{IB}} \in \mathbb{R}^3$ is the vector from the origin of coordinate system $\mathcal{I}$ to the origin of coordinate system $\mathcal{B}$, represented in the coordinate system $\mathcal{I}$.
Similarly, ${{}_{\mathcal{B}}\omega_{\mathcal{I}\mathcal{B}} \in \mathbb{R}^3}$ describes the angular velocity of $\mathcal{B}$ with respect to $\mathcal{I}$, represented in coordinate system $\mathcal{B}$.
Single right-hand subscripts refer to vectors that are fully described by their direction and magnitude.
${}_{\mathcal{I}}v_{\mathcal{B}}\in \mathbb{R}^3$ describes the translational velocity of the origin of coordinate system $B$, represented in coordinate system ${\mathcal{I}}$.
${}_{\mathcal{I}}r_{g}\in \mathbb{R}^3$ describes the direction of the gravitational field, represented in coordinate system ${\mathcal{I}}$.
3D rotations and rigid transformations are represented by the capital letters $R$ and $T$ respectively.
${R_{\mathcal{I}\mathcal{B}}} \in SO(3)$ indicates the orientation, ${T_{\mathcal{I}\mathcal{B}}} \in SE(3)$ the orientation and translation of frame $\mathcal{B}$ with respect to reference frame $\mathcal{I}$.
We use passive rotations and transformations to map 3D vectors from one frame to another.
For example, ${}_{\mathcal{I}}v_{\mathcal{B}} = {R_{\mathcal{I}\mathcal{B}}}~ {}_{\mathcal{B}}v_{\mathcal{B}}$ maps the velocity represented in frame $\mathcal{B}$ to frame $\mathcal{I}$.
}

\section{Synthetic Aperture Radar Imaging} \label{sec:SAR}
\begin{figure*}
\begin{subfigure}{.54\textwidth}
\centering
\includegraphics[height=2.5in]{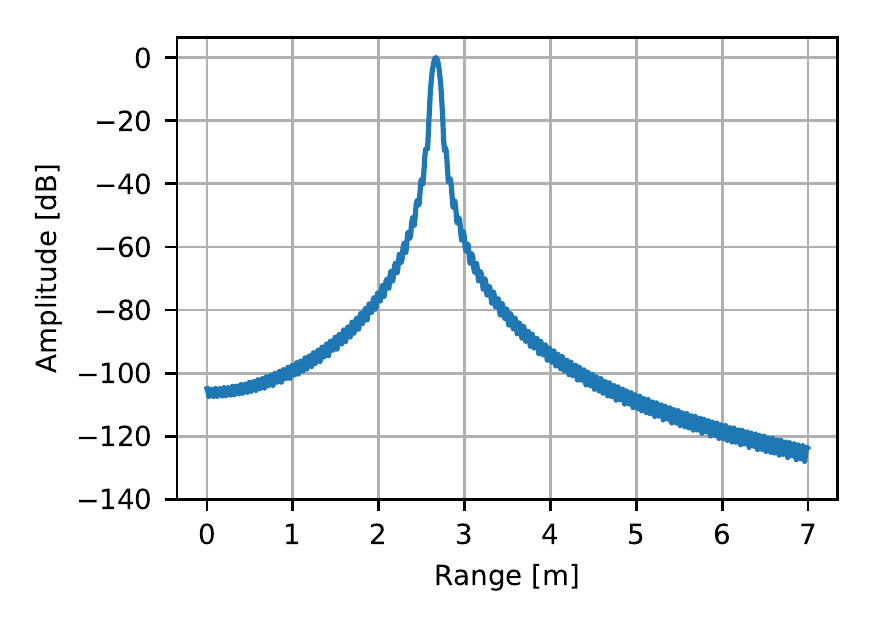}
\caption{Radar response of a single point target at \SI{2.67}{\metre}.}
\label{fig:fft}
\end{subfigure} 
\begin{subfigure}{.45\textwidth}
\centering
\includegraphics[height=2.5in]{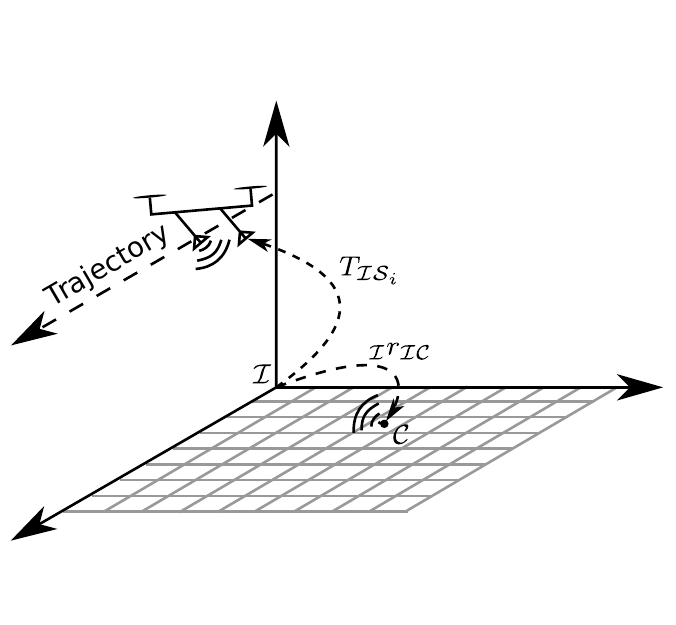}
\caption{Back projection of radar response into grid cell.}
\label{fig:backprojection}
\end{subfigure}
\centering
\caption{A back projection algorithm forms the radar image. Multiple radar range measurements are accumulated into a 3D voxel grid to resolve the bearing ambiguity and mitigate noise. This process requires a well defined measurement trajectory, accurate radar antenna localization and an accurate \ac{DSM}.}
\label{fig:sar}
\end{figure*}

Synthetic aperture radar imaging is the process of forming a high resolution 3D image from a collection of radar range measurements.
\reffig{fig:fft} exemplifies the idealized range response of a single target.
Typically, a single radar chirp contains not only one target but the sum of noisy responses of multiple targets with different scattering characteristics.
In order to filter out the 3D location of prominent targets, the platform averages multiple measurements from different viewpoints gathered along a measurement trajectory (see \reffig{fig:backprojection}).
Observing a radar target from different view points resolves the bearing ambiguity and reveals the location of strong scatterers, which can include both plastic and metal landmines. 
The particular back projection algorithm used in this work enforces specific design decisions on the motion of the platform, the radar antenna localization, and the generation of a \ac{DSM} \cite{zaugg2015generalized,schartel2021signalverarbeitungskonzepte}.
In this section we elaborate the radar imaging process and identify the implications on the system design.

The system in this work has a side-looking single-input single-output radar.
With each measurement, the radar emits \iac{FMCW} that reflects at the interface between materials with different permittivity.
A frequency mixer creates the difference signal between the still transmitting \ac{FMCW} and its returning time-delayed copies.
For a single scatterer and assuming no antenna movement for the duration of the chirp, the difference signal has a constant frequency depending on the distance to the object and the propagation speed of the electromagnetic wave.
\Iacl{ADC} discretizes the difference signal and \iacl{DFT} isolates the frequency components.
The resulting frequency spectrum correlates with the distance to all scatterers in the field of view.

In the next step every single measurement is back projected into every grid cell of a 3D voxel map to evaluate the complex pixel value $A({}_{\mathcal{I}}r_{\mathcal{I}\mathcal{C}}) \in \mathbb{C}$ of each cell located at ${}_{\mathcal{I}}r_{\mathcal{I}\mathcal{C}} \in \mathbb{R}^3$, where $\mathcal{I}$ denotes the inertial coordinate frame and $\mathcal{C}$ is the image cell center.
Given the position of the radar antennas with respect to the cell location and the electromagnetic wave propagation speed in air and soil as well as possible refraction at the ground surface, the amplitude and phase of the measured, range-compressed radar signal $S_\eta$ is determined, see \reffig{fig:fft}.
The resulting signal at the image cell location is multiplied with the negative expected response if a perfect point target was located inside the cell and summed up over all measurements $\eta$.
\begin{align}
    A({}_{\mathcal{I}}r_{\mathcal{I}\mathcal{C}}) &= \sum_{\eta = 1}^N w_\eta ~ S_\eta ~ e^{-j \phi_\eta}, \label{eq:sar}
\end{align}
where $N$ is the total number of measurements in one trajectory, $w_\eta$ is a normalization weight to compensate for non-uniform sampling, and $\phi_\eta$ is the expected phase of the point target.
This phase depends on the radar antenna position, the location of the imaging cell, and the \ac{DSM} defining the interface between air and soil and thus the expected wave propagation time. 

The algorithm has the following assumptions to perform range estimation and obtain focused images:
\begin{itemize}
    \item An exact pose estimation $T_{\mathcal{I}\mathcal{S}_i}$ of the transmitting and receiving radar antennas
    \item An exact \ac{DSM}
    \item An  exact model of soil permittivity and ground refraction
\end{itemize}
These assumptions often lead to the conclusion that the range of the antenna to the target needs to be known within $\frac{1}{16}$ or $\frac{1}{8}$ of the wavelength, which in our case would be less than \SI{10}{\milli\metre} \cite{schartel2021signalverarbeitungskonzepte,fernandez2018synthetic}.
Otherwise, the coherent addition in \refequ{eq:sar} will not correlate due to the radar signals being out of phase for the same target observed from different viewpoints. 
In practice, however, the algorithm can image targets with less accurate range estimation.
In the case of less accurate antenna position estimates, biases in the range estimate will split the energy of a single target across multiple image cells and cause defocus or \textit{smearing}.
The goal of this work is to minimize the errors that occur due to radar antenna positioning. In the next section we provide an overview of our complete airborne \ac{GPSAR} system.
In \refsec{sec:navigation} we introduce vehicle navigation principles that enforce uniform sampling of target cells from different view points as well as accurate trajectory tracking control.
In \refsec{sec:timing} and \refsec{sec:localization} we address time synchronization and vehicle localization to improve the position estimation of the radar antennas.

% ----------------------------------------------------------------------------- %
% ---------------------------------- SECTION ---------------------------------- %
% ----------------------------------------------------------------------------- %

\section{System Overview} \label{sec:system}
Our system can generate georeferenced aerial imaging products as well as execute low-altitude \ac{GPSAR} missions.
This requires a specific hardware setup as well as mission workflow to efficiently combine both functionalities into a single system.
\subsection{\ac{MAV} Platform Setup}
\begin{figure*}
\begin{subfigure}{.49\textwidth}
\centering
\includegraphics[height=2.48in]{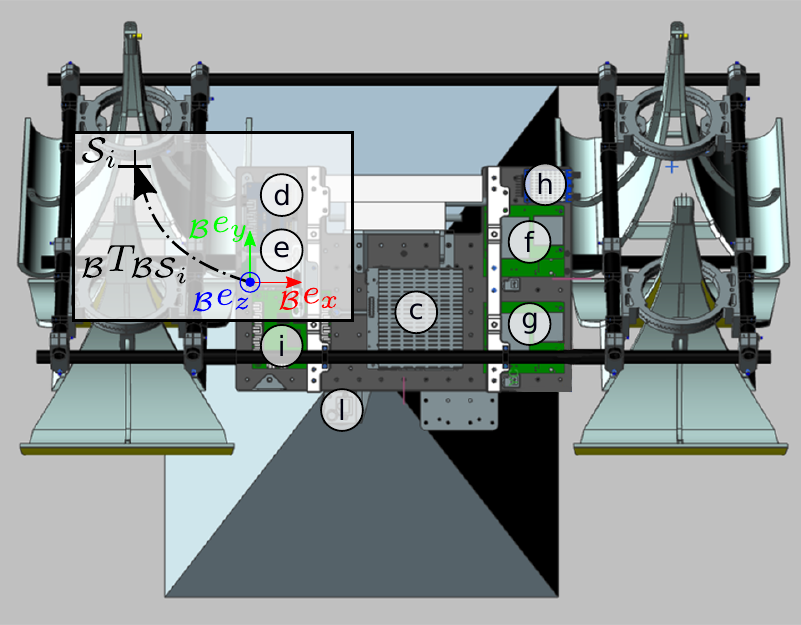}
\caption{Top view of sensor payload.}
\label{fig:pod_top}
\end{subfigure}
\begin{subfigure}{.49\textwidth}
\centering
\includegraphics[height=2.48in]{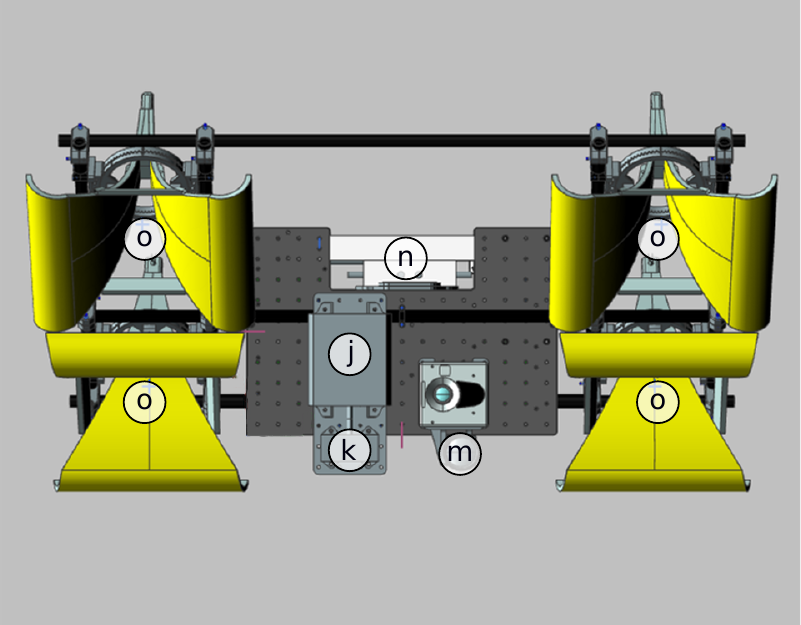}
\caption{Bottom view of sensor payload.}
\label{fig:pod_bottom}
\end{subfigure}
\centering
\caption{The sensor pod holds all sensors necessary for optical and radar imaging as well as platform localization and control. \reviewchanges{The carbon tube construction fixes the radar antennas in place such that the \ac{CAD} model fully defines the position and orientation of the antenna phase centers with respect to the \ac{IMU}.}}
\label{fig:sensor_pod}
\end{figure*}

% Hardware setup
Our \ac{MAV} setup consists of a commercially available DJI M600 Pro platform equipped with a custom sensor pod payload.
The sensor pod holds all components that are necessary to implement the airborne \ac{GPSAR}.
We divide the setup into an upper compartment, that includes all processing, communication and interoceptive sensing (\reffig{fig:pod_top}) and a lower compartment holding all exteroceptive sensors (\reffig{fig:pod_bottom}).
At the center of the upper compartment is an UP Squared single board computer (c), sufficiently powerful to log the sensor data, compute the online navigation solution, and control the \ac{MAV} through its autopilot and \ac{ROS}.
Next to it is the \textit{VersaVIS} (d), \iac{MCU} with peripherials to interface the sensors.
The \ac{IMU} (e) and two \ac{GNSS} receivers (f) and (g), together with two helical triple band antennas, and the corrections modem (h) form the basis of the high precision navigation solution necessary for exact platform localization.
All components are powered by a single power board (i) capable of converting the \ac{MAV}'s \SI{18}{\volt} power supply from the onboard batteries to \SI{5}{\volt}, \SI{12}{\volt}, and \SI{24}{\volt}.
On the bottom of the sensor pod the radar altimeter (j) and \ac{LiDAR} altimeter (k) measure the platform \ac{AGL}.
The camera (l) and lens (m) create aerial imagery to build a photogrammetric map for path planning and radar imaging.
Next to them, the custom \SIrange{1}{4}{\giga\Hz} \ac{FMCW} radar (n), and its four horn antennas (o) with different polarization direction are used to create \ac{GPSAR} images \cite{burr2018design}.
The four antenna phase center locations with respect to the \ac{IMU} ${}_{\mathcal{B}}T_{\mathcal{B}{\mathcal{S}_i}}$ for $i\in\{1\ldots4\}$ are well defined through the \ac{CAD}. 
The camera pose with respect to the \ac{IMU} ${}_{\mathcal{B}}T_{\mathcal{B}{\mathcal{S}_5}}$ is determined with a calibration software~\cite{furgale2013unified}.

\begin{table}
\centering
\begin{tabular}{@{}llll@{}} 
\toprule 
ID  & Item & Description & Rate \\ 
\midrule a & DJI M600 Pro   & Hexacopter with \ac{ROS} onboad SDK & - \\
         b & Sensor payload & Sensor pod (prepreg carbon, milled alloy, nylon 3D print) & -\\
         c & UPS-APLP4-A20-0864 & Onboard processing unit & -\\
         d & VersaVIS  & Sensor interface based on SAMD21 \ac{MCU} & - \\
         e & ADIS16448BMLZ & Industrial grade MEMS \ac{IMU} & \SI{1}{\kilo\Hz} \\ 
         f & Piksi Multi reference & \ac{RTK} \ac{GNSS} position receiver  & \SI{10}{\Hz}\\ 
         g & Piksi Multi attitude & \ac{RTK} \ac{GNSS} moving baseline receiver & \SI{5}{\Hz} \\ 
         h & RFD868x & Long range \ac{RTK} \ac{GNSS} correction modem & -\\ 
         i & Power distribution board & \SIrange{16}{27}{\volt} DC input, \SI{5}{\volt}, \SI{12}{\volt}, and \SI{24}{\volt} DC output & - \\  
         j & US-D1 & \SI{24}{\giga\Hz} radar altimeter & \SI{100}{\Hz}  \\ 
         k & LIDAR-Lite v3HP & \ac{LiDAR} altimeter & \SI{10}{\Hz} \\ 
         l & BFLY-PGE-31S4C-C & Global shutter \SI{3.2}{\mega\pixel} color camera & \SI{2}{\Hz}\\ 
         m & Edmund Optics \#35-139 & \SI{6}{\milli\metre} fixed focal length lens & - \\ 
         n & Custom radar & \SIrange{1}{4}{\giga\Hz} \ac{FMCW} ground penetrating radar & \SI{200}{\Hz}\\
         o & Horn antennas & Transversal electromagnetic horn antenna for \ac{GPSAR} & -\\  
         - & 33-HC882-28    & Helical triple band \ac{GNSS} antennas & - \\
\bottomrule
\end{tabular}
\caption{Component list of the landmine detecting \ac{MAV}. The ID refers to the labels in \reffig{fig:dji} and \reffig{fig:sensor_pod}.}
\label{tab:sensors}
\end{table}

\subsection{Mission Overview}
\label{sec:mission_overview}
\begin{figure*}
    \begin{subfigure}{.49\textwidth}
    \centering
    \includegraphics[height=2.48in]{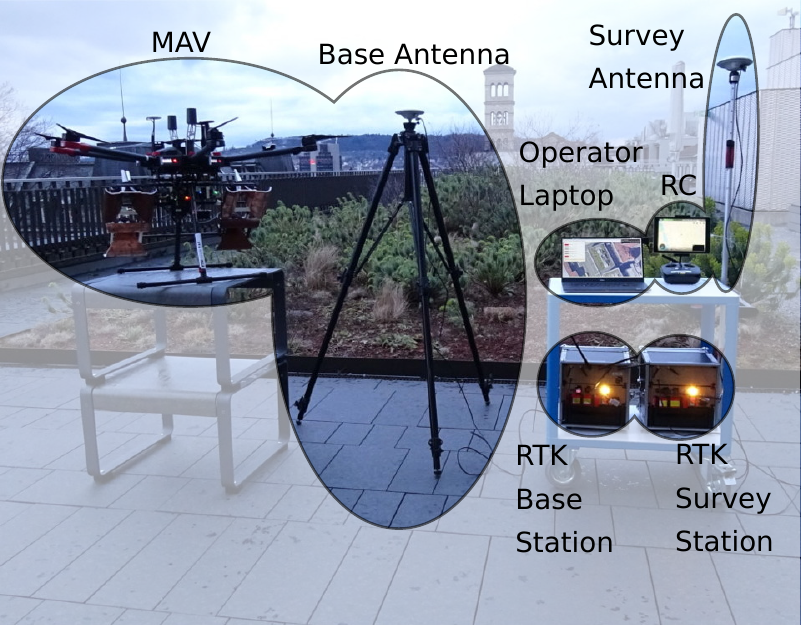}
    \caption{The field test setup.\newline\newline}
    \label{fig:setup}
    \end{subfigure}
    \begin{subfigure}{.49\textwidth}
    \centering
    \includegraphics[height=2.48in]{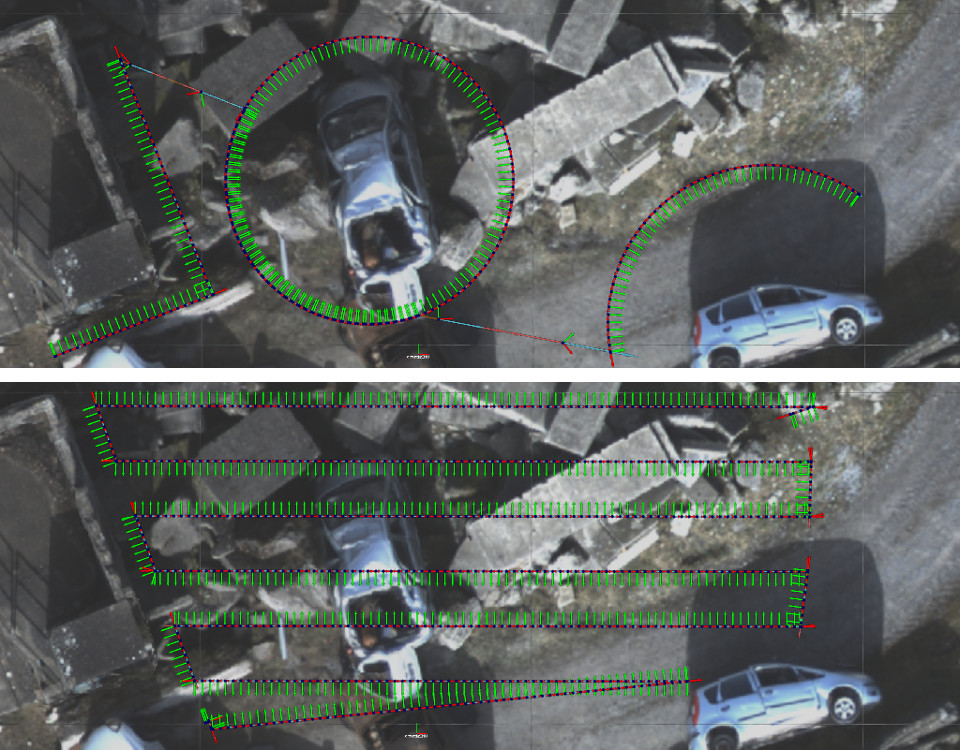}
    \caption{\reviewchanges{(top) User-defined combination of constant velocity linear and circular trajectories or (bottom) automatic} coverage mission planning.}
    \label{fig:mission_planning}
    \end{subfigure}\\
    \begin{subfigure}{.99\textwidth}
    \centering
    \vspace{0.3cm}
    \includegraphics[width=.99\linewidth]{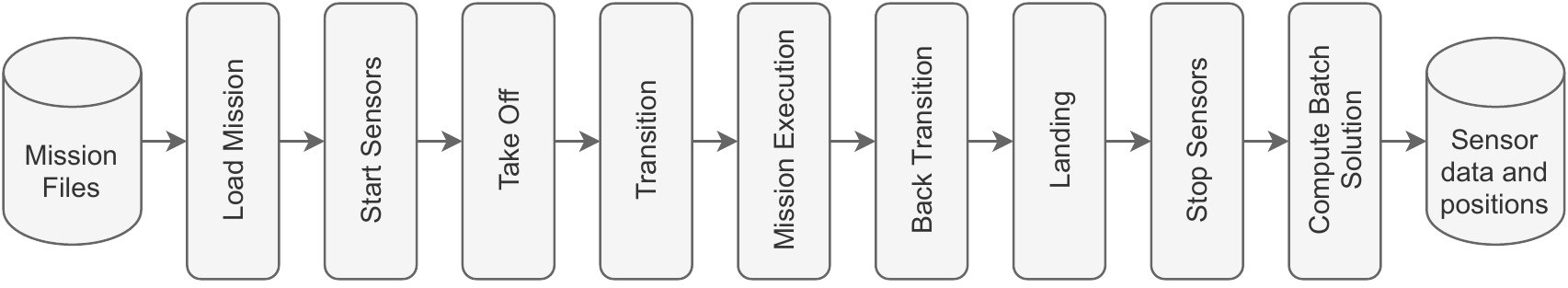}
    \caption{The \acl{FSM} controlling the autonomous mission execution.}
    \label{fig:fsm}
    \end{subfigure}
    \caption{An overview of the system setup, mission planning and execution. The operator plans the missions offline and the \ac{MAV} executes the flight and sensor logging autonomously.}
\end{figure*}

% Mission overview
The complete system setup, shown in \reffig{fig:setup}, includes the copter with payload, \iac{RTK} base station that supplies WiFi, 4G communication and \ac{RTK} corrections, a survey station to measure geodetic \acp{GCP} and an operator laptop running \ac{ROS} to monitor the platform status and select, configure and start the autonomous mission.
For safety, the operator can always interrupt a mission with \iac{RC}.
The mission itself can be planned in the field or in the office on the operator laptop with respect to a georeferenced map.
\reffig{fig:mission_planning} shows two example missions consisting of concatenated motion primitives, e.g., circles or lines or an automatically generated boustrophedon coverage pattern \cite{bahnemann2021revisiting}.
After surveying the \ac{RTK} \ac{GNSS} base station position, the mission is executed autonomously with all processing and data logging happening on board the platform controlled by the \acl{FSM} depicted in \reffig{fig:fsm} \cite{pradalier2017task}.
The \acl{FSM} starts sensor recording and takes off the \ac{MAV} automatically.
The platform ascends to a collision-free altitude, e.g, \SI{30}{\metre}, and transitions to a location above the start of the mission.
It then descends to the mission altitude and executes the preplanned trajectory. 
After finishing, the platform transitions back to high altitude, returns to the take off position and automatically lands.
After landing the system post processes the sensor position batch solution.
The sensor and positioning data are downloaded from the onboard computer and passed to the optical or radar imaging processor.

\section{Navigation}
\label{sec:navigation}
According to the back projection \refequ{eq:sar}, the radar response of a single radar image cell is the weighted sum of all radar measurements illuminating that cell.
In order to observe all targets in a scene and simplify the weighting, it is important that the \ac{MAV} illuminates all cells uniformly and from different view points.
For this purpose we generate smooth, constant velocity radar measurement trajectories as well as a trajectory tracking controller that allows terrain following control.

\subsection{Polynomial Trajectory Generation}
\label{sec:trajectory_generation}
\begin{figure*}
\begin{subfigure}{.62\textwidth}
\centering
\includegraphics{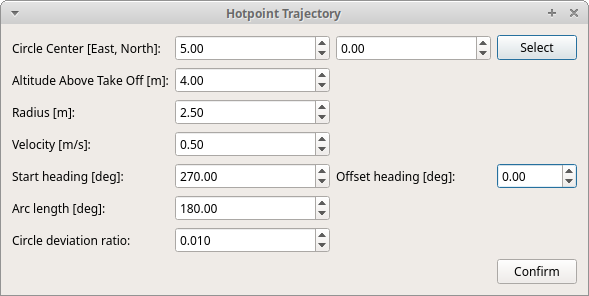}
\caption{\reviewchanges{User interface.}}
\end{subfigure}
\begin{subfigure}{.37\textwidth}
\centering
\includegraphics{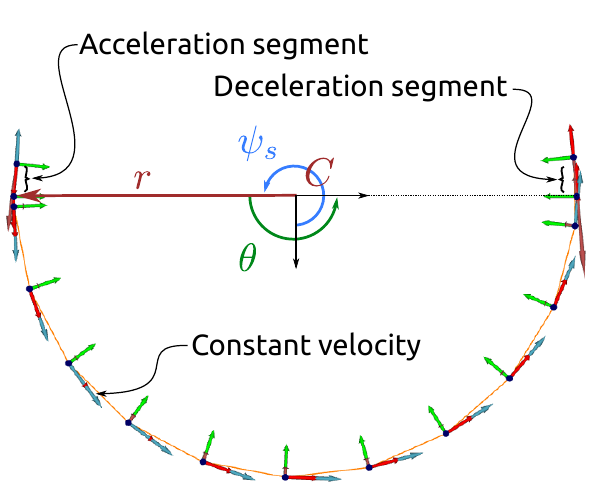}
\caption{\reviewchanges{Dynamically feasible circular trajectory.}}
\label{fig:circle_acceleration}
\end{subfigure}
\centering
\caption{The user plans continuous, smooth and feasible polynomial measurement trajectories (orange) via high-level inputs. A trajectory fully defines the \ac{MAV} position and orientation (coordinate frames), velocity (blue arrows) and acceleration (red arrows) at every time step during the mission.}
\label{fig:trajectorygeneration}
\end{figure*}

Typical radar imaging modes are \textit{stripmap} and \textit{circular} \ac{GPSAR}.
\textit{Stripmap} refers to flying over the target area along a straight line, while in \textit{circular} \ac{GPSAR} the platform orbits the target area with the radar pointing towards the center.
Our trajectory generation framework allows combining these two imaging modes in an arbitrary fashion as shown in \reffig{fig:mission_planning}.
The trajectories are formed based on high-level user input and are automatically transformed into continuous, smooth, and feasible polynomial trajectories based on a differentially flat multirotor model~\cite{richter2016polynomial,burri2015real}.
For example, in \reffig{fig:trajectorygeneration}, a user defines a circular arc by its center $C$, radius $r$, start angle $\psi_s$ and subtended angle $\theta$, as well as altitude, platform velocity and heading while flying along the circle.
In an intermediate step, this user input is converted into waypoint constraints that serve as polynomial support vertices.
In the case of the circle these are $M+1$ vertices equally spaced along the perimeter.
$M$ is computed automatically as 
\begin{align}
    M &= \ceil*{\frac{\theta}{\arccos{\left(2 \left( 1 - d \right)^2 - 1 \right)}}},
\end{align}
where $d \in \mathopen(0, 1\mathclose)$ is a parameter that describes the relative deviation of the vertices approximating the circle from an actual circle.
In the case of a straight line the support vertices are simply the start and goal point of the line.
Once the support vertices' positions have been defined, their velocity, i.e., the first derivative of the polynomial, is set to the user input velocity. In the \textit{circular} case, this is angled tangentially to the trajectory, while in the \textit{stripmap} the velocity vector points towards the next waypoint.
All other derivatives, depending on the maximum polynomial degree, are set to zero, as the platform should move with constant velocity along the measurement trajectory.
Similarly, the heading polynomial is fully constrained by pointing into the direction of the velocity vector.
Furthermore, the segment transition times between two vertices are fully defined by the path length and constant velocity.
The constrained vertices and segment times completely describe the polynomial coefficients and thus the full desired state of the \ac{MAV} at every time.

Every measurement trajectory is started from rest.
In order to bring the platform up to speed at the first vertex and slow it down after the final vertex we introduce acceleration and deceleration segments.
At both the start and stop positions, an additional vertex is connected to the measurement trajectory.
The time allocation of this segment is unknown and subject to physical feasibility constraints of the platform.
A binary search minimizes the segment time such that the acceleration and deceleration motion is as short as possible but still feasible with respect to minimum and maximum thrust, velocity, and rotation rates \cite{mueller2015computationally}.
Similarly, consecutive measurement trajectories are connected by rest-to-rest trajectories using the minimum segment time search.

\subsection{Tracking Controller}
\label{sec:controller}
\begin{figure}
    \centering
    \includegraphics[height=1.99in]{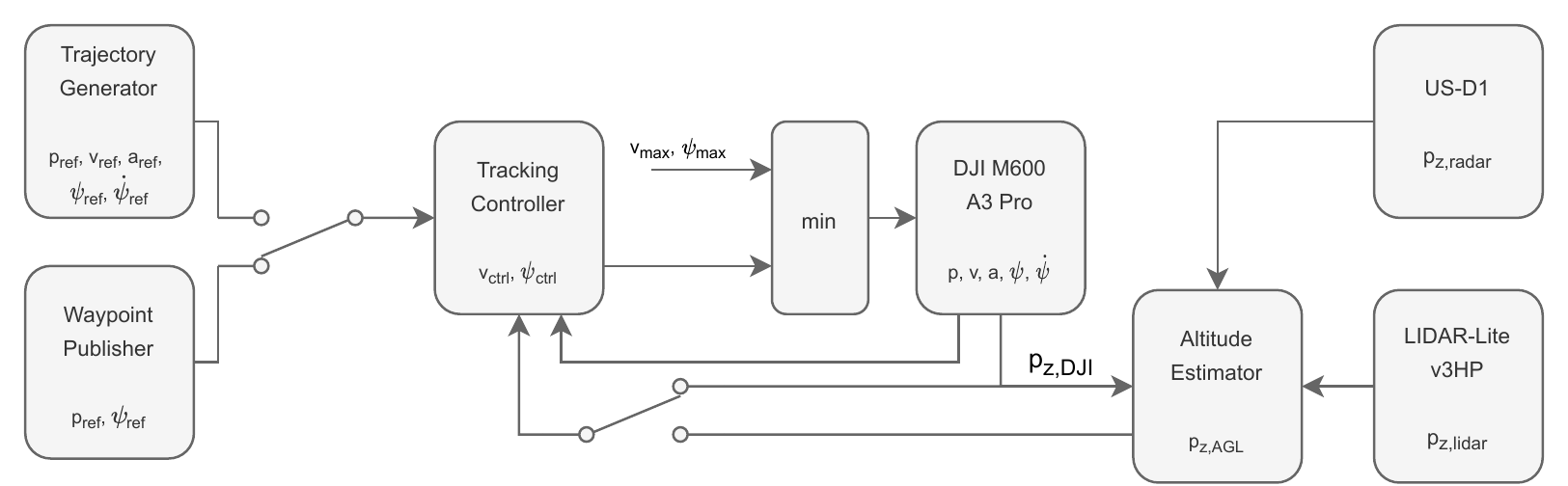}
    \caption{The platform control loop. The \acl{FSM} switches between trajectory tracking or waypoint tracking and between control at \ac{GNSS} altitude or \ac{AGL}.}
    \label{fig:controls}
\end{figure}

\reffig{fig:controls} displays the control loop that runs on board the \ac{MAV}.
The \acl{FSM} has two options to control the flight of the \ac{MAV}. 
If the goal is to go as fast as possible from one waypoint to another, e.g., in the transition maneuver, it can directly send the desired position $p_{\mathrm{ref}}$ and heading  $\psi_{\mathrm{ref}}$ to the tracking controller.
If smooth, well-defined trajectories are desired, it can send polynomial missions as described above to the controller.
The controller continuously evaluates the desired position, velocity $v_{\mathrm{ref}}$, acceleration $a_{\mathrm{ref}}$, heading, and yaw rate $\dot{\psi}_{\mathrm{ref}}$, and compares it to the current position, velocity and acceleration estimated by the DJI autopilot $(p, v, a)$ to compute the controlled translational velocity $v_{\mathrm{ctrl}}$ and yaw rate $\dot{\psi}_{\mathrm{ctrl}}$.
\begin{align}
    v_{\mathrm{ctrl}} &= K_p \left( p_{\mathrm{ref}} - p \right) + K_v \left( v_{\mathrm{ref}} - v \right) + K_a \left( a_{\mathrm{ref}} - a \right) \\
    \dot{\psi}_{\mathrm{ctrl}} &= K_\psi \left( \psi_{\mathrm{ref}} - \psi \right) + K_{\dot{\psi}} \left( \dot{\psi}_{\mathrm{ref}} - \dot{\psi} \right),
\end{align}
where $K_i$ are user-defined gains.
\reviewchanges{The reference trajectory and DJI state are represented in a local Cartesian \ac{ENU} coordinate system that is set by the DJI autopilot at the start of the mission.
A formal introduction of the DJI \ac{ENU} frame and DJI body frame with calligraphic letters is omitted for brevity.}
Note that since we are sending velocity references to the autopilot, the position error acts like an integration term and the acceleration error acts like a damping term on the control loop.
The control signal is passed through a limiter before being sent to the autopilot to avoid infeasible input commands.

\subsection{Altitude Estimation}
\label{sec:altitude_estimator}
Besides switching between trajectory tracking and waypoint tracking, the \acl{FSM} can also switch between flying with DJI's proprietary altitude estimate and flying relative to the ground surface.
The latter mode is particularly useful when flying radar missions at \SIrange{2}{8}{\metre} because the absolute surface height is not necessarily known and we wish to take measurements from well-defined altitudes.
Furthermore, the regular DJI altitude is subject to large fluctuations which makes terrain tracking control the safer option when flying below \SI{8}{\metre}.
As \reffig{fig:controls} shows, the altitude estimator fuses the DJI altitude with two different altimeter measurements to compute the \acl{AGL} $p_{z,\mathrm{AGL}}$.
Multiple range sensors ensure robustness against outliers and sensor outages.

The implementation of the altitude estimator resembles a Kalman filter, with the change in DJI altitude governing the process model.
Both downward facing radar altimeter range $p_{z,\mathrm{radar}}$ and \ac{LiDAR} altimeter range $p_{z,\mathrm{lidar}}$ serve as independent measurement updates.
Since large roll and pitch angles invalidate the \ac{AGL} measurements, different measures are taken to correct those.
First, the range measurements are corrected by the current roll and pitch angle to represent the vertical distance to the ground assuming a plane environment.
Second, their standard deviation is scaled based on the attitude.
Third, a cutoff attitude rejects measurements when either of the platform's roll or pitch angles exceed a threshold.
Furthermore, the filter implements a Mahalanobis threshold to reject outlier measurements.
And the filter scales the sensor uncertainty based on the magnitude of the measured range, because with increasing height objects in the field of view may invalidate the measurement.

\begin{figure}
    \begin{subfigure}{.49\textwidth}
    \centering
    \includegraphics{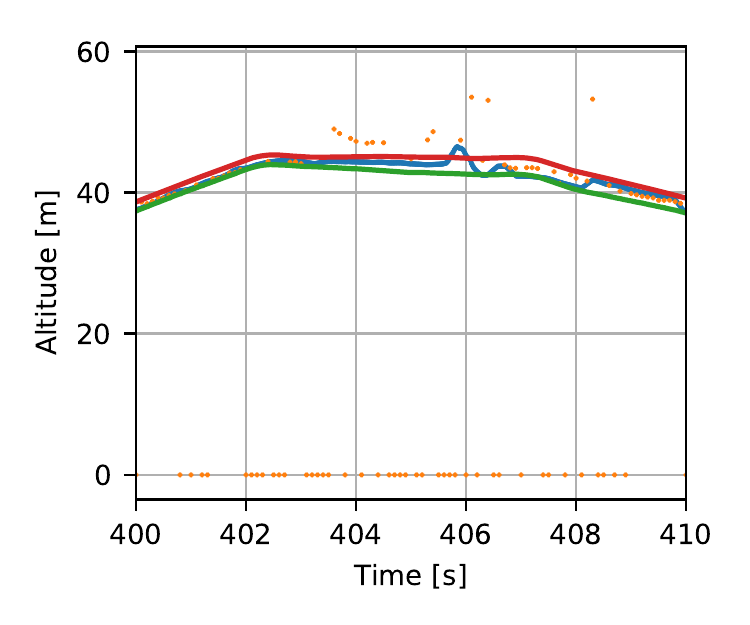}
    \caption{High altitude estimation.}
    \label{fig:high_altitude}
    \end{subfigure} 
    \begin{subfigure}{.49\textwidth}
    \centering
    \includegraphics{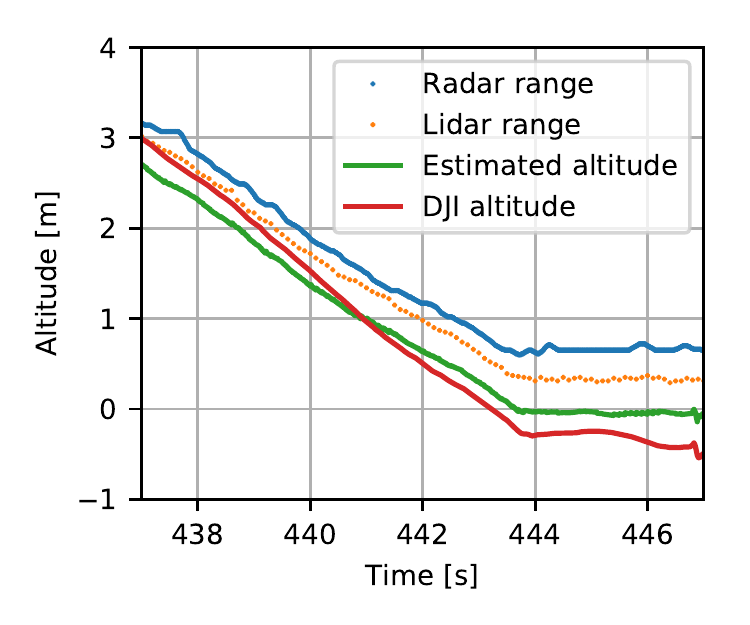}
    \caption{Low altitude estimation.}
    \label{fig:low_altitude}
    \end{subfigure} 
    \caption{The altitude estimator during a typical mission over flat terrain. At high altitude it rejects outlier range measurements. At low altitudes it compensates for DJI altitude drift, here allowing safe landing at \SI{0}{\metre}.}
    \label{fig:altitude}
\end{figure}

\reffig{fig:altitude} shows two example segments of the resulting altitude estimate.
In the high altitude segment in \reffig{fig:high_altitude}, the filter shows robustness to outlier range measurements.
The \ac{LiDAR} does not have the necessary maximum range and is prone to large roll and pitch angles during acceleration movements.
The radar altimeter has a relatively large measurement cone and at $t=\SI{406}{\second}$ measures the wrong \ac{AGL}.
The estimator completely ignores the invalid \ac{LiDAR} measurments and only slowly adapts to the radar altimeter measurements while trusting the DJI altitude most.
In the low altitude segment in \reffig{fig:low_altitude}, when the distance to the surface decreases, the quality of the sensor measurements improves.
The filter trusts the range measurements more than the DJI altitude which in this case allows safe landing at \SI{0}{\metre} altitude.
Note, that during this flight the DJI altitude drifted about \SI{50}{\centi\metre} from take off to landing.
Relying only on the DJI altitude would thus potentially lead to crashes at low altitude.

\section{Time Synchronization}
\label{sec:timing}
\begin{figure}
\centering
\includegraphics{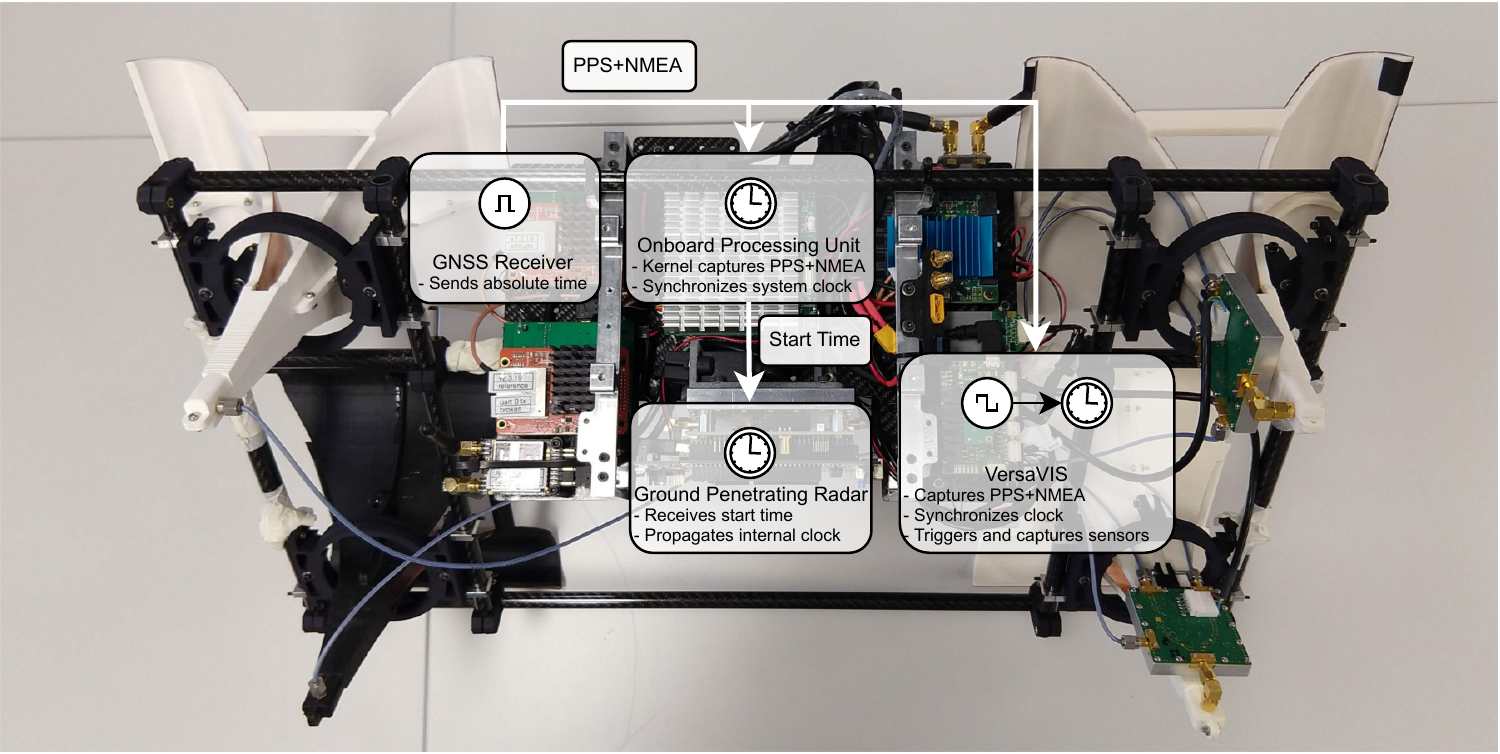}
\caption{Top view of the sensor pod overlayed by the sensor timing overview. All sensors are time stamped with respect to \ac{GNSS} time. The radar receives the current \ac{GNSS} time from the onboard processing unit. The \textit{VersaVIS} directly synchronizes with respect to the \ac{PPS} and time stamps and triggers all other sensors.}
\label{fig:timingoverview}
\end{figure}

% Time synchronization
An important aspect in sensor fusion is time synchronization, that is, triggering and time stamping sensors accurately through the same clock source. 
For high-quality \ac{GPSAR} reconstruction it is particularly important to accurately time stamp the radar measurement with respect to the \ac{GNSS} and the \ac{IMU} to determine the exact radar antenna poses with every chirp.
In our system, all navigation sensors and radar messages are time-stamped with a globally consistent time (\ac{GNSS} time) to allow accurate motion estimation. 
\reffig{fig:timingoverview} shows an overview of the timing modalities.

The \ac{GNSS} provides the time reference of the system and has a typical precision of \SI{50}{\nano\second} \cite{kaplan2005understanding}.
Naturally, all \ac{GNSS} measurements have a \ac{GNSS} time stamp.
All other sensors are time stamped with respect to the onboard processing unit or \textit{VersaVIS} clock.
To synchronize these two systems, one of the \ac{GNSS} receivers emits a \ac{PPS} signal.
On the onboard processing unit a Linux kernel interrupt captures the \ac{PPS} signal and the \textit{chrony} protocol synchronizes its clock~\cite{chrony1997}.
On the \textit{VersaVIS} \ac{MCU} the \ac{PPS} continuously triggers a control loop that synchronizes a clock derived from a \SI{10}{\mega\Hz} external oscillator.
All navigation sensors and the RGB camera are triggered and time stamped through counters derived from the same oscillator.
This allows sensor timing precisions as fine as \SI{0.1}{\micro\second} with respect to \ac{GNSS} as shown in our evaluations below.
The ground penetrating radar system is not interfaced by the \textit{VersaVis} board, as the radar driver board has its own internal clock that is synchronized to the primary onboard clock.
It receives the onboard processing unit's time through \acs{USB} at the beginning of a measurement stream and then open-loop propagates its internal clock.
Due to the clock drift, this accumulates a clock offset of approximately \SI{10}{\milli\second} during a \SI{15}{\minute} mission which corresponds to \SI{1}{\ppm}.

\subsection{VersaVIS to \ac{GNSS} Synchronization}
\begin{figure}
    \begin{subfigure}{2.9in}
    \centering
    \includegraphics{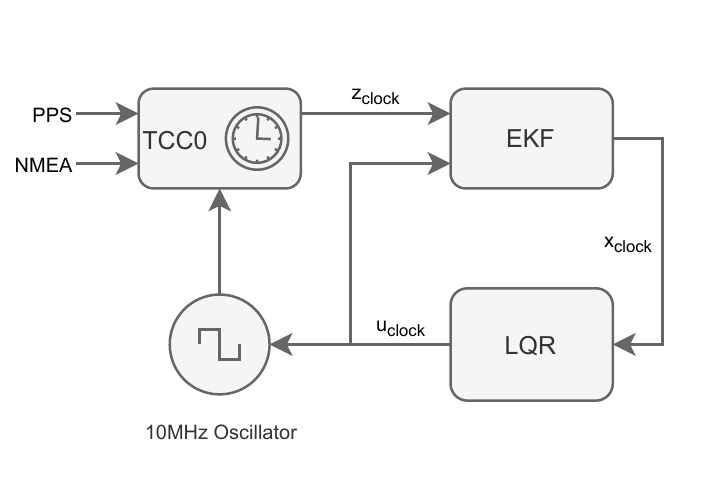}
    \caption{The \textit{VersaVIS} clock control loop.}
    \label{fig:clock_control}
    \end{subfigure} 
    \begin{subfigure}{.54\textwidth}
    \centering
    \includegraphics{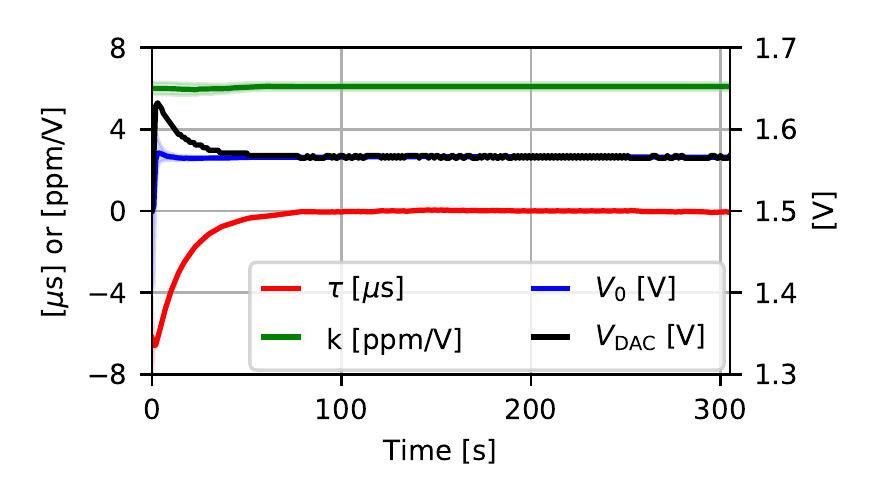}
    \caption{A typical synchronization response with $2\sigma$-bounds.}
    \label{fig:clock_converge}
    \end{subfigure} 
    \caption{A \ac{PPS} driven control loop continuously steers the \textit{VersaVIS} clock offset from \ac{GNSS} time to zero. The controller modifies the frequency of an external oscillator that drives the clock.}
    \label{fig:clock_sync}
\end{figure}

The \textit{VersaVIS} has an internal record of time by counting the pulses coming from the \SI{10}{\mega\Hz} external oscillator on a \SI{24}{\bit} \ac{TCC}.
The first \ac{PPS} pulse in conjunction with its \ac{NMEA} time and date sentence initializes this process.
The \ac{TCC} then wraps around every \SI{e7}{} ticks to increment one second.
Synchronization delays on the \ac{MCU} introduce an initial offset between \ac{PPS} and internally propagated time.
Furthermore, temperature changes and oscillator resolution cause a time varying clock drift.
Hence, a \SI{1}{\Hz} control loop steers the external oscillator frequency to drive the time offset to zero. 
\reffig{fig:clock_sync} shows the closed-loop system that synchronizes the \textit{VersaVIS} to \ac{GNSS} time and a typical control response.

The key idea behind the synchronization mechanism shown in \reffig{fig:clock_control} is that applying an input voltage to the external oscillator changes its frequency and allows driving the difference between internally tracked time and \ac{GNSS} time to zero.
With every captured \ac{PPS} pulse \iac{EKF} estimates the current time offset $\tau$, the constant nominal control voltage $V_0$ to set the oscillator frequency to exactly \SI{10}{\mega\Hz} and the constant factor $\cclk$ that converts oscillator input voltages to expected clock drift.
The measurement input is the time difference between when \iac{PPS} pulse was captured by the \ac{TCC} and its actual time according to the \ac{NMEA} signal.
The following non-linear continuous-time state-space equations with state variable $\xclk$, control input $\uclk$ and measurement input $\zclk$ describe the system:
\begin{align}
    \xclk &= 
    \begin{bmatrix}
        \tau & V_0 & \cclk
    \end{bmatrix}^T,
    & \uclk &= V_{\mathrm{DAC}},
     \\
    \xclkdot &=
    \begin{bmatrix}
        \left( V_{\mathrm{DAC}} - V_0 \right) \cclk \\
         0 \\
         0
    \end{bmatrix} + w \label{eq:clock_dynamics}, & w &\sim \mathcal{N}(0, Q),\\ 
    \zclk &= 
    \begin{bmatrix}
        1 & 0 & 0
    \end{bmatrix} \xclk + v, & v &\sim \mathcal{N}(0, R),
\end{align}
where the \ac{DAC} voltage $V_{\mathrm{DAC}}$ is the control input and $w$ and $v$ are normally distributed, additive white noise with variances $Q$ and $R$, respectively.

Given the full state estimate $\xclk$, \iac{LQR} computes and sets the control voltage $V_{\mathrm{DAC}}$ to accelerate or decelerate the external oscillator to drive the estimated time offset $\tau$ to zero.
The optimal gain $K_c$ is computed offline based on clock dynamics in \refequ{eq:clock_dynamics} discretized and linearized about the nominal state.
The output voltage is clamped to remain within oscillator control voltage limits.
\begin{align}
    V_{\mathrm{DAC}} &= -K_c ~ \tau + V_0 
\end{align}
Typically, the control loop converges within \SI{100}{\second} seconds as \reffig{fig:clock_converge} shows.
Here, the tracked time on the \textit{VersaVIS} has an initial delay of \SI{6}{\micro\second}.
Thus the controller increases the input voltage to accelerate the external oscillator.
While the offset $\tau$ is settling, the control input $V_{\mathrm{DAC}}$ settles as well towards the nominal control voltage $V_0$.
During the course of the run, the controller keeps alternating the control voltage stepwise to remain within zero time offset given the resolution of the \ac{DAC}.

\subsection{Sensor Measurement Time Stamps}
\begin{figure}
    \centering
    \begin{subfigure}{\textwidth}
    \centering
    \includegraphics{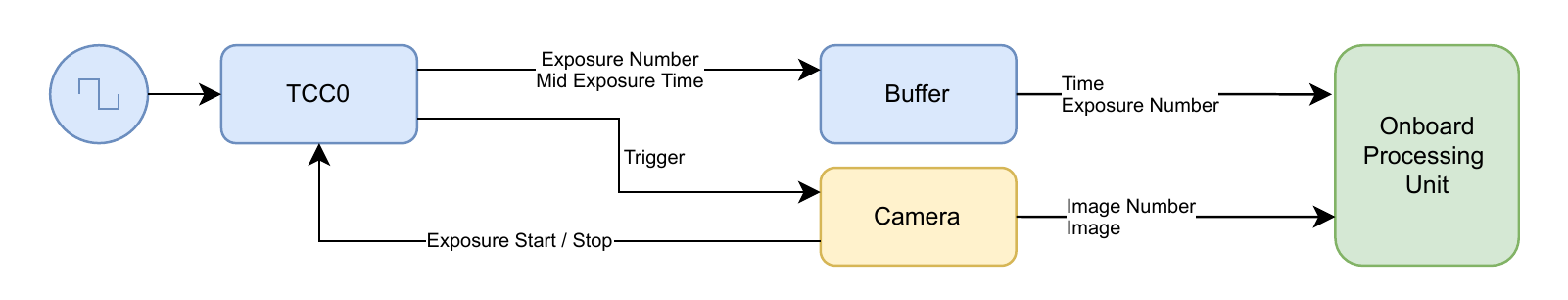}
    \caption{The \ac{TCC}0 triggers the color camera and captures its mid-exposure time. The onboard processing unit matches exposure time stamps with images.}
    \label{fig:timer_bfly}
    \end{subfigure} \\
    \begin{subfigure}{\textwidth}
    \centering
    \includegraphics{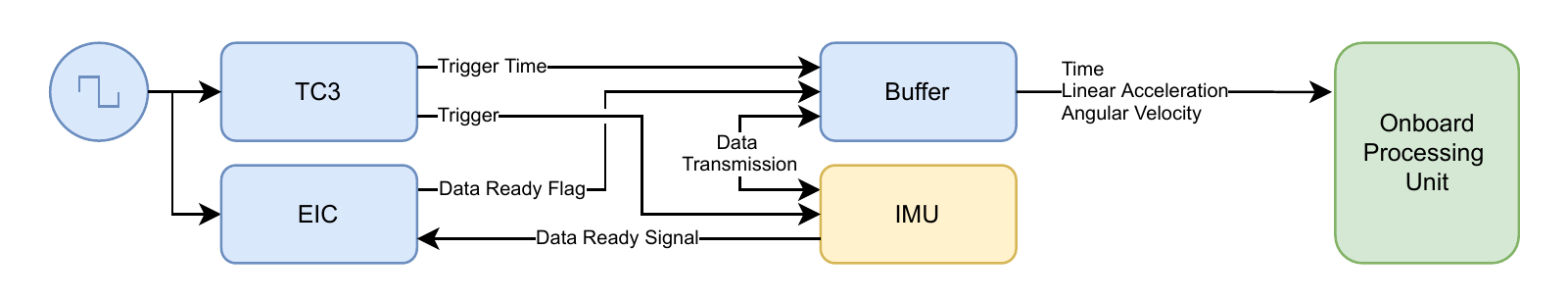}
    \caption{The \ac{TC}3 provides an external clock input to the \ac{IMU}. The data ready signal allows safe data acquisition via \ac{SPI}.}
    \label{fig:timer_imu}
    \end{subfigure} \\
    \begin{subfigure}{\textwidth}
    \centering
    \includegraphics{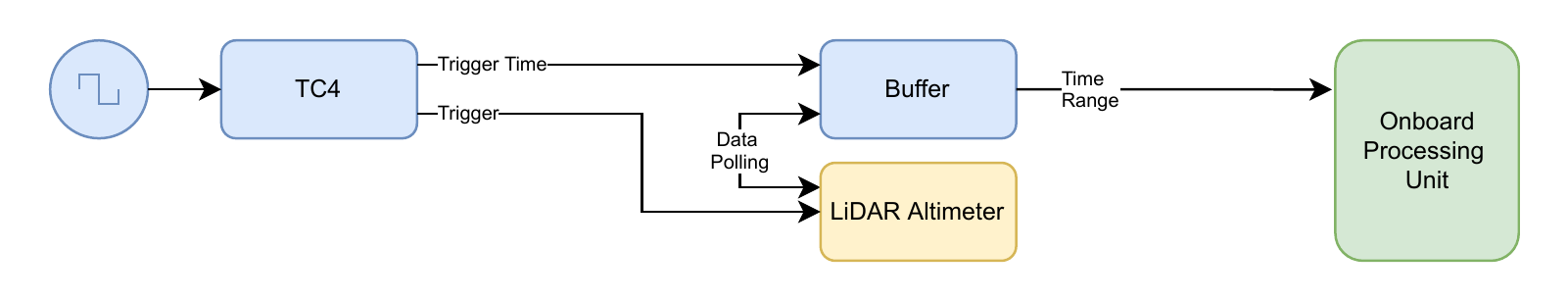}
    \caption{The \ac{TC}4 triggers and stamps \ac{LiDAR} measurements. The measurements are polled via \ac{I2C}.}
    \label{fig:timer_lidar}
    \end{subfigure} \\
    \begin{subfigure}{\textwidth}
    \centering
    \includegraphics{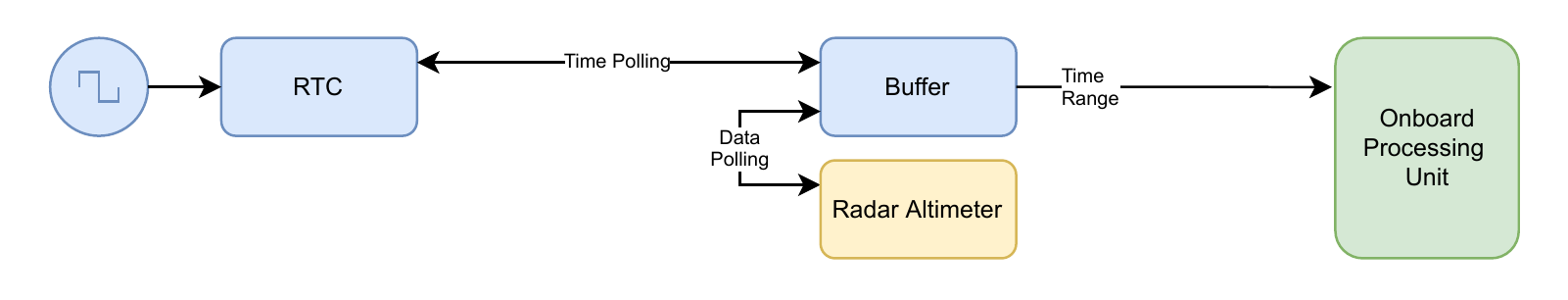}
    \caption{The radar altimeter does not have an external trigger input. It is continuously polled via \ac{UART}. The measurement time is retrieved from \iac{RTC} with \si{\milli\second} precision.}
    \label{fig:timer_radar}
    \end{subfigure} \\
    \begin{subfigure}{\textwidth}
    \centering
    \includegraphics{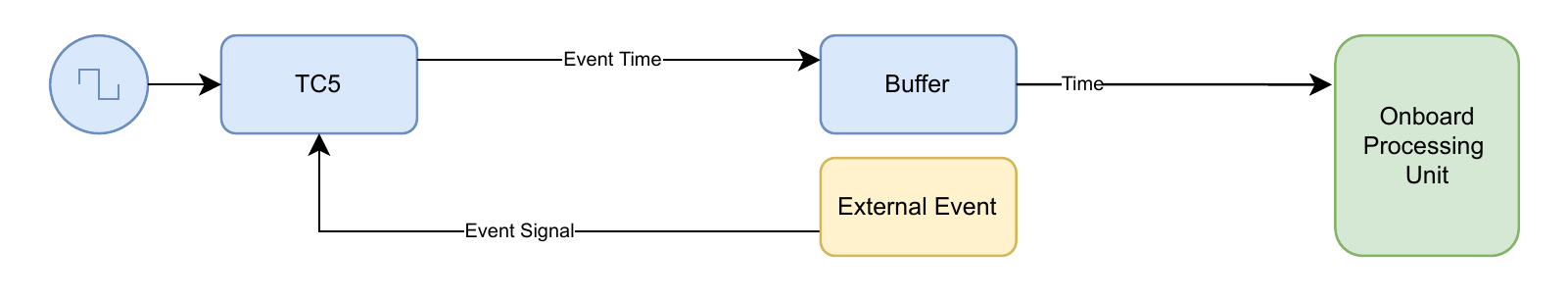}
    \caption{The \ac{TC}5 captures miscellaneous external events on a full resolution counter.}
    \label{fig:timer_ext}
    \end{subfigure}
    \caption{The time stamping and data retrieval mechanisms for the different sensor interfaces. Every sensor has a dedicated counter that time stamps the data synchronously to the known external oscillator ticks.}
    \label{fig:sensor_timers}
\end{figure}

The synchronized \ac{TCC}0 is the basis for the synchronization of all sensors that connect to the \textit{VersaVIS} peripherals.
All \acp{TCC}, \acp{TC} and the \ac{RTC} count the \SI{10}{\mega\Hz} external clock ticks synchronously with the \ac{TCC}0 to generate \acp{PWM} to trigger and to provide capture channels to time stamp sensor measurements.
Typically, the sensors supply \aclp{GPIO} to start a measurement or indicate its exposure.
\reffig{fig:sensor_timers} provides the timing schematics for all sensors that rely on the \textit{VersaVIS} clock.

\reffig{fig:timer_bfly} shows the image capture mechanism of the color camera.
A \ac{PWM} triggers the camera exposure.
The camera returns a strobe and the \ac{TCC} captures start and stop to time stamp the mid-exposure time.
The time stamp and the incremented image counter are stored in a circular buffer and sent to the onboard computer whenever possible.
Based on the image number, the time stamp is associated to an image that is sent to the computer via Ethernet \cite{tschopp2020versavis}. 

\reffig{fig:timer_imu} shows the \ac{IMU} synchronization.
The \ac{TC} generates a \ac{PWM} that serves as an external sampling clock to the \ac{IMU}.
Once the \ac{IMU} has sampled its \ac{MEMS}, it sends out a data ready signal that is captured on the \textit{VersaVIS} \ac{EIC}.
The data ready flag indicates that the measurement can safely be read via \ac{SPI}.
The \textit{VersaVIS} stores the measurement together with the trigger time in a circular buffer and sends it out to the onboard computer.

The \ac{LiDAR} synchronization shown in \reffig{fig:timer_lidar} works similarly.
The \ac{TC} generates a \ac{PWM} to engage a measurement with known time stamp.
After triggering, the \ac{MCU} polls the \ac{I2C} to receive the associated range measurement.
Both time stamp and measurement are stored in a circular buffer and sent to the onboard computer.

The radar altimeter does not have a particular synchronization interface.
Thus we poll the \ac{UART} when resources are available and time stamp the measurement via \ac{RTC} as shown in \reffig{fig:timer_radar}.
Furthermore, our firmware provides an external event channel.
\reffig{fig:timer_ext} shows how a \ac{TC} captures the external input and forwards the event time to the system.
We use this mechanism to evaluate the absolute timing accuracy of the proposed sensor synchronization scheme.

\subsection{Timing Evaluation}
\begin{figure}
\begin{subfigure}{2.9in}
\centering
\includegraphics{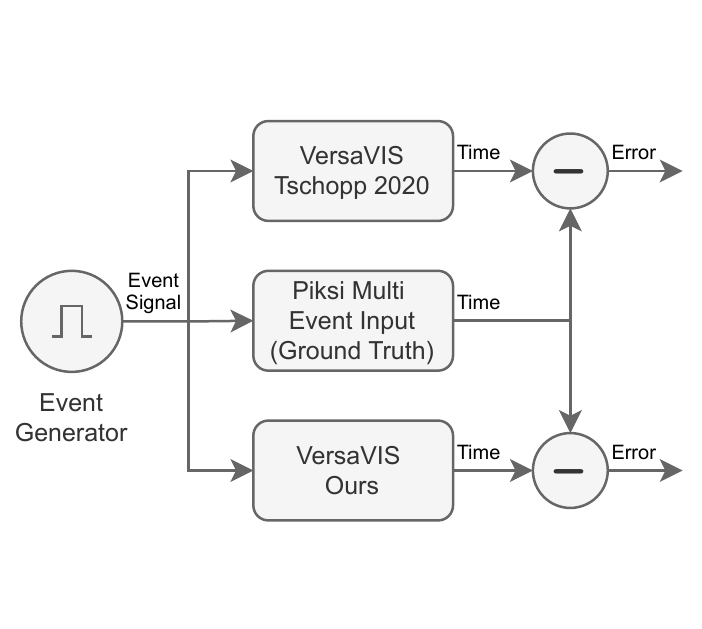}
\caption{External signal and ground truth comparison.}
\label{fig:timesyncevaluation}
\end{subfigure}
\begin{subfigure}{.54\textwidth}
\centering
\includegraphics{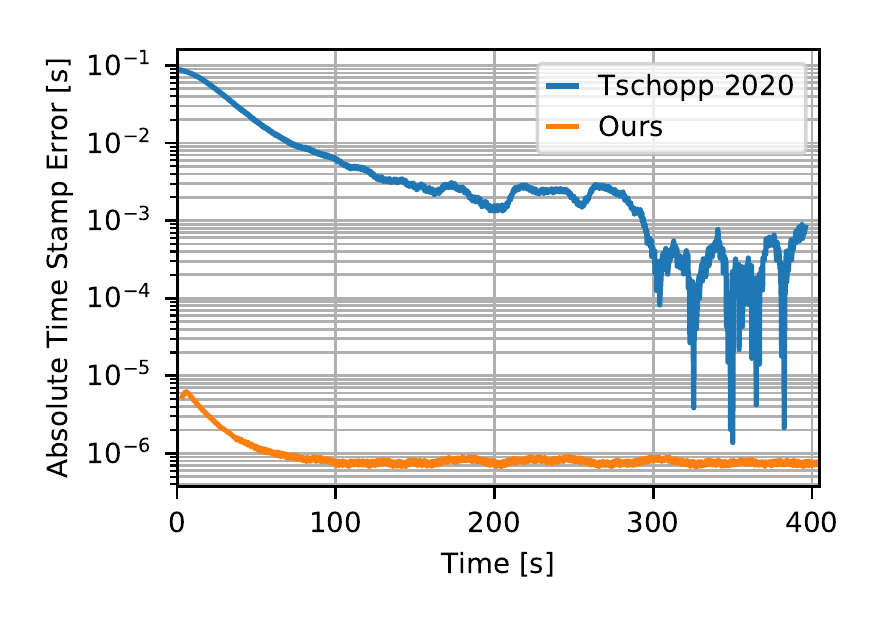}
\caption{Time stamping error over time.\\}
\label{fig:stamperrorseries}
\end{subfigure}
\centering
\caption{Time stamp accuracy evaluation. Our hardware-capture-based synchronization method reaches sub \SI{}{\micro\second} accuracy. In comparison to the previous software-based method it shows better accuracy, precision, and convergence time.}
\end{figure}

We compared our \ac{MCU} timing architecture to the original software trigger presented by \cite{tschopp2020versavis} both in terms of accuracy and precision.
A \SI{5}{\Hz} external signal generator was used to simulate a generic sensor strobe signal as depicted in \reffig{fig:timesyncevaluation}.
The rising edge of the signal was time stamped on both firmware versions and on a ground truth Piksi Multi receiver which has a timing accuracy of \SI{\pm60}{\nano\second}~\cite{piksi2019}.
The resulting time stamps were compared to the ground truth time stamps.
The error over time in \reffig{fig:stamperrorseries} shows that our system is capable of time stamping sensor data with a mean accuracy of \SI{0.8}{\micro\second} and a standard deviation of \SI{0.05}{\micro\second}.
The original firmware, which synchronizes only with respect to the onboard computer via \ac{USB} and not directly to \ac{PPS}, reached a mean accuracy of \SI{180}{\micro\second} and standard deviation of \SI{312}{\micro\second} after \SI{300}{\second} convergence time as opposed to \SI{100}{\second}.
Our \SI{0.05}{\micro\second} precision shows that we can consistently time stamp sensor data with a full clock resolution of \SI{0.1}{\micro\second} provided by the \SI{10}{\mega\hertz} oscillator.
The remaining \SI{0.8}{\micro\second} time stamp offset probably results from a combination of signal edge rise times, \ac{GNSS} \ac{PPS} signal accuracy and steady state offset of our synchronization controller.

Note that, in practice every sensor will introduce some small uncompensated delay in addition to the \SI{0.8}{\micro\second} system accuracy.
For example, the \ac{IMU} requires \SI{100}{\micro\second} to sample and average its \ac{MEMS}~\cite{adis2018}.
\reviewchanges{In the case of \ac{GNSS}-inertial navigation this additional delay is negligible. 
First, because other error sources, e.g., the gravity model and frame vibrations induced by rotor rotations, outweigh such small timing inaccuracies (see~\refsec{sec:radar_calibration}).
Second, because the timing accuracy is still sufficient given the vehicle dynamics.
According to~\cite{ding2008time}, \ac{IMU} time delay $\Delta t$ does not significantly alter the localization result if it fulfills the following inequality:
\begin{equation}
    \lvert \Delta t \rvert  \ll \frac{\lvert h_{pos}\left(x, c\right) - z_{pos} \rvert}{\lvert \iint j~dt^2 \rvert},
\end{equation}
where the numerator represents the measurement innovation during a \ac{GNSS} update and the denominator represents the double integrated change in acceleration (jerk) in between two \ac{GNSS} measurements.
Small timing errors are required either by high-accuracy sensors which lead to a small numerator or highly dynamic systems which lead to a large denominator.
If we assume perfect \ac{IMU} integration and \ac{GNSS} lever arm calibration, our system's lower bound on the innovation is determined by the \ac{RTK} \ac{GNSS} measurement error which is expected to be about~\SI{5}{\milli\metre}.
The maximum jerk we determined in regular operation was about~\SI{100}{\metre\per\second\cubed}.
Double integration of the jerk over a \ac{GNSS} measurement period of \SI{0.1}{\second} generates a velocity of \SI{0.5}{\metre\per\second}.
Thus the time delay has to be significantly smaller than \SI{10}{\milli\second}, i.e., at most \SI{1}{\milli\second}.
When considering half of the \SI{100}{\micro\second} \ac{IMU} sampling period as uncompensated delay, \SI{0.8}{\micro\second} synchronization offset and $2\sigma$-jitter of \SI{0.1}{\micro\second}, our system has an uncompensated delay of \SI{50.9}{\micro\second} between \ac{IMU} and \ac{GNSS} measurements, approximately 1/20th of the required \SI{1}{\milli\second}.
This margin prepares it for future use cases when motion sensing will become more precise (smaller numerator) and greater flight dynamics will be reached (greater denominator).
Following the same calculation, the original firmware from \cite{tschopp2020versavis} with \SI{180}{\micro\second} synchronization offset and \SI{624}{\micro\second} $2\sigma$-jitter has an expected maximum uncompensated delay of \SI{854}{\micro\second} which would still be borderline sufficient.
}

\begin{figure}
    \centering
    \includegraphics[width=.99\linewidth]{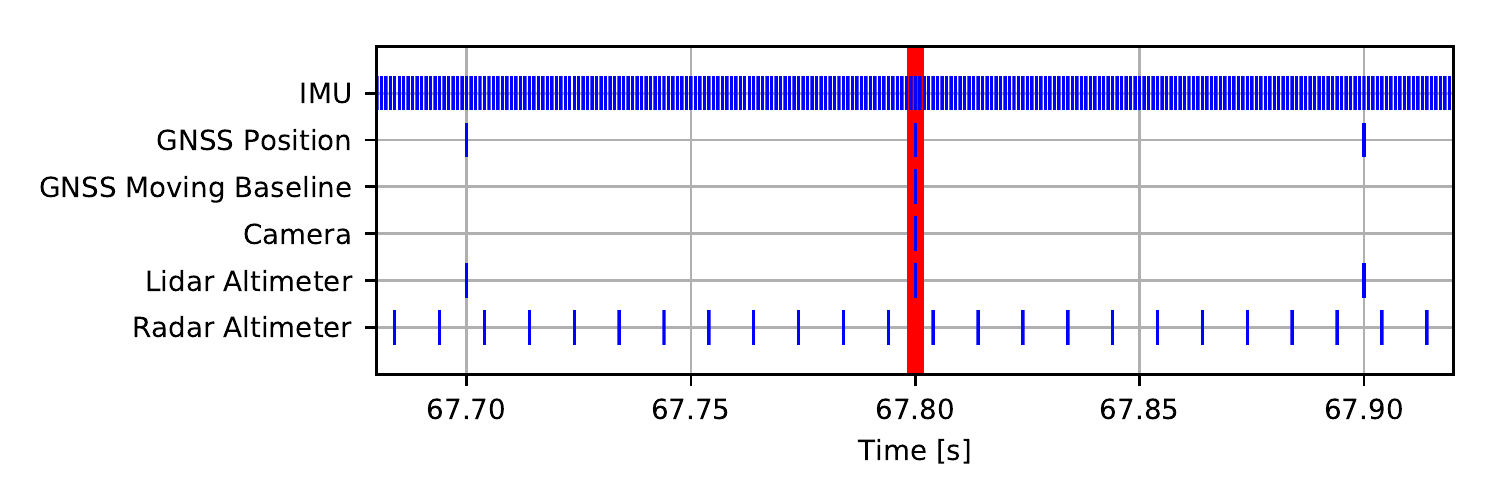}
    \caption{Timeline of \SI{200}{\milli\second} synchronized sensor data time stamps. The vertical red line indicates simultaneous capture of all sensors but the polled radar altimeter at GNSS time stamps. The \ac{IMU} is sampled uniformly every \SI{1}{\milli\second}.}
    \label{fig:sensor_timeline}
\end{figure}

Finally, \reffig{fig:sensor_timeline} shows that all sensors except the polled US-D1 radar altimeter can be simultaneously evaluated at \ac{GNSS} times due to the parallel running hardware triggering and capture mechanism.
This also includes an image exposure compensation scheme as presented in \cite{nikolic2014synchronized} that ensures mid-exposure stamps of the camera at desired \ac{GNSS} times.
The parallel processing leads to jitter-free sensor sampling with well-defined time stamps.
For example, the \ac{IMU} is evaluated uniformly every \SI{1}{\milli\second}. 
\reviewchanges{As we will see in \refsec{sec:algorithm}, having deterministic sensor time stamps simplifies the fusion algorithm design.}

\section{Localization} \label{sec:localization}
\begin{figure}
    \begin{subfigure}{.45\textwidth}
        \centering
        \includegraphics[height=2.5in]{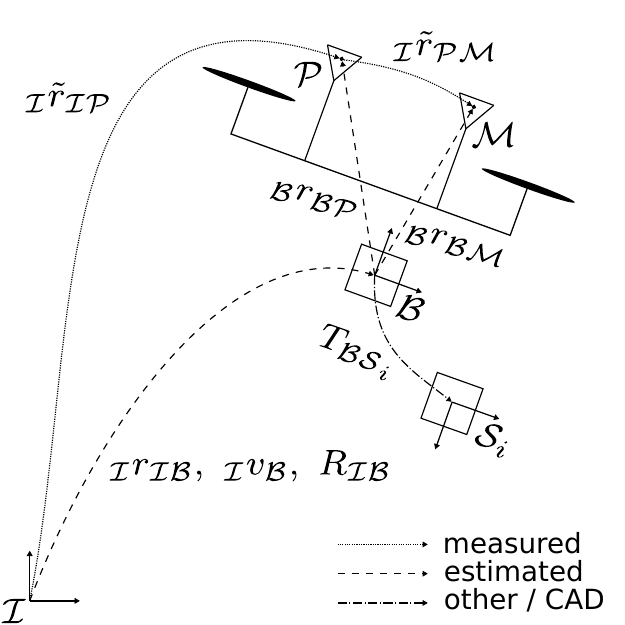}
        \caption{The geometric relationships.}
        \label{fig:kinematic}
    \end{subfigure}
    \begin{subfigure}{.54\textwidth}
        \centering
        \includegraphics[height=2.5in]{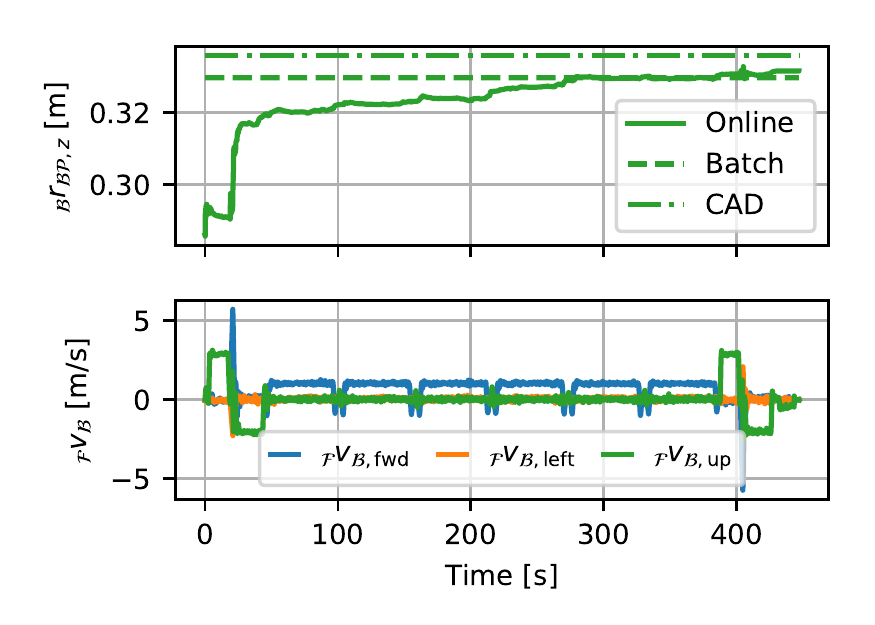}
        \caption{\ac{GNSS} antenna position self-calibration.}
        \label{fig:ant_estimate}
    \end{subfigure}
    \caption{The localization algorithm estimates the 3D pose of the \ac{MAV} and its sensors.
    It can self-calibrate \ac{GNSS} antenna positions and \ac{IMU} biases.}
    \label{fig:model}
\end{figure}

For localization our robot uses a \SI{1}{\kilo\Hz} \ac{IMU}, a \SI{10}{\Hz} \ac{RTK} \ac{GNSS} position receiver, and a \SI{5}{\Hz} \ac{RTK} \ac{GNSS} moving baseline receiver.
The \ac{IMU} measures the linear acceleration ${{}_{\mathcal{B}}\Tilde{a}_{\mathcal{B}} \in \mathbb{R}^3}$ and angular velocity ${{}_{\mathcal{B}}\Tilde{\omega}_{\mathcal{I}\mathcal{B}} \in \mathbb{R}^3}$ of the platform in body coordinates $\mathcal{B}$, where $\mathcal{B}$ aligns with the measurement axes of the \ac{IMU}.
The position receiver measures the position ${{}_{\mathcal{I}}\Tilde{r}_{\mathcal{IP}} \in \mathbb{R}^3}$ of \ac{GNSS} antenna $\mathcal{P}$ with respect to inertial frame $\mathcal{I}$.
Without loss of generality we define frame $\mathcal{I}$ to correspond to a cartesian frame with orientation aligned with the local \acl{ENU} frame and origin at the \ac{RTK} \ac{GNSS} base station position.
The moving baseline receiver measures the baseline vector ${{}_{\mathcal{I}}\Tilde{r}_{\mathcal{PM}} \in \mathbb{R}^3}$ between \ac{GNSS} antennas $\mathcal{P}$ and $\mathcal{M}$ rigidly attached to the platform, see \reffig{fig:kinematic}.

These four measurements let us estimate the sensor pod's position ${{}_{\mathcal{I}}r_{\mathcal{IB}} \in \mathbb{R}^3}$, velocity ${{}_{\mathcal{I}}v_{\mathcal{B}} \in \mathbb{R}^3}$, and orientation ${R_{\mathcal{I}\mathcal{B}}} \in SO(3)$.
The navigation state vector $x$ concatenates these quantities.
We choose an \ac{IMU}-centered navigation framework, where the state represents the state of the \ac{IMU} frame $\mathcal{B}$.
All three state quantities are expressed with respect to the inertial frame $\mathcal{I}$.
Apart from the sensor pod state, we estimate sensor-model-specific calibration parameters $c$.
These are the slowly changing \ac{IMU} accelerometer biases $b_a \in \mathbb{R}^3$, gyroscope biases $b_g \in \mathbb{R}^3$, the constant position receiver antenna phase center ${}_{\mathcal{B}}r_{\mathcal{BP}} \in \mathbb{R}^3$ and the moving baseline receiver antenna phase center ${}_{\mathcal{B}}r_{\mathcal{BM}} \in \mathbb{R}^3$. 
Note that, for simplicity, we assume a single phase center for each \ac{GNSS} antenna, even though the receivers utilize $L_1$, $L_2$, and $L_5$ bands.
\reffig{fig:ant_estimate}  shows an example self-calibration, where the \ac{GNSS} antenna position is refined during the flight to improve global accuracy of the estimate.
\begin{align}
    x &= \begin{bmatrix}
        {}_{\mathcal{I}}r_{\mathcal{IB}} ,~ {}_{\mathcal{I}}v_{\mathcal{B}} ,~ R_{\mathcal{I}\mathcal{B}}  
        \end{bmatrix}^T & 
    c &= \begin{bmatrix}
        b_{a},~ b_{g},~  {}_{\mathcal{B}}r_{\mathcal{BP}} ,~ {}_{\mathcal{B}}r_{\mathcal{BM}}
        \end{bmatrix}^T
\end{align}

The following continuous time model determines the propagation of the state vector $x$ and parameters $c$.
The state is modeled based on kinematic relations, the bias parameters of the \ac{IMU} are modeled as Brownian motion with standard deviations $\sigma_{ba}$ and $\sigma_{bg}$, and the antenna phase centers are assumed to be constant. 
\begin{align}
    \dot{x} &= 
    \begin{bmatrix}
    {}_{\mathcal{I}}v_{\mathcal{B}} \\
    {}_{\mathcal{I}}a_{\mathcal{B}} \\
    R_{\mathcal{I}\mathcal{B}} ~ \left[{}_{\mathcal{B}} \omega_{\mathcal{I}\mathcal{B}}\right]_\times
    \end{bmatrix}
    &
    \dot{c} &= 
    \begin{bmatrix}
    \eta_{ba}    \\
    \eta_{bg}  \\
    0 \\
    0 
    \end{bmatrix} 
    & \eta_{ba} &\sim \mathcal{N}(0, \sigma_{ba}^2 ~ I)
    & \eta_{bg} &\sim \mathcal{N}(0, \sigma_{bg}^2 ~ I)
    \label{eq:state}
\end{align}
The $\left[\cdot\right]_\times$ operator maps a vector to its skew symmetric matrix, ${}_{\mathcal{I}}a_{\mathcal{B}}$ is the linear acceleration of the base frame and ${}_{\mathcal{B}}\omega_{\mathcal{I}\mathcal{B}}$ its angular velocity.

Let $\mathcal{X}$, $\mathcal{C}$ and $\mathcal{Z}$ be the sets of all discrete states, parameters, and measurements available up to the current time $T$.
\begin{align}
    \mathcal{X} &= \{ x_i \}_{i=0}^T & \mathcal{C} &= \{ c_i \}_{i=0}^T & \mathcal{Z} &= \{ z_i \}_{i=0}^T
\end{align}
Our estimator constantly reevaluates the \ac{MAP} state estimates $\mathcal{X}^\ast$ and parameter estimates $\mathcal{C}^\ast$ with every new \ac{GNSS} measurement available.
\begin{align}
    \left(\mathcal{X}^\ast, \mathcal{C}^\ast\right) &= \arg \max_{(\mathcal{X},\mathcal{C})} ~ p\left(\mathcal{X},\mathcal{C} \vert \mathcal{Z}\right) \label{eq:joint_pdf}
\end{align}
To solve this inference problem efficiently, we use the GTSAM framework \cite{dellaert2017factor}.
The framework uses factor graphs to model the joint \ac{PDF} $p\left(\mathcal{X},\mathcal{C} \vert \mathcal{Z}\right)$.
Essentially, this decomposes the joint \ac{PDF} into a product of simpler \acp{PDF}, where each \ac{PDF} is dependent only on a single sensor measurement $z_i \in \mathcal{Z}$.
For Gaussian noise distributions the \ac{MAP} estimation problem of \refequ{eq:joint_pdf} can then be rewritten as the minimization of the sum of residuals
\begin{align}
   \arg \min_{(\mathcal{X},\mathcal{C})}  \sum_i \vert\vert h_i(\mathcal{X},\mathcal{C}) - z_i \vert\vert^2_{\Sigma_i},\label{eq:minimization}
\end{align}
where $\vert\vert e \vert\vert^2_{\Sigma} = e^T \Sigma^{-1} e $ is the squared Mahalanobis distance and $\Sigma$ the covariance matrix corresponding to the measurement.
The measurement function $h_i$ defines how the states and parameters are related to the respective measurement and needs to be defined for each sensor modality.
Furthermore, to minimize \refequ{eq:minimization} the partial derivatives of $h_i$ with respect to the state $x$ and parameters $c$ need to be known.

\begin{figure}
    \centering
    \includegraphics[width=.99\linewidth]{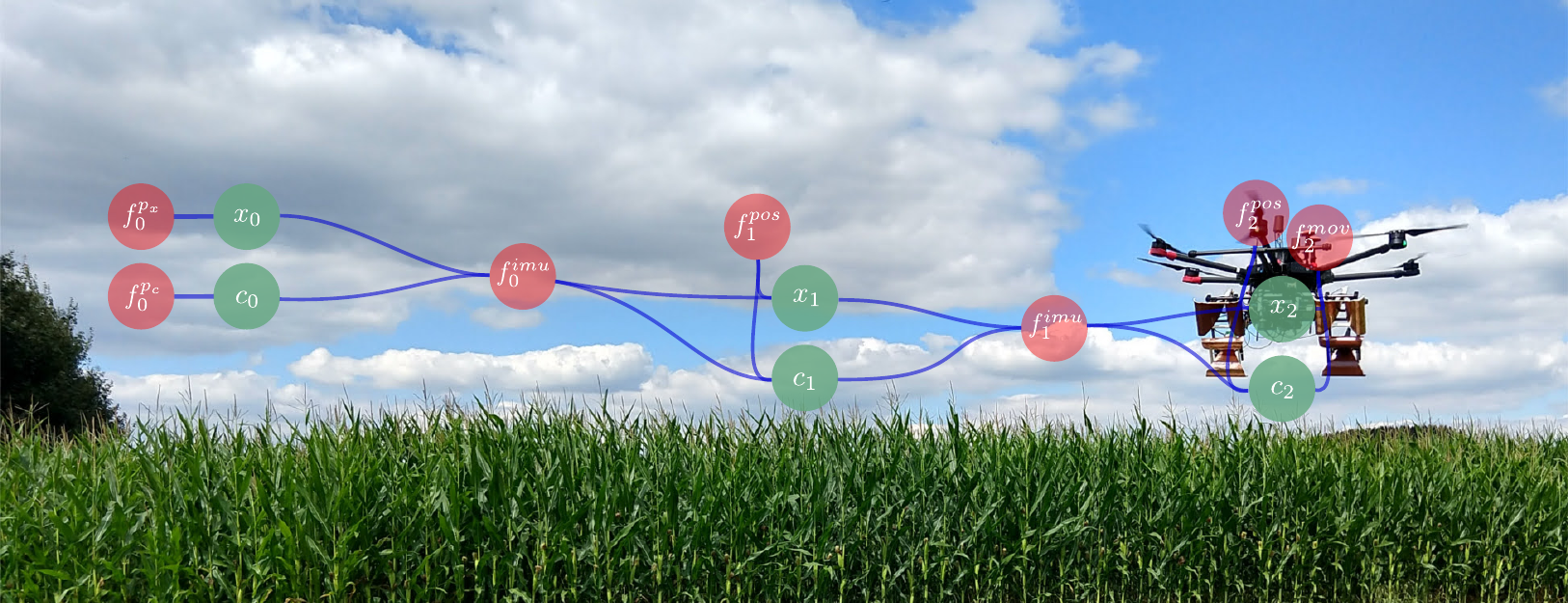}
    \caption{Illustration of the factor graph.}
    \label{fig:graph}
\end{figure}

\reffig{fig:graph} sketches the graph structure of our estimation problem.
The estimation process starts with a prior factor $f_0^{p_x}$ constraining the first state and $f_0^{p_c}$ constraining the belief about the initial calibration parameters.
Discrete optimization variables $x_i$ and $c_i$ are introduced with every \ac{GNSS} measurement.
These are modeled through unary position factors $f_i^{pos}$ and moving baseline factors $f_i^{mov}$.
The odometry between the discrete times is determined by preintegrated \ac{IMU} factors $f_i^{imu}$. In the following, we describe the formulation of each of these factors.
\typeout{* line width: \the\linewidth}

\subsubsection*{Prior Factor}
The states and calibration parameters in the factor graph need to be initialized to start the iterative smoothing.  
The TRIAD method determines the initial orientation of the platform \cite{black1964passive}.
This method computes the rotation matrix between two coordinate frames, i.e., the orientation $R_{\mathcal{IB}}$ of the base frame $\mathcal{B}$ with respect to the inertial frame $\mathcal{I}$, based on two linearly independent vectors expressed in both frames.
The first vector pair is the direction of gravity.
${}_{\mathcal{I}}r_g$, the gravity vector in inertial coordinates, points in the downward-direction.
The corresponding gravity vector in base coordinates, ${}_{\mathcal{B}}r_g$, is the linear acceleration ${}_{\mathcal{B}}\Tilde{a}_{\mathcal{B}}$ measured by the \ac{IMU} at standstill.
\begin{align}
    {}_{\mathcal{I}}r_g &= \begin{bmatrix}0, 0, -1\end{bmatrix}^T &  {}_{\mathcal{B}}r_g &= -{}_{\mathcal{B}}\Tilde{a}_{\mathcal{B}}(0)
\end{align} 
The second vector pair is the baseline vector between the two \ac{GNSS} antennas.
${}_{\mathcal{I}}r_b$, the baseline in inertial coordinates, is measured by the \ac{GNSS} moving baseline receiver ${}_{\mathcal{I}}\Tilde{r}_{PM}$.
The baseline vector in base coordinates is the difference between the attitude antenna phase center and position antenna phase center, which is known approximately a priori from \ac{CAD} data.
\begin{align}
    {}_{\mathcal{I}}r_b &= {}_{\mathcal{I}}\Tilde{r}_{\reviewchanges{\mathcal{PM}}}(0) & {}_{\mathcal{B}}r_b &= {}_{\mathcal{B}}r_{\reviewchanges{\mathcal{BM}}}(0) - {}_{\mathcal{B}}r_{\reviewchanges{\mathcal{BP}}}(0)
\end{align}
Following TRIAD, the initial orientation is,
\begin{align}
    R_{\mathcal{IB}}(0) &=\begin{bmatrix} 
    \frac{{}_{\mathcal{I}}r_g}{\vert\vert{}_{\mathcal{I}}r_g\vert\vert}, \frac{{}_{\mathcal{I}}r_g \times {}_{\mathcal{I}}r_b}{\vert\vert{}_{\mathcal{I}}r_g \times {}_{\mathcal{I}}r_b\vert\vert}, \frac{{}_{\mathcal{I}}r_g}{\vert\vert{}_{\mathcal{I}}r_g\vert\vert} \times \frac{{}_{\mathcal{I}}r_g \times {}_{\mathcal{I}}r_b}{\vert\vert{}_{\mathcal{I}}r_g \times {}_{\mathcal{I}}r_b\vert\vert}
    \end{bmatrix} \cdot \begin{bmatrix} 
    \frac{{}_{\mathcal{B}}r_g}{\vert\vert{}_{\mathcal{B}}r_g\vert\vert}, \frac{{}_{\mathcal{B}}r_g \times {}_{\mathcal{B}}r_b}{\vert\vert{}_{\mathcal{B}}r_g \times {}_{\mathcal{B}}r_b\vert\vert}, \frac{{}_{\mathcal{B}}r_g}{\vert\vert{}_{\mathcal{B}}r_g\vert\vert} \times \frac{{}_{\mathcal{B}}r_g \times {}_{\mathcal{B}}r_b}{\vert\vert{}_{\mathcal{B}}r_g \times {}_{\mathcal{B}}r_b\vert\vert}
    \end{bmatrix}^T.
\end{align}

The initial position is determined from the first \ac{GNSS} position measurement and the initial position receiver phase center position.
\begin{align}
    {}_{\mathcal{I}}r_{IB}(0) &= {}_{\mathcal{I}}\Tilde{r}_{\mathcal{IP}} - R_{\mathcal{IB}}(0) ~ {}_{\mathcal{B}}r_{\reviewchanges{\mathcal{BP}}}(0)
\end{align}
All other parameters, i.e., velocity, gyroscope bias, and accelerometer bias are initialized to zero.
Again we assume the vehicle is stationary at startup.
This also allows us to calibrate the \ac{IMU} gyroscope biases to zero at startup by averaging the first gyroscope measurements.
The same process cannot be assumed for the accelerometer because we cannot ensure a perfectly leveled platform.
The initial uncertainty is left as a tuning parameter.

\subsubsection*{\ac{IMU} Factor}
The \ac{IMU} measures the linear acceleration ${}_{\mathcal{B}}\Tilde{a}_{\mathcal{B}}$ and angular velocity ${}_{\mathcal{B}}\Tilde{\omega}_{\mathcal{I}\mathcal{B}}$ of the platform. 
Ignoring effects due to earth's rotation, the measurements are modelled as,
\begin{align}
    {}_{\mathcal{B}}\Tilde{a}_{\mathcal{B}} &= R_{\mathcal{I}\mathcal{B}}^T ~ \left( {}_{\mathcal{I}}a_{\mathcal{B}} - {}_{\mathcal{I}} g \right) + b_{a} + \eta_a & \eta_a &\sim \mathcal{N}(0, \sigma_a^2 ~ I) \\
    {}_{\mathcal{B}}\Tilde{\omega}_{\mathcal{I}\mathcal{B}} &= {}_{\mathcal{B}}\omega_{\mathcal{I}\mathcal{B}} + b_{g} + \eta_g & \eta_g &\sim \mathcal{N}(0, \sigma_g^2 ~ I),
\end{align}
where ${}_{\mathcal{I}}g$ is the gravity vector in inertial coordinates, $b_{a}$ and $b_{g}$ are the \ac{IMU} biases modeled as random walk in \refequ{eq:state} and $\eta_a$ and $\eta_g$ are additive white noise with standard deviation $\sigma_a$ or $\sigma_g$ respectively.
The set of \ac{IMU} measurements between two \ac{GNSS} measurements are summarized into a single preintegrated factor as described in \cite{forster2015imu}.

\subsubsection*{Position Factor}
The \ac{GNSS} position receiver measures the distance between the inertial frame and position antenna phase center ${}_{\mathcal{I}}\Tilde{r}_{\mathcal{IP}}$.
We can express the position of the position antenna as a function of the current position and orientation estimate.
\begin{align}
    h_{pos}\left(x, c\right) &= {}_{\mathcal{I}}r_{\mathcal{IB}} + R_{\mathcal{IB}} ~ {}_{\mathcal{B}}r_{\mathcal{BP}}
\end{align}
The non-zero partial derivatives \cite[p.~6]{dellaert2020} are,
\begin{align}
    \frac{\partial h_{pos}\left(x, c\right)}{\partial {}_{\mathcal{I}}r_{\mathcal{IB}}} &= I &
    \frac{\partial h_{pos}\left(x, c\right)}{\partial R_{\mathcal{IB}}} &= -R_{\mathcal{IB}} ~ \left[_{\mathcal{B}}r_{\mathcal{BP}}\right]_\times &
    \frac{\partial h_{pos}\left(x, c\right)}{\partial {}_{\mathcal{B}}r_{\mathcal{BP}}} &= R_{\mathcal{IB}}.
    \label{eq:pos}
\end{align}
The partial derivatives show that the position measurement not only gives direct feedback on the position estimate ${}_{\mathcal{I}}r_{\mathcal{IB}}$ but, due to the influence of the lever arm ${}_{\mathcal{B}}r_{\mathcal{BP}}$, also indirectly measures the orientation $R_{\mathcal{IB}}$ under certain motions.
Note that this factor holds for both the \ac{RTK} \ac{GNSS} solution, which measures the baseline between the \ac{RTK} base station antenna and rover antenna, and the \ac{SBAS} \ac{GNSS} solution, which measures the position in the world-fixed WGS84 frame, under the assumption that the \ac{RTK} base station position is determined bias-free with respect to the WGS84 frame.
Thus the estimator is robust to switching between the \ac{RTK} fixed and the \ac{SBAS} solution.

\subsubsection*{Moving Baseline Factor}
Since the platform orientation from a single \ac{GNSS} receiver is only observed under certain motions, a second \ac{GNSS} receiver is added that measures the baseline vector ${_{\mathcal{I}}\Tilde{r}_{\mathcal{PM}}}$ between two \ac{GNSS} antennas rigidly attached to the platform. 
This measurement gives direct information about the orientation.
The measurement is actually given at the local WGS84 tangent plane centered at the position antenna $\mathcal{P}$ but we assume that for small distances from the inertial frame, this tangent plane is parallel to the inertial frame's east-north-plane.
The resulting moving baseline measurement function maps the current estimated difference of the \ac{GNSS} antenna positions into in the inertial frame.
\begin{align}
    h_{mov}\left(x, c\right) &= R_{\mathcal{IB}} ~ \left( {}_{\mathcal{B}}r_{\mathcal{BM}} - {}_{\mathcal{B}}r_{\mathcal{BP}} \right)
\end{align}
The non-zero partial derivatives \cite[p.~6]{dellaert2020} are,
\begin{align}
    \frac{\partial h_{mov}\left(x, c\right)}{\partial R_{\mathcal{IB}}} &= R_{\mathcal{IB}} ~ \left[ {}_{\mathcal{B}}r_{\mathcal{BP}} - {}_{\mathcal{B}}r_{\mathcal{BM}} \right]_\times &
    \frac{\partial h_{mov}\left(x, c\right)}{\partial {}_{\mathcal{B}}r_{\mathcal{BP}}} &= -R_{\mathcal{IB}} & 
    \frac{\partial h_{mov}\left(x, c\right)}{\partial {}_{\mathcal{B}}r_{\mathcal{BM}}} &= R_{\mathcal{IB}}.
    \label{eq:att}
\end{align}
The first partial derivative states that the orientation perpendicular to the moving baseline vector, in our case roll and heading, is directly measureable through the moving baseline factor.
The other two derivatives, similar to the last derivative in \refequ{eq:pos}, indicate, that the \ac{GNSS} antenna positions are observable under orientation changes.
This observability is validated in \reffig{fig:ant_estimate}.
Here the antenna position estimate is plotted alongside the velocity of the base frame represented in \acl{FLU} coordinates $\mathcal{F}$.
The antenna position is particularly well observed during fast forward flight at $t=\SI{21}{\second}$ and $t=\SI{405}{\second}$, i.e., where the platform tilts heavily to generate accelerations.

\reviewchanges{\subsection{Algorithm}\label{sec:algorithm}}
The localization algorithm runs in three parallel processing threads.
The highest priority is given to the navigation thread \cite{indelman2013information}.
This thread integrates incoming \ac{IMU} measurements given the latest available \ac{MAP} state estimate and delivers a high rate odometry estimate for control purposes.

The second thread iteratively solves the inference problem stated in \refequ{eq:joint_pdf}.
Given the synchronization between \ac{IMU} and \ac{GNSS}, and the fixed \ac{GNSS} solution rate, the time stamps of new factors are known a priori.
When a new \ac{IMU} measurement arrives, it is preintegrated into a new \ac{IMU} factor up to the next expected \ac{GNSS} measurement time stamp.
Once the \ac{IMU} factor is completed, it is inserted into the factor graph.
The \ac{GNSS} measurements typically arrive delayed and are inserted at the respective positions into the factor graph.
The solver is polled whenever a new factor is added to the graph and the thread is available.
We use ISAM2 with a configurable sliding window to reduce inference time and bound graph size and thus relinearization efforts \cite{kaess2012isam2}.

The third thread can be called on demand, e.g., after completing the mission, and computes the full batch solution for \ac{GPSAR} mapping purposes.
While running the sliding window estimator, the program stores a copy of the full graph, the estimator solution, and all \ac{IMU} measurements.
When the batch solver is triggered, it uses a Levenberg-Marquardt optimizer to find the \ac{MAP} estimate given the full graph and prior online solution.
Additionally, it predicts intermediate states propagating the stored \SI{1}{\kilo\Hz} \ac{IMU} measurements.
The sensor poses $T_{\mathcal{IS}_i}$, i.e., the ground penetrating radar antenna phase center and RGB camera poses, can then be evaluated given the known sensor locations with respect to the \ac{IMU} $T_{\mathcal{BS}_i}$,
\begin{align}
T_{\mathcal{IS}_i} &= T_{\mathcal{IB}}~T_{\mathcal{BS}_i}.
\end{align}

\section{Results}\label{sec:results}
In this section we evaluate the state estimation accuracy and overall functionality of the proposed system.
\reviewchanges{As \refsec{sec:SAR} lays out, the final application of airborne buried object detection is a function of the flight path, antenna localization, knowledge of the terrain, and millimeter wave signal processing.
Thus, evaluating individual components with respect to the downstream task is impossible.
Instead, we design individual experiments that allow testing the components presented earlier in isolation.
The goal is to practically verify the correct implementation of each sub-system, set a system performance benchmark, and last but not least determine system limitations and performance margins.
This methodology allows the reader to draw conclusions for future system designs.}

In the first experiment, we validate the algorithmic soundness of the derived estimator in a ground truth motion tracking setting.
We demonstrate that the introduction of the second \ac{GNSS} receiver improves positioning accuracy, in particular for initialization.
Furthermore, the experiment shows that the full batch inference computed after landing improves overall estimation smoothness and accuracy over online sliding window estimation. 
Following this, we conducted two field trials with the \ac{GPSAR}.
First we present mapping of known radar calibration targets placed on the ground surface.
The experiments \reviewchanges{show} an overall system sensitivity that allows identifying the scattering characteristics of individual targets.
The results also highlight the importance of accurate \ac{GNSS} antenna calibrations.
\reviewchanges{Furthermore, the experiments suggest that the presented localization method exceeds the resolving power of the underlying imaging solution.
As a result of this we present and evaluate a simple alternative solution to localize the radar antennas that still generates focused radar images.}
Finally, we show the detection of buried metal can lids as a proof-of-concept for landmine detection from an aerial vehicle.

\subsection{Localization Ground Truth Evaluation}\label{sec:ground_truth}
\begin{figure}
    \begin{subfigure}{.49\textwidth}
        \centering
        \includegraphics{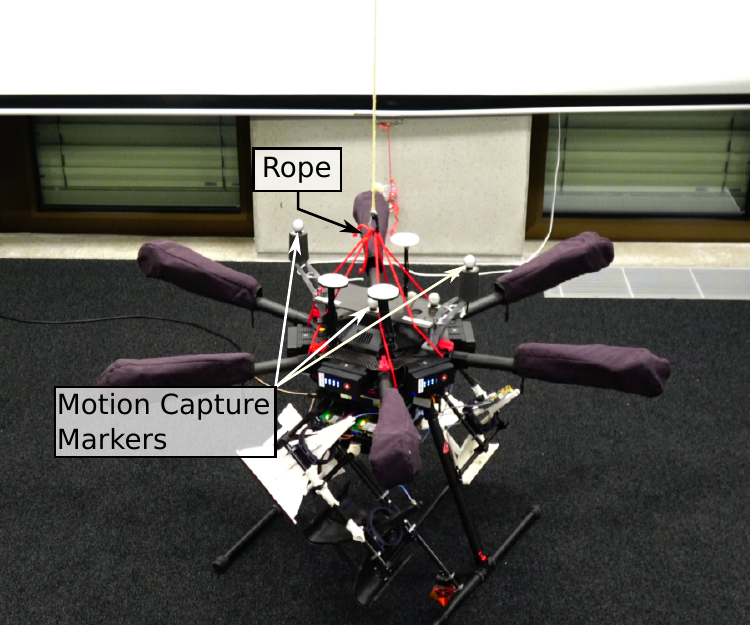}
        \caption{Motion tracking setup.}
        \label{fig:gt_setup}
    \end{subfigure}
    \begin{subfigure}{.49\textwidth}
        \centering
        \resizebox{\textwidth}{!}{%
        \begin{tabular}{l c r r}
        \toprule 
            Configuration & Legend & \acs{RMSE} & Improvement \\
        \midrule
            Ground truth & \legenddot{E377C2} & \SI{0}{\milli\metre} & \SI{100}{\percent}\\
            Batch (position + moving baseline) & \legendline{D62728} & \textbf{\SI{4.24}{\milli\metre}} & \textbf{\SI{59}{\percent}}\\
            Batch (position only) & \legendline{9467BD} & \SI{5.02}{\milli\metre} & \SI{52}{\percent} \\
            Batch (position only + heading offset) & \legendline{8C564B} & \SI{6.29}{\milli\metre} & \SI{40}{\percent} \\
            Online (position + moving baseline) & \legendline{1F77B4} & \SI{6.74}{\milli\metre} & \textbf{\SI{36}{\percent}} \\
            Online (position only) & \legendline{FF7F0E} & \SI{9.41}{\milli\metre} & \SI{10}{\percent} \\
            Online (position only + heading offset) & \legendline{2CA02C} & \SI{10.46}{\milli\metre} & \SI{0}{\percent} \\
        \bottomrule
        \end{tabular}}
        \caption{\ac{RMSE} of the different configurations.}
        \label{tab:rmse}
    \end{subfigure}
    \begin{subfigure}{.49\textwidth}
        \centering
        \includegraphics[height=2in]{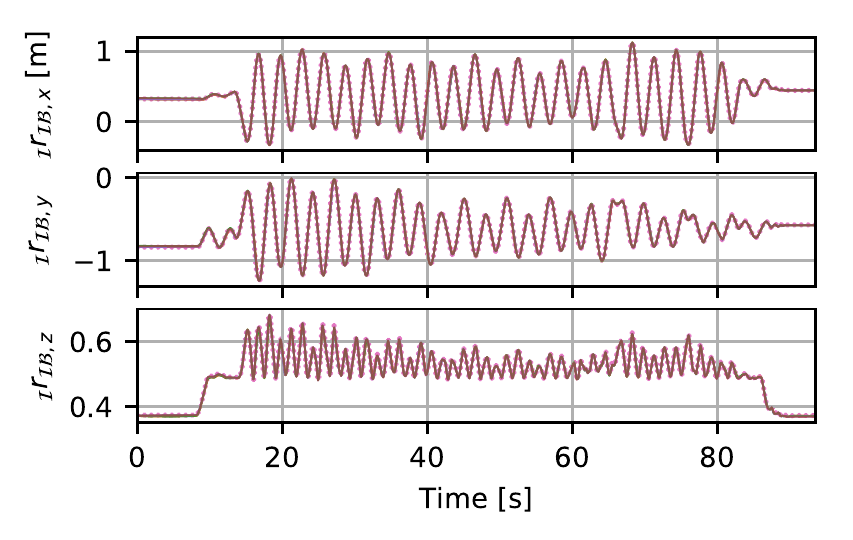}
        \caption{The experiment trajectory.}
        \label{fig:gt_pos}
    \end{subfigure}
    \begin{subfigure}{.49\textwidth}
        \centering
        \includegraphics[width=.99\linewidth]{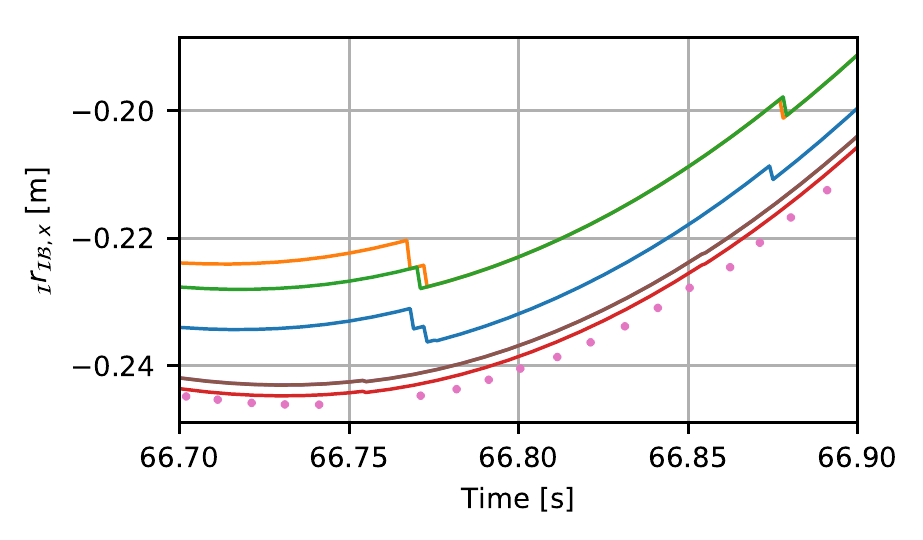}
        \caption{Detailed position estimate.}
        \label{fig:gt_pos_detail}
    \end{subfigure}
    \begin{subfigure}{.49\textwidth}
        \centering
        \includegraphics[width=.99\linewidth]{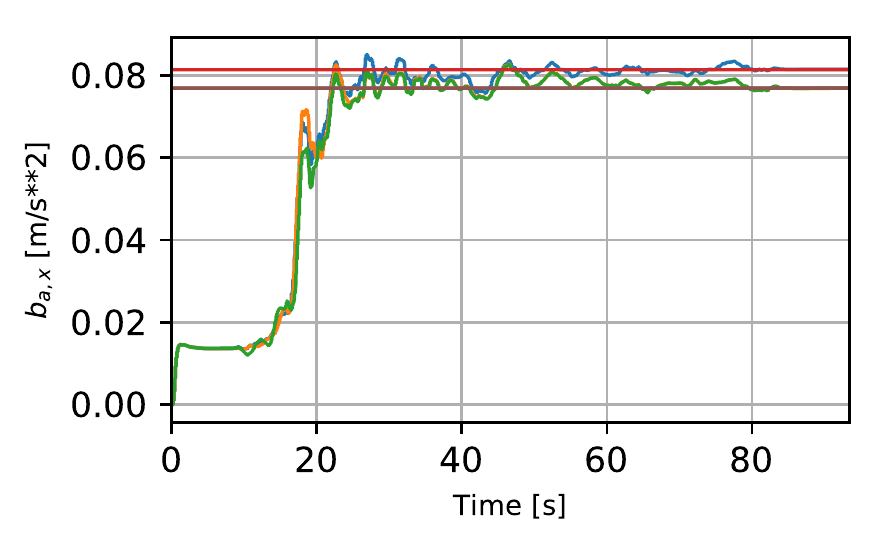}
        \caption{Accelerometer bias estimate.}
        \label{fig:gt_bias}
    \end{subfigure}
    \begin{subfigure}{.49\textwidth}
        \centering
        \includegraphics[width=.99\linewidth]{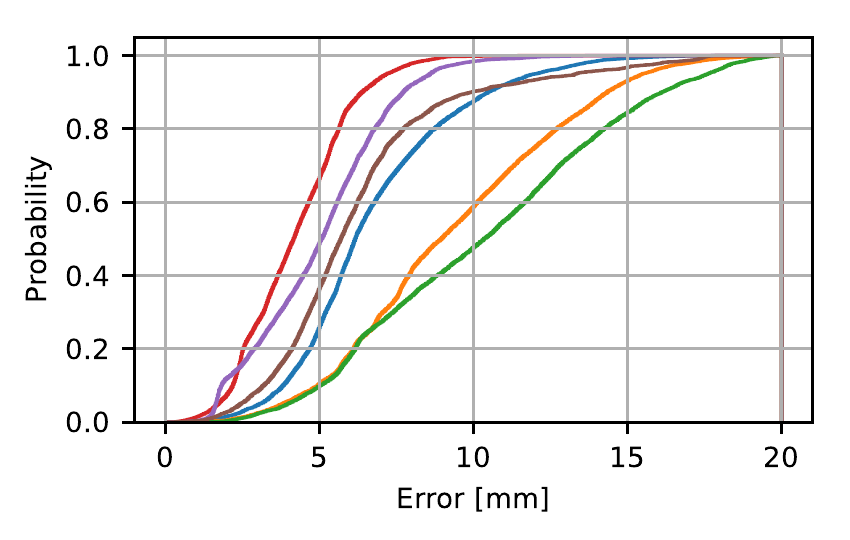}
        \caption{Cumulative absolute error.}
        \label{fig:gt_error}
    \end{subfigure}
    \centering
    \caption{The localization ground truth evaluation using a Vicon motion tracking system. 
    Batch estimation and dual \ac{GNSS} receivers improve the accuracy and precision of the localization algorithm.}
    \label{fig:gt_results}
\end{figure}

The dual \ac{GNSS} receivers are one of the defining features of the factor graph-based localization algorithm.
In this experiment we evaluate their influence to the overall localization accuracy.
In particular, we show the benefits of \ac{GNSS} moving baseline measurements for initialization and during the online sliding window estimate.
Since it is difficult to obtain outdoor ground truth position and orientation measurements that are more accurate than the \ac{RTK} \ac{GNSS} used, we use a Vicon motion tracking system to evaluate the state estimation algorithm~\cite{vicon2014}.
\reffig{fig:gt_setup} shows the experiment setup.

Motion capture markers are attached to the platform to measure the ground truth position and orientation.
The measurements simulate noise-free \ac{GNSS} position and moving baseline measurements at \SI{10}{\Hz} and \SI{5}{\Hz}, respectively.
Due to limited flight space, platform motion is emulated by swinging randomly on a rope for approximately \SI{1.5}{\minute}.
Before the experiment, the \ac{IMU}, \ac{GNSS}, and prior noise models were tuned iteratively, with the goal of obtaining steady bias and \ac{GNSS} phase center estimates.
In post-processing, we run the estimator multiple times with different configurations: with two emulated \ac{GNSS} receiver factors (position + moving baseline), with only one emulated \ac{GNSS} receiver factor (position only), and with one emulated \ac{GNSS} receiver where the initial state is offset by \SI{5}{\degree} in heading (position only + heading offset).
The last configuration resembles a setting where the platform was equipped with only one \ac{GNSS} receiver and a magnetometer was used for initialization.
Besides the three different configurations we also compare the full batch inference with the online estimator with a \SI{3}{\second} sliding window.

\reffig{fig:gt_pos} plots the complete trajectory over our experiment.
At this scale, the different configurations (online, batch, only position measurements, position and moving baseline measurements or with initial heading offset) cannot be distinguished.
All estimates agree with the ground truth position.
Only the detailed view of the position estimates in \reffig{fig:gt_pos_detail} reveals that the batch estimation is always smoother than the online estimation.
This is expected, since the batch optimizer incorporates all measurements, future and past, while the online estimator only has knowledge of past measurements.
The jumps in the online estimation occur because of modeling mismatch between the propagated \ac{IMU} measurements and the emulated \ac{GNSS} measurements, e.g., erroneous \ac{GNSS} phase center calibration, biases, noise models, and time delay. 
Note that these jumps are in the order of millimeters.

Evaluating the error (\reffig{fig:gt_error}) shows that the batch estimation incorporating both position and moving baseline measurements achieves the best \ac{RMSE}.
\reftab{tab:rmse} is sorted by ascending \ac{RMSE} and shows that batch estimation significantly improves estimation accuracy.
One reason is that the online estimator is subject to the availability of computational resources and thus has to rely on \ac{IMU} integration for longer periods before the factor graph is re-evaluated.
Another reason is, again, the additional information available to the batch optimizer.
For example, the batch optimization has a steady estimate of the \ac{IMU} acceleration bias over the whole trajectory length exemplified in \reffig{fig:gt_bias}.
A converged bias estimate has a positive influence on the estimation accuracy.
An incorrect acceleration bias estimate leads to an attitude estimation error, since the gravity vector is observed incorrectly.
This also stresses the importance of (1) finding well-bounded initial estimates for the \ac{GNSS}'s extrinsic calibration and (2) fixing the gyroscope biases before startup to obtain precise results.
\refequ{eq:pos} and \refequ{eq:att} reveal that the dynamics of the antenna positions, accelerometer and gyroscope bias are strongly coupled.
Estimating all of them simultaneously without good initial guesses would require complex platform movements and extremely accurate modeling assumptions.

The \ac{RMSE} analysis of \reffig{fig:gt_error} and \reftab{tab:rmse} also shows that incorporating moving baseline measurements principally improves the accuracy, even when using it only for initialization (position only).
However, it is remarkable that the batch estimate with only position measurements is almost as good as with additional moving baseline measurements.
On the contrary, the online estimation with moving baseline measurements is significantly more accurate than the online estimate with only position measurements.
We suppose that this is due to the influence of the position measurement on the orientation estimate stated in \refequ{eq:pos}.
The orientation is observable under certain motions due to the lever arm between \ac{IMU} and \ac{GNSS} antenna.
The batch estimation with only position measurements finds a good orientation estimate over all the estimates, while the online estimation first needs to converge.
In the case where the orientation is directly measured through the moving baseline measurement, the online estimation can correct the estimate immediately. 

Note that this motion tracker experiment does not allow us to draw conclusions on the accuracy and precision of the localization estimate with actual \ac{GNSS} measurements.
Both simulated \ac{GNSS} measurements and the ground truth have been derived from the same Vicon measurements.
This generally biases the estimation towards the ground truth and correlates the position and moving baseline measurements.
Additionally, no noise has been added to the simulated measurements, which is not the case for true \ac{GNSS} measurements.
On the other hand, it can be expected that the actual \ac{GNSS} timing is better than the Vicon timing.
In addition, the \ac{GNSS} phase center position can probably be determined more accurately than the Vicon marker positions with respect to the \ac{IMU}.
Nevertheless, this experiment still lets us draw the following conclusions:
\begin{itemize}
    \item Batch processing always improves the pose estimation and should be used for mapping purposes;
    \item Online pose estimation benefits significantly from \ac{GNSS} moving baseline measurements;
    \item Initializing the estimator as accurately as possible improves overall accuracy. This includes orientation initialization from \ac{GNSS} moving baseline measurements but also finding well-constrained initial parameters for the gyroscope biases or \ac{GNSS} antenna locations.
\end{itemize}

\subsection{Radar Calibration Target Detection}
\label{sec:radar_calibration}
\begin{figure*}
\begin{subfigure}{.48\textwidth}
\centering
\includegraphics{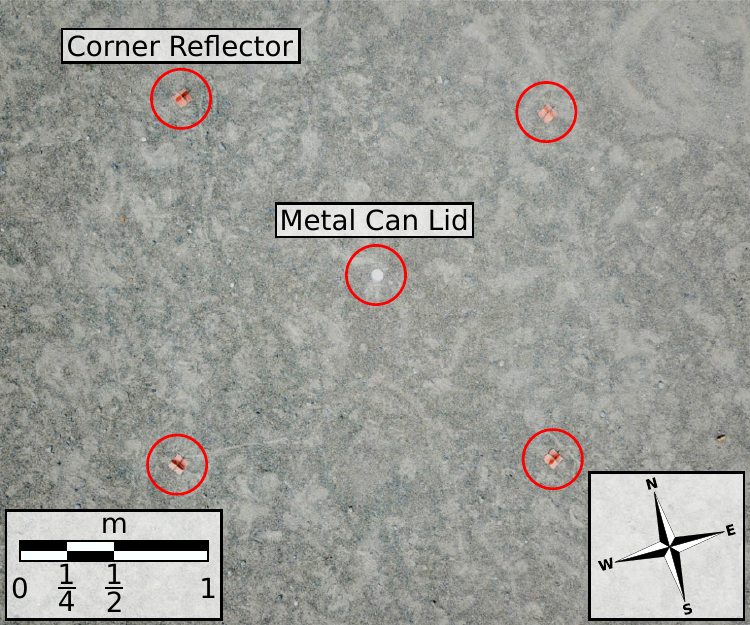}
\caption{Target placement.}
\label{fig:radar_target_setup}
\end{subfigure}
\begin{subfigure}{.51\textwidth}
\centering
\includegraphics{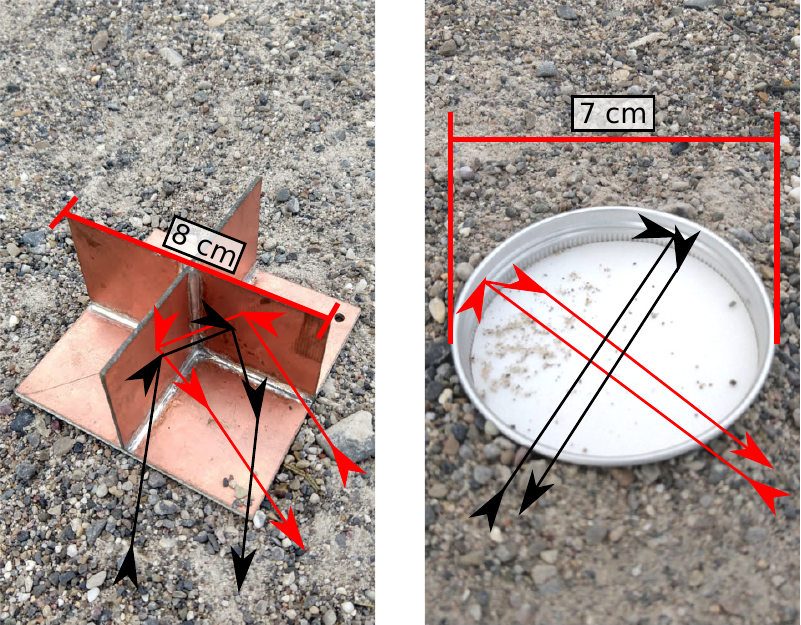}
\caption{Left: Corner reflector, right: metal can lid.}
\label{fig:radar_targets}
\end{subfigure}
\caption{We placed radar calibration targets in a distinctive pattern on the ground. This allows radar back projection independent of the ground surface reflection and soil properties and lets us draw conclusions about the radar antenna positioning performance. (b) The corner reflectors and metal can lid produce distinctive patterns in the radar image.}
\label{fig:calibration}
\end{figure*}

To evaluate the quality of the radar imaging and thus the quality of the overall \reviewchanges{navigation and} radar antenna localization, we performed a well-controlled radar calibration target imaging experiment.
This experiment allows excluding environmental effects, i.e., the necessity for an exact \ac{DSM}, knowledge of the ground surface refraction and soil properties.
\reffig{fig:radar_target_setup} shows a photo of the setup.
Four corner reflectors were placed on the ground in a square with approximately \SI{2}{\metre} distance.
In the center of the square was a metal can lid.
The targets are expected to reflect the radar waves with certain characteristics.
The corner reflectors, as depicted in \reffig{fig:radar_targets}, should show four distinctive features along the approximate compass directions in the resulting radar image.
The metal can lid should show a point symmetrical reflection.

\begin{figure*}
\begin{subfigure}{.99\textwidth}
\centering
\includegraphics{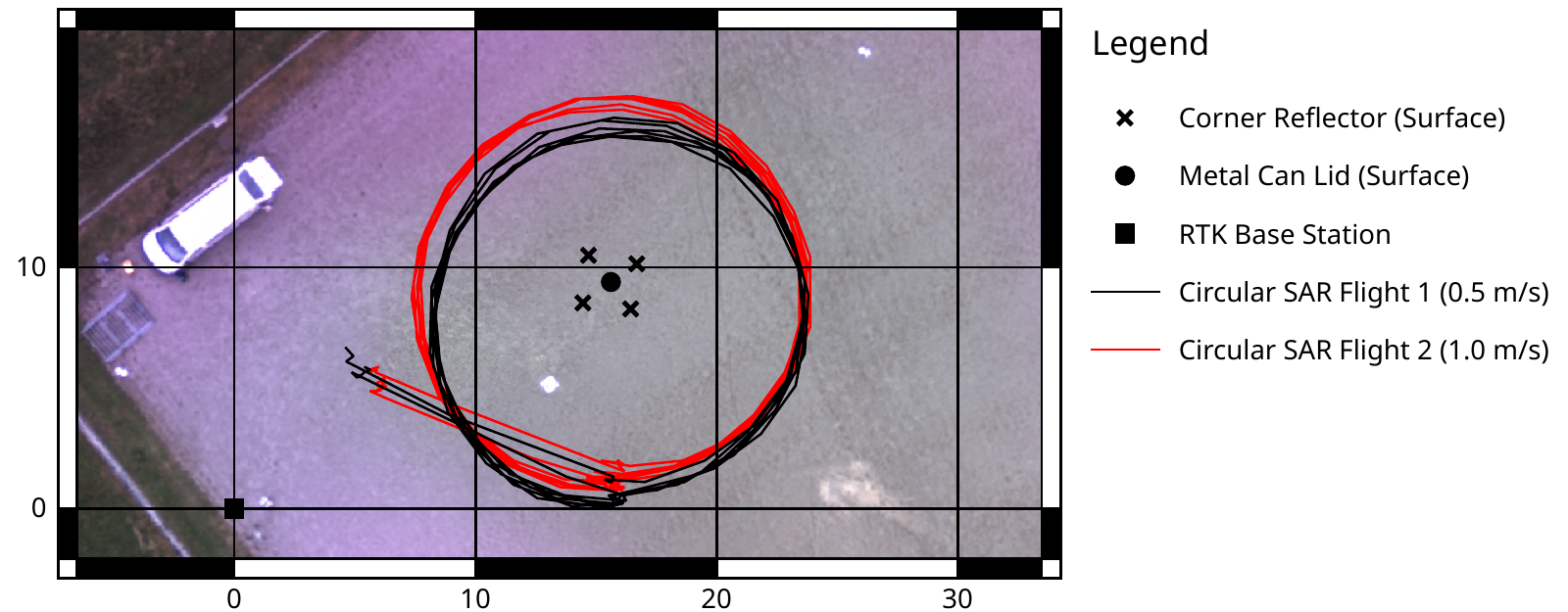}
\caption{The platform transitions to the circle start at \SI{45}{\metre} \ac{AGL} and performs $6$ circles from \SIrange{2}{4}{\metre} \ac{AGL}.}
\label{fig:sst_hoengg}
\end{subfigure} \\
\begin{subfigure}{.5\textwidth}
\centering
\includegraphics{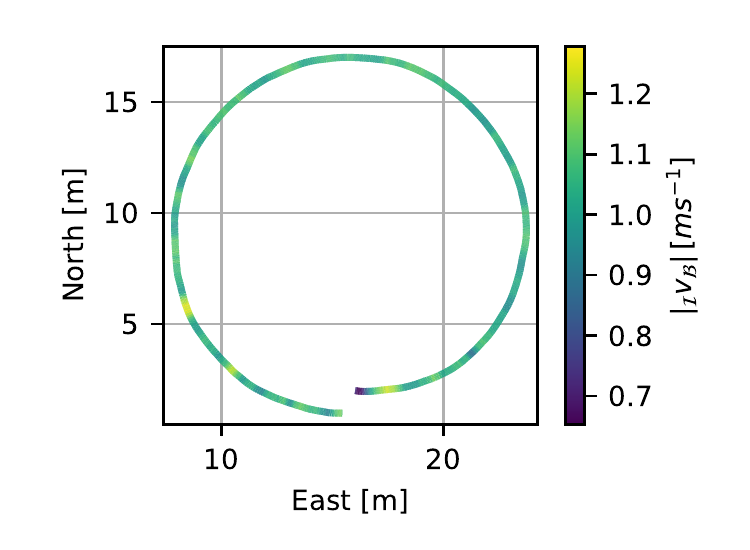}
\caption{\reviewchanges{Masked circular trajectory at \SI{2}{\metre} \ac{AGL} and \SI{1}{\metre\per\second}.}}
\label{fig:velocity_circle}
\end{subfigure}
\begin{subfigure}{.5\textwidth}
\centering
\includegraphics{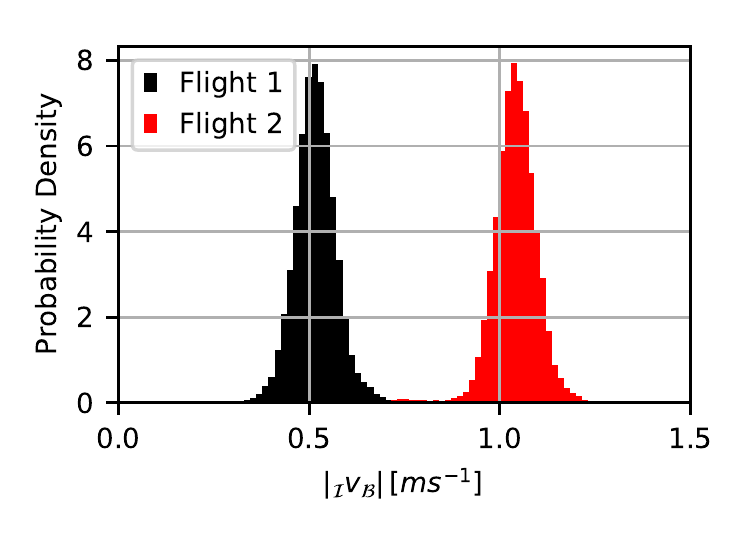}
\caption{\reviewchanges{Velocity distribution of all circular trajectories.}}
\label{fig:velocity_hist}
\end{subfigure}
\caption{Radar experiment with known surface targets. \reviewchanges{The dedicated polynomial circular trajectories enforce constant velocity and thus uniform radar sampling. This holds for slow and fast flights.}}
\end{figure*}

We used the \ac{RTK} survey station to measure the targets' ground truth position and the imaging plane.
The target positions are visualized in the georeferenced map in \reffig{fig:sst_hoengg}.
The platform flew \reviewchanges{two missions with} six circles \reviewchanges{each} with \SI{15}{\metre} diameter centered around the targets.
The platform velocity was \reviewchanges{set to \SI{0.5}{\metre\per\second} in the first flight and} \SI{1}{\metre\per\second} \reviewchanges{in the second flight.} 
The flight altitude increased from \SIrange{2}{4}{\metre} \ac{AGL}.
\reviewchanges{To generate constant velocity circular trajectories, the platform thrust was constrained to \SI[separate-uncertainty=true,multi-part-units=single]{9.81(100)}{\metre\per\second\squared}, the maximum roll and pitch rate to \SI{\pi/12}{\radian\per\second}, and the maximum yaw acceleration to \SI{\pi/2}{\radian\per\second\squared}.
Consequently, the automatically generated acceleration and deceleration segments of the polynomial trajectory (see 
\reffig{fig:circle_acceleration}) were sufficient to reach cruising speed at the circle entry and come to a stop after finishing the circle.
\reffig{fig:velocity_circle} shows the velocity profile of one of the six circles at \SI{1}{\metre\per\second}.}

\reviewchanges{
To generate the radar image for each mission, the radar measurements were masked along the constant velocity circles based on the segment time stamps.
As described in \refsec{sec:trajectory_generation}, we know at the planning stage how long it takes the \ac{MAV} to complete any part of the trajectory.
This leads to a constant velocity profile and thus a uniform sampling rate across all six circles.
\reffig{fig:velocity_hist} shows that this holds both for slow and fast trajectories. 
The only additional radar measurement filtering step was to exclude saturated radar measurements and the \ac{RC} frequency band.
For each circle the coherent radar image was formed and for each individual flight all six images were added coherently to form the final products.
}

\begin{figure*}
\begin{subfigure}{.99\textwidth}
\centering
\includegraphics{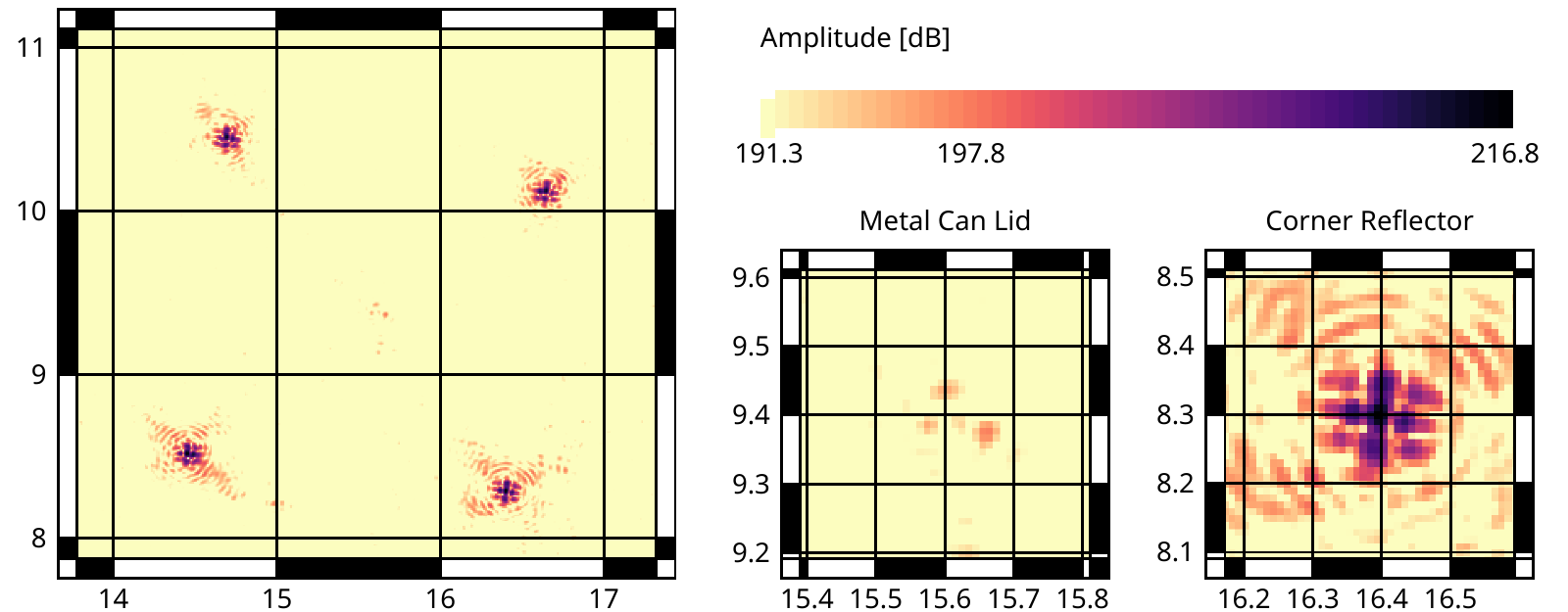}
\caption{\reviewchanges{Radar image of flight 1 (\SI{0.5}{\metre\per\second}) evaluated at ground surface level with our localization solution and horizontally polarized antennas. The amplitude is cutoff at the top \SI{2}{\percent}.}}
\label{fig:sar_hoengg_05}
\end{subfigure}
\begin{subfigure}{.99\textwidth}
\centering
\includegraphics{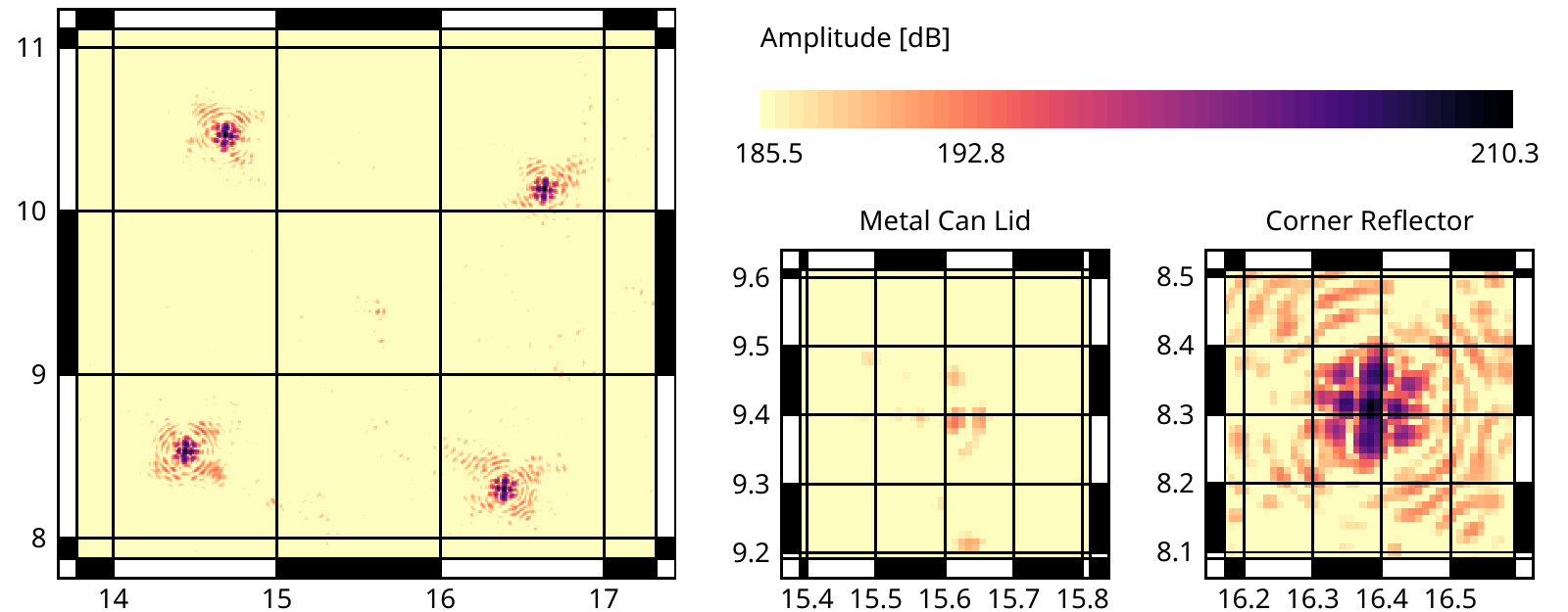}
\caption{Radar image \reviewchanges{of flight 2 (\SI{1.0}{\metre\per\second})} evaluated at ground surface level with our localization solution and horizontally polarized antennas. The amplitude is cutoff at the top \SI{2}{\percent}.}
\label{fig:sar_hoengg}
\end{subfigure}
\caption{\reviewchanges{Both flights deliver comparable results with a greater amplitude in the slow flight due to approximately twice the number of radar measurements. The similarity of the images shows system repeatability.} The metal can lid and the four corner reflectors are detectable. The details show a circular response for the can. The corner reflectors create a well-defined signature with visible corners approximately along the compass directions and surrounding antenna side lobes.}
\label{fig:calibration_results}
\end{figure*}

\reviewchanges{\reffig{fig:calibration_results}} shows the \reviewchanges{two} resulting radar image\reviewchanges{s} at ground surface level with \SI{1}{\centi\metre} cell resolution.
\reviewchanges{Despite the two flight paths in \reffig{fig:sst_hoengg} being offset horizontally by approximately \SI{1}{\metre} due to the DJI \ac{GNSS} position control (see \refsec{sec:controller}) and different reference velocities, both images in \reffig{fig:sar_hoengg_05}~and~\ref{fig:sar_hoengg} are similar.
This highlights the capability of our system to perform repeatable experiments.
The main difference is a greater amplitude in the slow flight due to almost twice as many radar measurements back projected into the image.
Furthermore, the corner reflector signature, seen in the bottom right of both figures, shifts north-west by approximately \SI{1.5}{\pixel} (\SI{1.5}{\centi\metre}).
This systematic shift may be caused by radar imaging artefacts which emerge from the shifted flight paths and the geometry of the corner reflectors~\cite{moreira2013tutorial}.}
\reviewchanges{Still, o}ne can clearly identify the four radar reflector locations and the location of the metal can lid \reviewchanges{in both images}.
A detailed view on the individual targets reveals a point symmetrical response of the metal can lid and the geometric pattern of the corner reflectors.
In particular, the four corners stand out pointing towards north, east, south and west.  
Furthermore, the center of the target has a strong response and the four corner edges of the copper base plate stick out. This demonstrates the sensitivity of the proposed system \reviewchanges{as well as the repeated localization precision to generate focused radar images}.
Additionally, periodic circular bands show up around the target, these indicate the side lobes.
\reviewchanges{In the remainder of this section we will investigate only the second flight in detail.}

\begin{figure*}
\begin{subfigure}{.31\textwidth}
\centering
\includegraphics{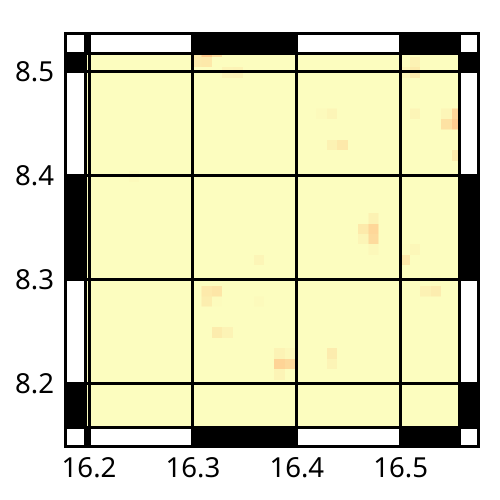}
\caption{DJI only.}
\label{fig:DJI_only_good_ant}
\end{subfigure} 
\begin{subfigure}{.31\textwidth}
\centering
\includegraphics{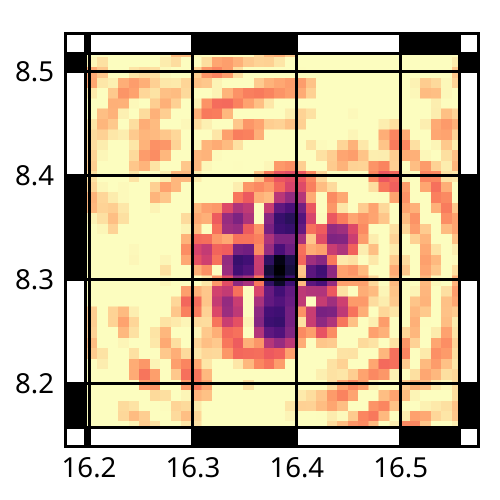}
\caption{DJI + \ac{RTK}.}
\label{fig:DJI_RTK_good_ant}
\end{subfigure} 
\begin{subfigure}{.31\textwidth}
\centering
\includegraphics{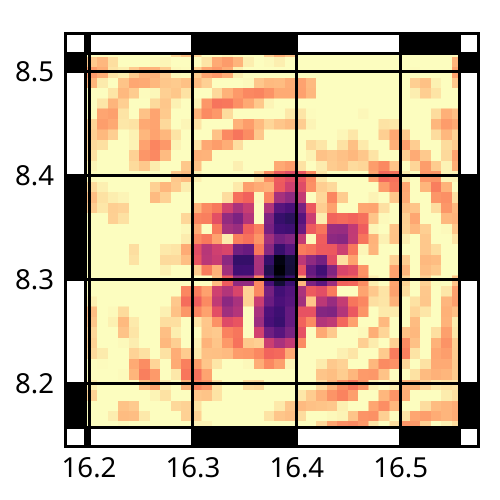}
\caption{Ours.}
\label{fig:ours_good_ant}
\end{subfigure} \\
\begin{subfigure}{.49\textwidth}
\centering
\includegraphics{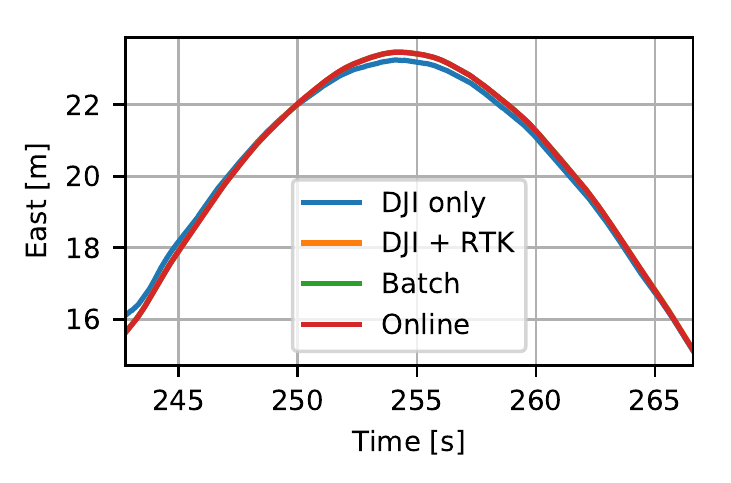}
\caption{Position estimate east.}
\label{fig:hoeng_east}
\end{subfigure} 
\begin{subfigure}{.49\textwidth}
\centering
\includegraphics{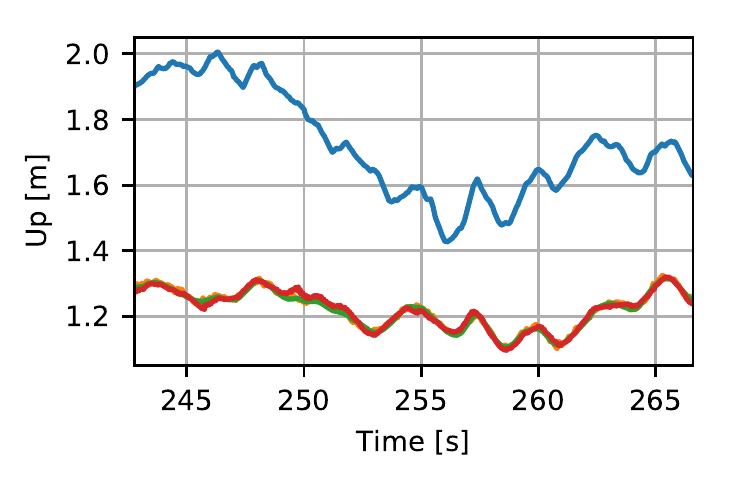}
\caption{Position estimate up.}
\label{fig:hoeng_up}
\end{subfigure} \\
\begin{subfigure}{.49\textwidth}
\centering
\includegraphics{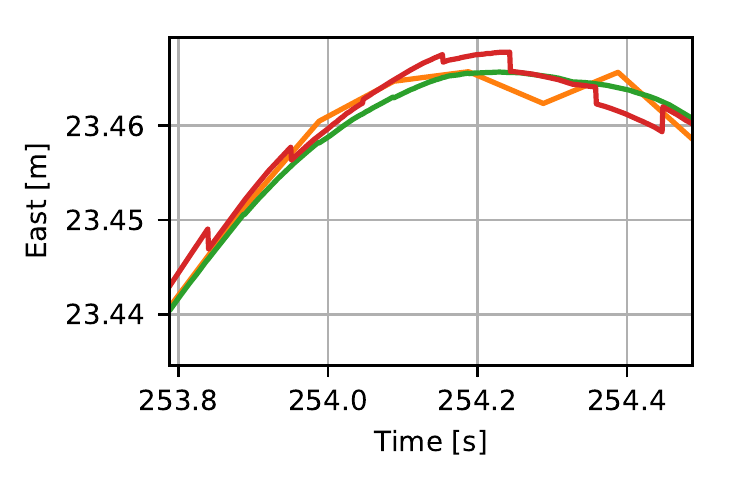}
\caption{\reviewchanges{Position estimate detail east.}}
\label{fig:hoengg_east_detail}
\end{subfigure} 
\begin{subfigure}{.49\textwidth}
\centering
\includegraphics{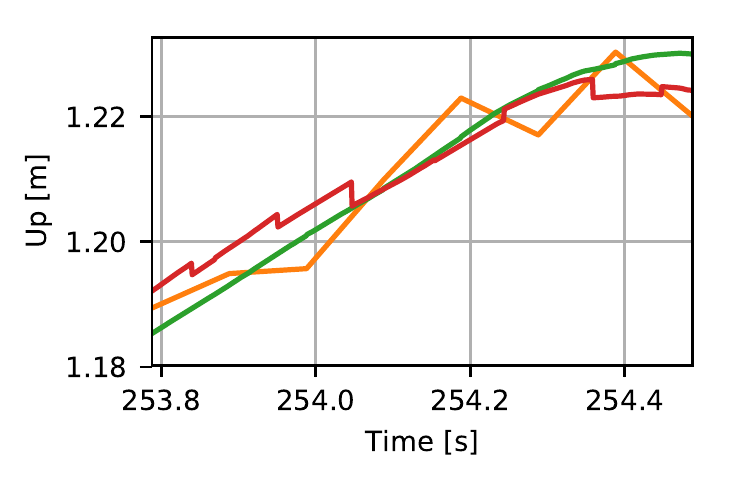}
\caption{\reviewchanges{Position estimate detail up.}}
\label{fig:hoengg_up_detail}
\end{subfigure} 
\centering
\caption{Ablation study analyzing the radar imaging performance of different antenna localization solutions \reviewchanges{in flight 2}. \reviewchanges{Our \ac{GTSAM} batch solution delivers the smoothest and possibly most accurate position estimation result. However, a simple DJI+RTK solution is also sufficient for the radar imaging task if the transformations between radar and \ac{GNSS} antenna are known.} The standalone DJI solution is not sufficient.}
%Note that generating (a) requires a known transformation between the DJI frame and the base frame, while generating (b) requires a known transformation between the \ac{RTK} \ac{GNSS} position receiver and the base frame. On the other hand, our solution is able to directly solve for the necessary transforms (shown in \reffig{fig:kinematic}) as part of the \ac{GTSAM} optimization.}
\label{fig:good_antenna_estimate}
\end{figure*}

In the next step, we performed an ablation study to investigate the specific impact of our \reviewchanges{batch} antenna localization algorithm. 
For comparison, we processed the radar returns using only the localization solution provided out-of-the-box by DJI (DJI only) and the solution interpolated from the raw DJI attitude outputs and the raw \ac{RTK} \ac{GNSS} positioning (DJI+RTK). 
We highlight the differences in the resulting radar images by zooming in on a single corner reflector in \reffig{fig:good_antenna_estimate}.

The differences between the three images are most prominent between the DJI only solution and the other two which use \ac{RTK} \ac{GNSS} measurements. 
The radar image in \reffig{fig:DJI_only_good_ant} does not show a focused image in the expected target position.
The pure DJI position estimation accuracy is not sufficient to run the back projection algorithm.
\reffig{fig:hoeng_east} shows that the DJI only solution is offset by about \SI{20}{\centi\metre} from the RTK solutions in the horizontal direction.
The altitude estimation in \reffig{fig:hoeng_up} shows an even more severe drifting offset of up to \SI{70}{\centi\metre}.

On the other hand, both the DJI+RTK and our localization solution provide comparable radar image quality.
\reviewchanges{This is surprising because \reffig{fig:hoengg_east_detail} and~\ref{fig:hoengg_up_detail} show that our batch estimation solution is smoother and possibly more accurate than the DJI+RTK solution.
This indicates, that the downstream task of \ac{GPSAR} imaging is not sensitive enough to resolve a few millimeters difference in localization.
The experiment also demonstrates that the DJI+RTK solution is an attractive alternative to integrate \ac{GPSAR} on \iac{MAV} in situations where our level of accuracy as well as future sensor integration and self-calibration is not required. 
To obtain the pose of the individual antennas with the DJI+RTK solution $\hat{T}_{\mathcal{I}\mathcal{S}_i}$ we solve the following kinematic chain:  
\begin{align}
    \hat{T}_{\mathcal{I}\mathcal{S}_i} &= \tilde{T}_{\mathcal{I}\mathcal{D}}~T_{\mathcal{D}\mathcal{S}_i} = \tilde{T}_{\mathcal{I}\mathcal{D}}~T_{\mathcal{D}\mathcal{B}}~T_{\mathcal{B}\mathcal{S}_i},
\end{align}
where $\tilde{T}_{\mathcal{I}\mathcal{D}}$ is composed of the measured \SI{10}{\hertz} RTK position and interpolated DJI orientation and $T_{\mathcal{D}\mathcal{S}_i}$ describes the rigid transform from the RTK position antenna to the respective radar antenna.
Note that in our case the latter is obtained from a series of calibration efforts.
$T_{\mathcal{D}\mathcal{B}}$ is composed of the known rotation from the body frame to DJI frame $R_{\mathcal{BD}}$ and our self-calibrated RTK antenna position ${}_{\mathcal{B}}r_{\mathcal{B}\mathcal{P}}$.
$T_{\mathcal{B}\mathcal{S}_i}$ is known from \ac{CAD}.
Obtaining this transformation accurately is critical~\cite{yanghuan2011lever}.}
%However, it must be noted that both the DJI only and the DJI+RTK localization solutions require known transformations to the base frame. 
%These can be obtained manually, however, as we showed in \reffig{fig:ant_estimate}, it can be difficult to achieve precise measurements. 
%In contrast, our method is able to directly solve for these transformations within the \ac{GTSAM} optimization routine. 
%Indeed, the results in \reffig{fig:DJI_only_good_ant} and \reffig{fig:DJI_RTK_good_ant} both make use of transforms computed from our batch optimization.

\begin{figure*}
\begin{subfigure}{.31\textwidth}
\centering
\includegraphics[height=2in]{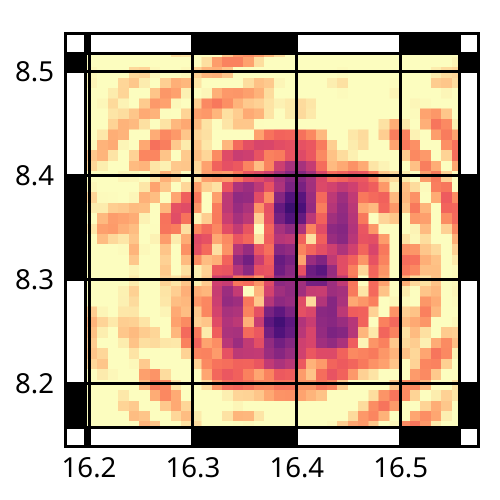}
\caption{DJI + \ac{RTK}.}
\label{fig:hoengg_raw_offset_detail}
\end{subfigure} 
\begin{subfigure}{.31\textwidth}
\centering
\includegraphics[height=2in]{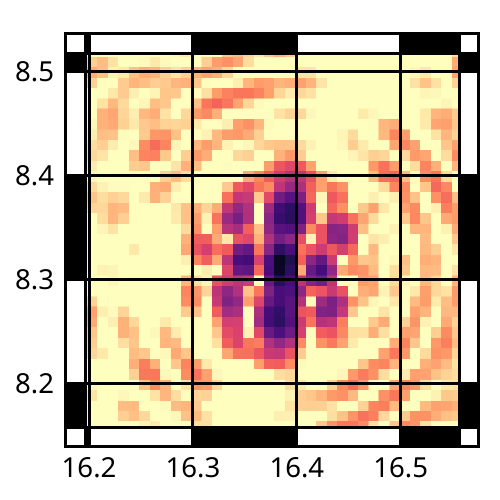}
\caption{Ours.}
\label{fig:hoengg_fusion_offset_detail}
\end{subfigure} 
\centering
\caption{Analysis of system sensitivity to \ac{RTK} \ac{GNSS} receiver position offsets \reviewchanges{in flight 2}. Both radar images are generated given a \SI{3}{\centi\metre} offset between the expected and true \ac{GNSS} receiver position. Our solution is able to perform self-calibration to account for this error, recovering a sharp radar image of similar quality to that shown in \reffig{fig:ours_good_ant}.}
\label{fig:bad_antenna_estimate}
\end{figure*}

\reviewchanges{To stress the importance of an accurate \ac{GNSS} to radar antenna calibration}, we investigated the effect of an inaccurate \ac{GNSS} receiver to base frame transform. 
In this case, we applied a \SI{3}{\centi\metre} offset in forward, left and up direction to the receiver antenna centers ${}_{\mathcal{B}}r_{\mathcal{B}\mathcal{P}}$ and ${}_{\mathcal{B}}r_{\mathcal{B}\mathcal{M}}$ and re-generated the resulting radar images for the interpolated DJI+RTK localization solution and our localization solution. \reffig{fig:bad_antenna_estimate} shows a sharp contrast between the image quality from the two solutions. 
With only the interpolated position and attitude measurements, the DJI+RTK solution has no way to identify and correct for the inaccurate receiver position, resulting in a blurry radar image. 
Our solution offers self-calibration in the \ac{GTSAM} fusion, which is shown for this particular flight in \reffig{fig:ant_estimate}, and is able to recover a sharp radar image despite the initially poor \ac{GNSS} receiver position estimate.

\reviewchanges{Finally, we discuss the online and batch solution of our localization algorithm in flight.
\reffig{fig:hoengg_east_detail} and~\ref{fig:hoengg_up_detail} show that the batch solution smoothes the online solution significantly, improving localization precision.
This is consistent with the Vicon motion tracking experiments in \refsec{sec:ground_truth}.
However, compared to the rope swinging experiment in~\reffig{fig:gt_pos_detail}, the jumps in the online estimate between \ac{IMU} integration and \ac{GNSS} update are greater in flight.
This indicates that additional effects such as frame vibrations and \ac{RTK} \ac{GNSS} noise influence the estimate.}

\reviewchanges{Summarized, this experiment lead to the following findings:
\begin{itemize}
    \item Polynomial trajectory generation is an adequate technique to generate dedicated \ac{GPSAR} trajectories that enforce uniform sampling. 
    \item Our self-contained positioning solution enables repeatable, high-resolution radar imaging. 
    \item Interpolating off-the-shelf \ac{RTK} \ac{GNSS} and DJI autopilot attitude is also a viable positioning solution for low-frequency coherent radar imaging. 
    \item Calibrating the radar antenna positions accurately with respect to the \ac{GNSS} antennas is critical. Our self-calibrating algorithm helps to determine the \ac{GNSS} antenna positions ${}_{\mathcal{B}}r_{\mathcal{B}\mathcal{P}}$, ${}_{\mathcal{B}}r_{\mathcal{B}\mathcal{M}}$ to generate focused images. 
    \item Batch estimation to improve localization precision has been confirmed in flight. Sensor noise such as frame vibrations and \ac{GNSS} measurement uncertainties may alter the estimation quality.
\end{itemize}}

%Summarized, the radar calibration target study implies an important result. The static calibrations of the \ac{GNSS} and radar antenna phase centers  ${}_{\mathcal{B}}r_{\mathcal{B}\mathcal{P}}$, ${}_{\mathcal{B}}r_{\mathcal{B}\mathcal{M}}$, ${}_{\mathcal{B}}T_{\mathcal{B}{\mathcal{S}_i}}$ have a substantial influence on the radar back projection quality and must be determined carefully. 
%Nevertheless, our sensor synchronization setup along with our batch optimization formulation is able to explicitly correct for these errors, recovering a sharp radar image despite an offset in the initial static calibration.

\subsection{Buried Object Detection} \label{sec:buried_objects}
\begin{figure*}
\centering
\includegraphics{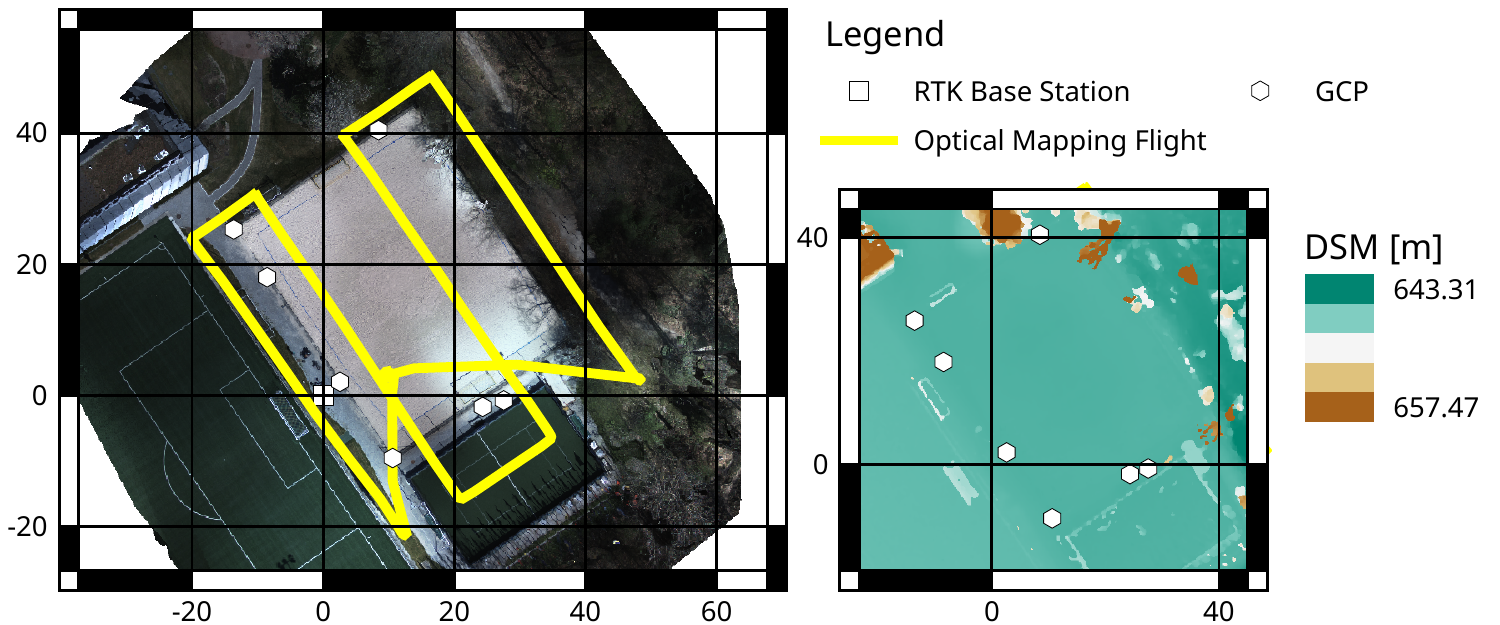}
\centering
\caption{A georeferenced orthomosaic and \ac{DSM} with \SI{2.5}{\centi\metre} \ac{GSD} created by our platform. The maps are used to plan the circular \ac{GPSAR} trajectory and compute the ground surface transition in the radar imaging algorithm.}
\label{fig:dsm}
\end{figure*}

Finally, we validated the whole system for buried object detection.
The experiment environment is a clean beach soccer field where we buried the same metal can lids as in the previous experiment. Four outer targets forming a square were buried at a depth of around \SIrange{2}{5}{\centi\metre} and a center target was buried at lower depth of around \SI{19}{cm}.
In a first flight we collected georeferenced images using our system at \SI{45}{\metre} altitude.
We created the orthomosaic and \ac{DSM} in \reffig{fig:dsm} with a \ac{GSD} of \SI{2.5}{\centi\metre} of the environment using off-the-shelf photogrammetry software~\cite{pix4d2021}.
The maps are georeferenced using a set of \acp{GCP} placed outside of the target area and surveyed with our equipment.

\begin{figure*}
\begin{subfigure}{.99\textwidth}
\centering
\includegraphics[width=0.99\linewidth]{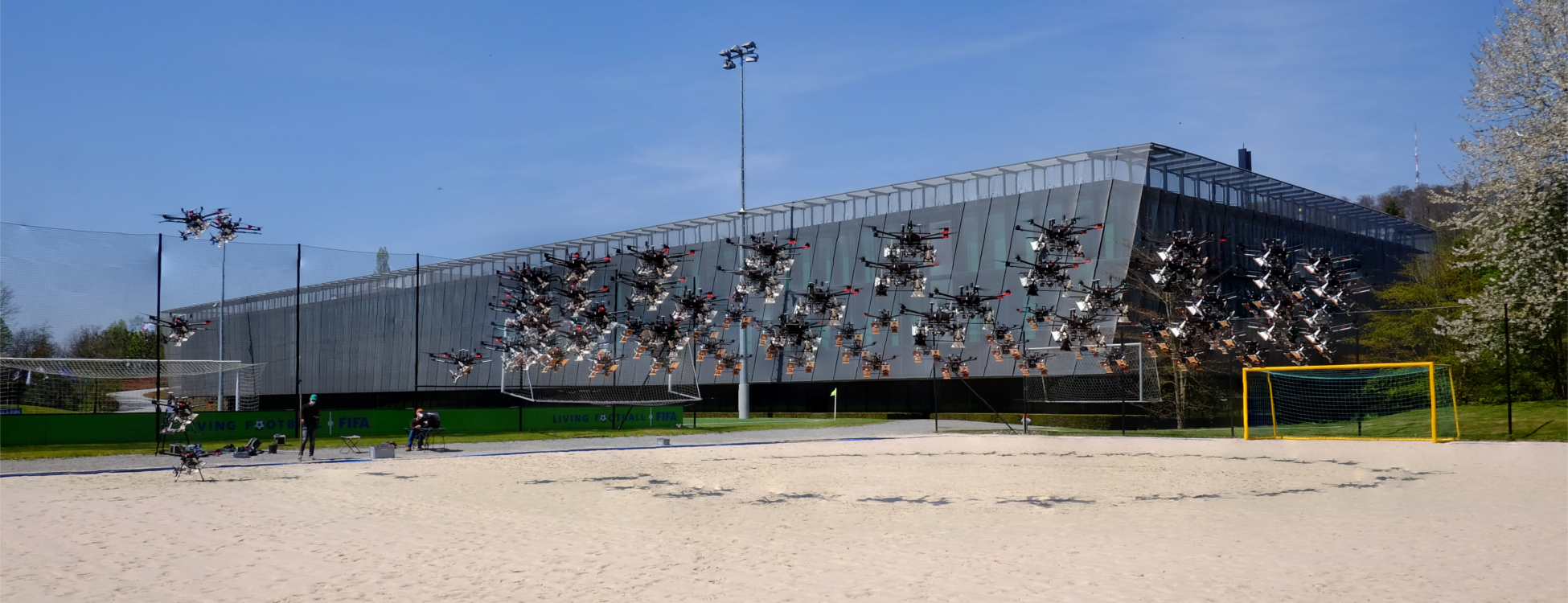}
\caption{Time lapse of an automated \SI{6}{\minute} circular \ac{GPSAR} trajectory with increasing altitudes.}
\end{subfigure}  \\
\begin{subfigure}{.99\textwidth}
\centering
\includegraphics{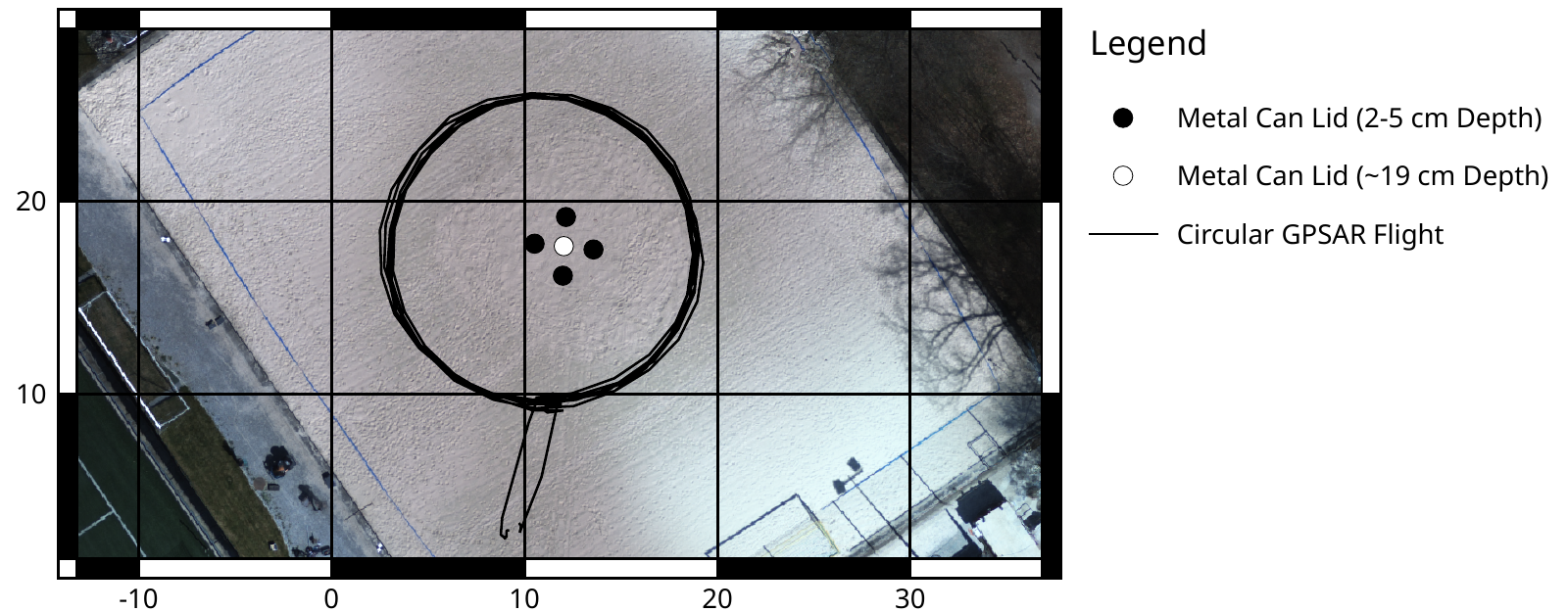}
\caption{The buried target positions and executed radar trajectory.}
\end{subfigure} 
\caption{Buried object detection experiment. The platform scanned a \SI{7 x 7}{\metre} patch without the operators entering the field.}
\label{fig:experiment}
\end{figure*}

\begin{figure*}
\begin{subfigure}{.99\textwidth}
\centering
\includegraphics{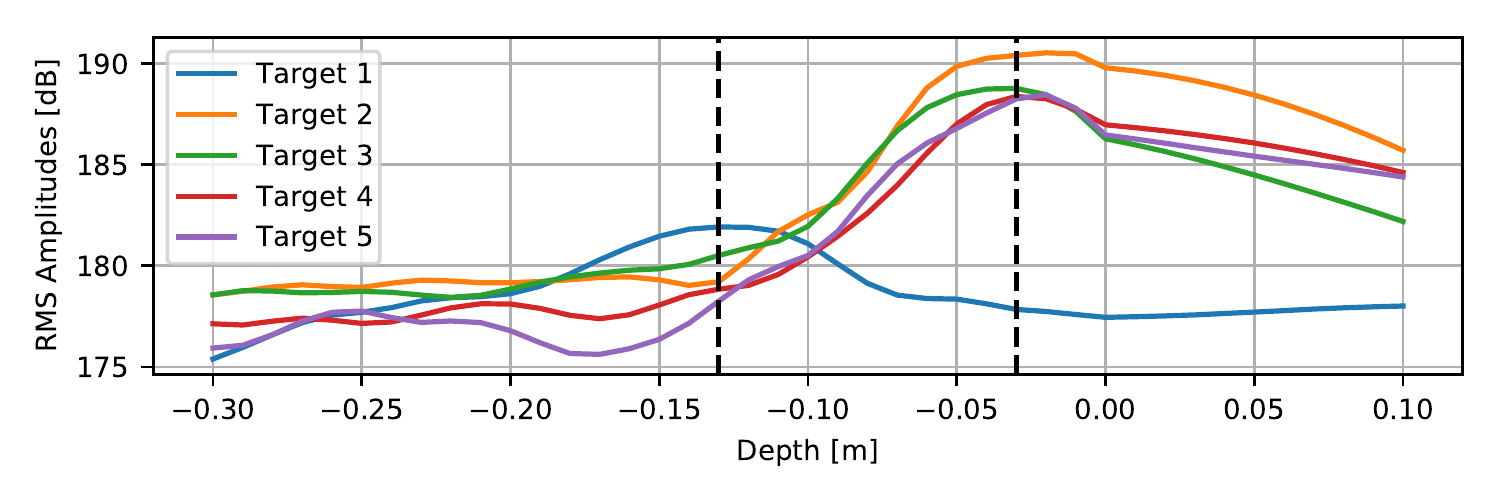}
\caption{Target \ac{RMS} amplitudes in a \SI{20 x 20}{\centi\metre} image patch around the ground truth position \reviewchanges{with our antenna positioning solution}.}
\label{fig:target_responses}
\end{subfigure}  \\
\begin{subfigure}{.99\textwidth}
\centering
\vspace{0.5cm}
\includegraphics{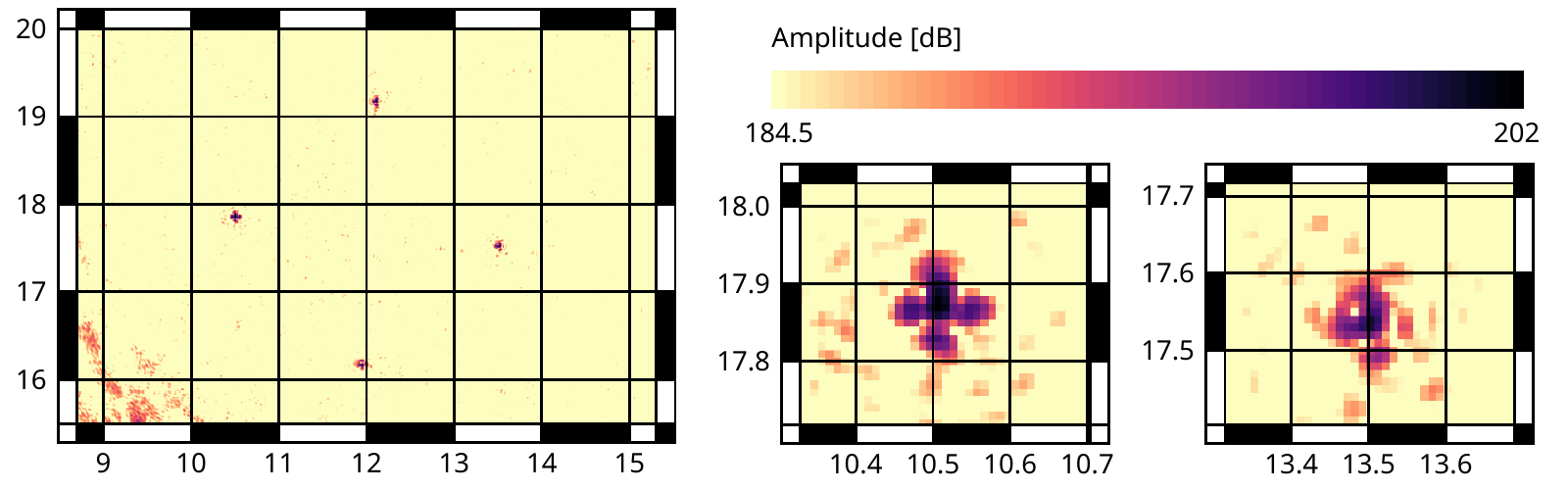}
\caption{\SI{3}{\centi\metre} underground with vertically polarized antennas \reviewchanges{with our positioning solution}.}
\label{fig:fifa_shallow_objects}
\end{subfigure}  \\
\begin{subfigure}{.99\textwidth}
\centering
\vspace{0.5cm}
\includegraphics{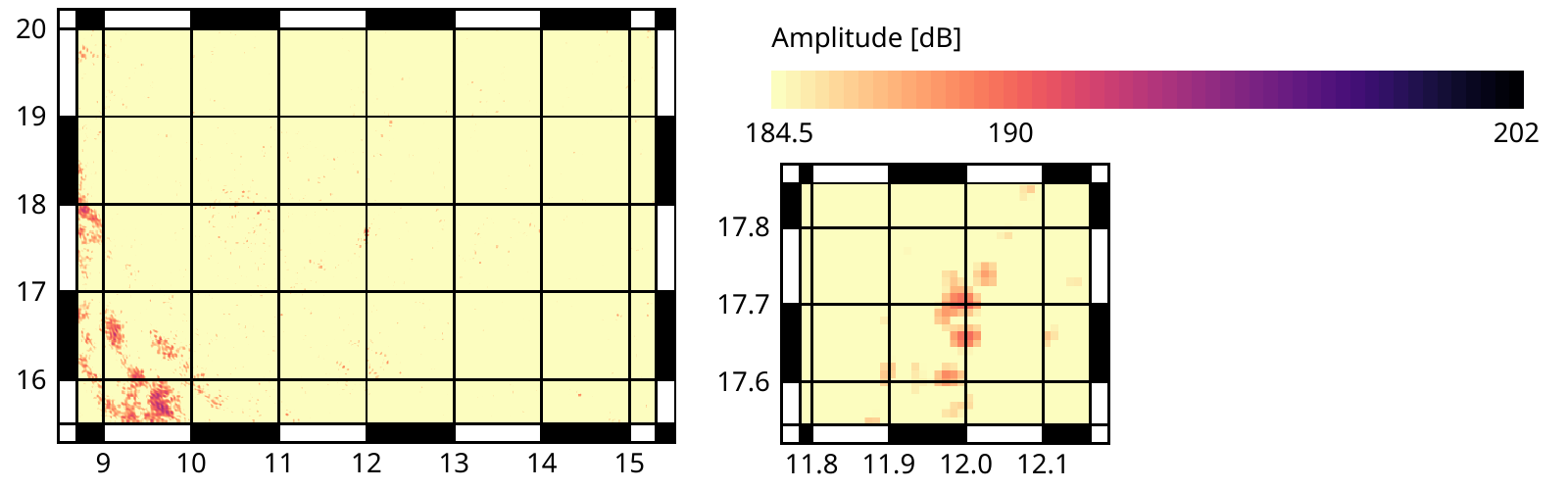}
\caption{\SI{13}{\centi\metre} underground with vertically polarized antennas \reviewchanges{with our positioning solution}.}
\label{fig:fifa_deep_objects}
\end{subfigure}
\caption{Radar images evaluated at different depths \reviewchanges{with our localization solution}. The shallow targets are clearly visible verifying the capability to detect landmines. The deep object is also detectable, however the challenges of deeper subsurface sensing are also highlighted by the attenuation of the radar signal in (c). Additional clutter shows up in the bottom left-hand corner of the image.}
\label{fig:buried_gpsar}
\end{figure*}

In a second flight we performed the actual \ac{GPSAR} survey from safe distance shown in \reffig{fig:experiment}.
The platform flew $6$ circles with \SI{15}{\metre} diameter centered around the target area.
The platform velocity was \reviewchanges{set to} \SI{1}{\metre\per\second} and the flight altitude increased from \SIrange{2}{4}{\metre} \ac{AGL} over approximately \SI{6}{\minute}.
\reviewchanges{The dynamic constraints were identical to the previous experiment, enforcing constant velocity along the circles.}
The collected radar responses\reviewchanges{, trajectory time stamp mask,} and  batch processed radar antenna poses are forwarded to a compute cluster to perform the \reviewchanges{coherent} radar back projection algorithm.
\reviewchanges{The permittivity of the soil was not modified and left at its default value of $\epsilon_r=8$.}
The resulting radar image extends \SI{7 x 7}{\metre} and spans a volume from \SIrange{-0.3}{0.1}{\metre} above surface level with a resolution of \SI{1}{\centi\metre}.
\reffig{fig:buried_gpsar} shows the resulting \ac{GPSAR} image at \SI{3}{\centi\metre} and \SI{13}{\centi\metre} depth as well as the \ac{RMS} radar image cell magnitudes around the ground truth position of the targets.
The metal can lids buried at shallow depth can be detected in the radar image.
This validates that the system presented in this paper can potentially be used to detect landmines.

The deep object shows the strongest response at \SI{13}{\centi\metre} depth.
The offset between the manually measured depth and the peak radar response may be due to several confounding factors that make subsurface sensing especially challenging. These include unmodeled complex effects of soil moisture and heterogeneity on the ground permittivity, digital surface map accuracy as well as errors induced through the manual ground truth surveying process.
\reviewchanges{As described in the experiment setup, we do not calibrate  $\epsilon_r$ to correct the depth.}

\begin{figure*}
\begin{subfigure}{.99\textwidth}
\centering
\includegraphics{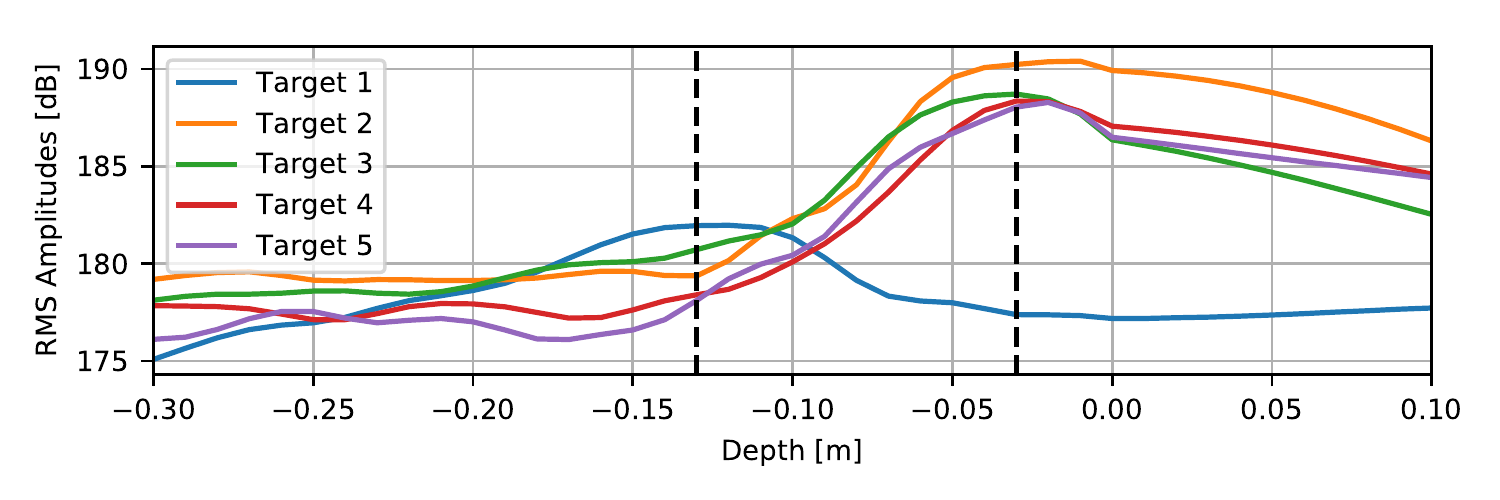}
\caption{\reviewchanges{Target \ac{RMS} amplitudes in a \SI{20 x 20}{\centi\metre} image patch around the ground truth position with the DJI+RTK antenna positioning.}}
\label{fig:target_responses_djirtk}
\end{subfigure}  \\
\begin{subfigure}{.99\textwidth}
\centering
\vspace{0.5cm}
\includegraphics{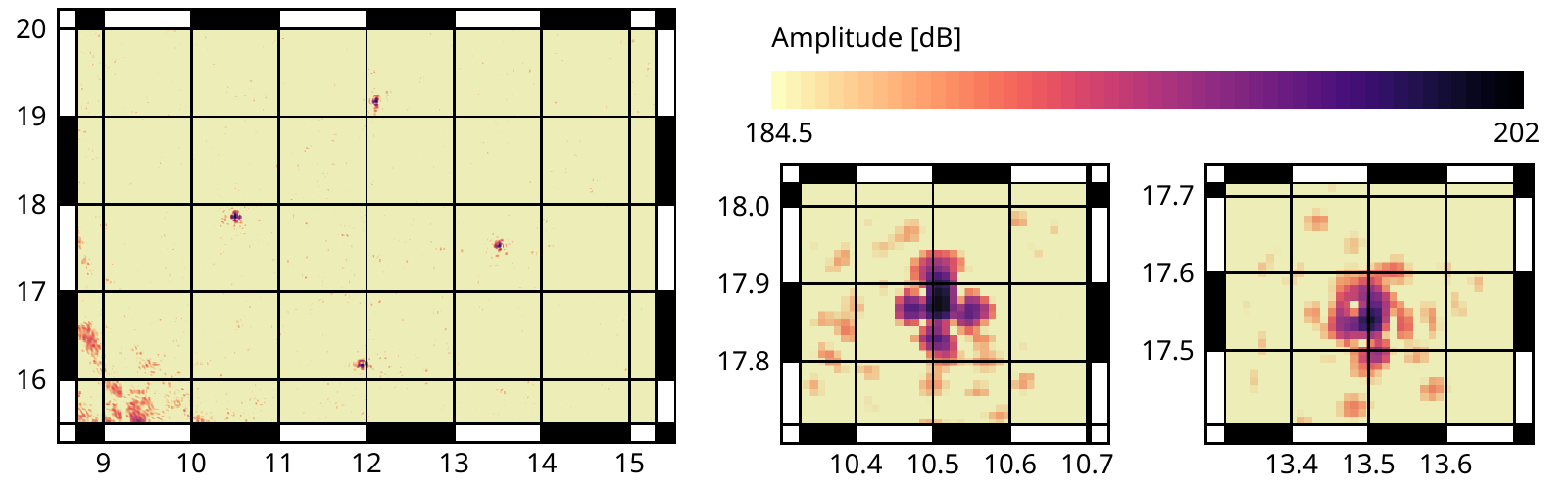}
\caption{\reviewchanges{\SI{3}{\centi\metre} underground with vertically polarized antennas with the DJI+RTK positioning.}}
\label{fig:fifa_shallow_objects_djirtk}
\end{subfigure}  \\
\begin{subfigure}{.99\textwidth}
\centering
\vspace{0.5cm}
\includegraphics{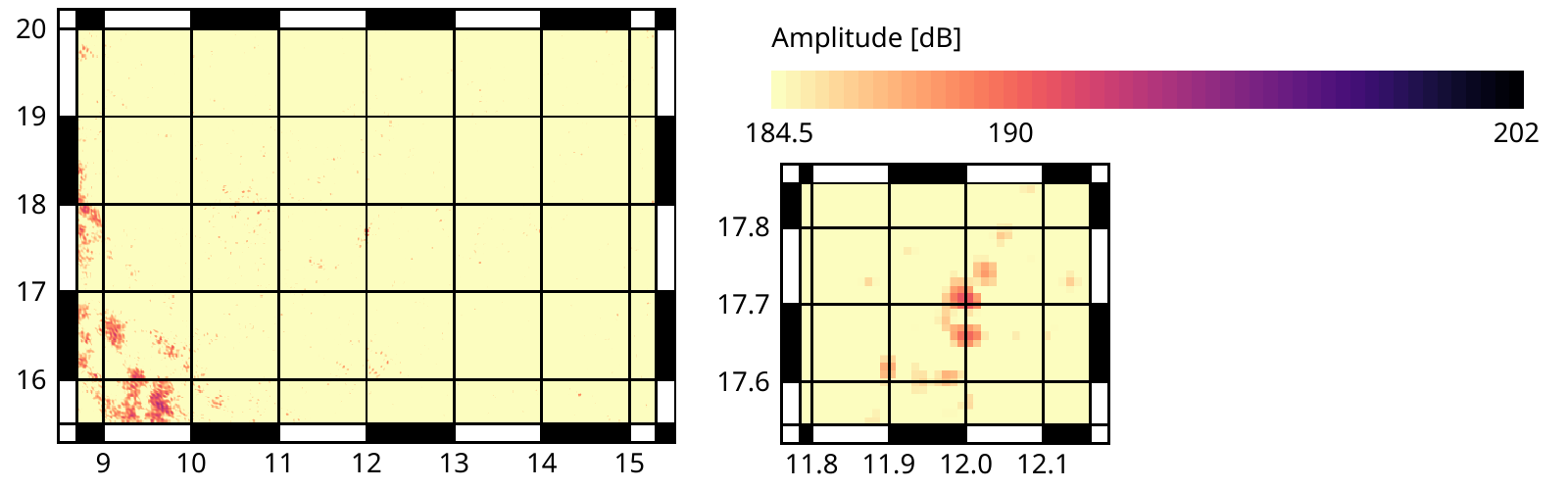}
\caption{\reviewchanges{\SI{13}{\centi\metre} underground with vertically polarized antennas with the DJI+RTK positioning.}}
\label{fig:fifa_deep_objects_djirtk}
\end{subfigure}
\caption{\reviewchanges{Radar images evaluated at different depths with the DJI+RTK antenna positioning solution. The similarity to \reffig{fig:buried_gpsar} confirms that DJI+RTK is a viable solution for \ac{GPSAR} imaging.
Since the images do not show significant differences, localization does not seem to be the bottleneck of \ac{GPSAR} anymore. Other radar processing errors, e.g., millimeter wave soil propagation modelling, seem to outweigh localization errors.}}
\label{fig:buried_gpsar_djirtk}
\end{figure*}

\reviewchanges{To confirm that DJI+RTK presented in the previous section is a viable solution for buried object detection as well, we repeat the imaging process.
The result in \reffig{fig:buried_gpsar_djirtk} resembles the result obtained with our localization solution in \reffig{fig:buried_gpsar}.
This indicates that localization for \ac{GPSAR} is not the bottle neck anymore and other error sources outweigh.
Possible significant error sources in \ac{GPSAR} imaging are clutter responses, modelling the millimeter wave propagation in heterogeneous soil, the frequency dependent radar antenna phase center positions, and errors in the surface model.
Note that a dedicated localization fusion algorithm such as the one presented in this paper may still give advantages over the DJI+RTK solution.
It is extendable to additional sensor modalities in \ac{GNSS} denied environments, it can include self-calibration routines, and it has greater precision as the previous section suggests.
The latter may be relevant to other mapping modalities where greater precision is required, e.g., high-frequency radar, or once the major modelling errors in \ac{GPSAR} are resolved.}

\section{Conclusion} \label{sec:conclusion}
% What did we do
In this paper we presented an airborne \ac{GPSAR} solution capable of detecting buried objects such as landmines.
The complete system consists of a navigation stack to fly user-defined autonomous survey missions, a time-synchronized sensor suite and high-accuracy online and offline localization algorithms.
% What did we show
Our experiments evaluated the individual contributions of the system components.
We demonstrated the robustness of the terrain tracking controller, the \SI{0.8}{\micro\second}~timing accuracy and \SI{0.1}{\micro\second}~precision of the sensor synchronization, \reviewchanges{the advantage of dedicated radar trajectory generation,} and the effectiveness of the localization algorithm due to self-calibration, batch optimization and a dual \ac{GNSS} receiver setup. 
Most notably, we demonstrated that this combination of features enables \reviewchanges{side-looking} \ac{GPSAR} imaging on a rotary wing \ac{MAV}, an imaging technique which heavily relies on an accurate estimation of the sensor trajectory for reconstruction.

% Impact
\reviewchanges{The system presented in this paper is a valid alternative to \ac{GPSAR} on large aerial vehicles~\cite{moussally2004ground} as well as \ac{GPSAR} on \acp{MAV} with downward-facing radar~\cite{garcia2020airborne}.
The system particularly stands out due to its full, self-contained autonomy.}
Camera data collected on an initial flight can be used to generate georeferenced \acp{DSM}, which allow users to safely plan radar flight missions close to the surface, even in regions beyond visual line of sight.
The system enables autonomous radar flights with localization accuracies that allow the generation of crisp radar images in which buried objects are clearly distinguishable.
This is an important step towards an aerial device that is capable of detecting landmines in the context of humanitarian demining. In addition, we have open-sourced each of the software components of our system to encourage reuse and further development by the community, the details are provided in \reftab{tab:opensource}.
\reviewchanges{In fact our research partners have been able to utilize the platform to continue improving radar imaging on \acp{MAV} without further assistance.
For example, the system enabled \ac{InSAR} to create \acp{DSM}~\cite{burr2021uav}, as well as the detection of avalanche victims~\cite{grathwohl2021experimental} or tripwires~\cite{schartel2020tripwire}.} 

\begin{table}[t]
\centering
\resizebox{\textwidth}{!}{%
\begin{tabular}{lll} 
\toprule 
Package  & Link & Description (Section) \\ 
\midrule mav\_findmine & \url{github.com/ethz-asl/mav_findmine}   & Main project repository\\
&& Platform user interface (\ref{sec:mission_overview}) \\
&& \Acl{FSM} (\ref{sec:mission_overview}) \\
&& Radar trajectory generation (\ref{sec:trajectory_generation}) \\
&& DJI tracking controller (\ref{sec:controller}) \\
&& Altitude estimator (\ref{sec:altitude_estimator}) \\
&& Orthomosaic generation utilities (\ref{sec:buried_objects}) \\
polygon\_coverage\_planning & \url{github.com/ethz-asl/polygon_coverage_planning}   & Optical survey mission planning (\ref{sec:mission_overview}, \ref{sec:buried_objects})\\
ethz\_piksi\_ros & \url{github.com/ethz-asl/ethz_piksi_ros}   & \ac{ROS} driver for \ac{RTK} \ac{GNSS}\\
&& \ac{PPS} sync Kernel module (\ref{sec:timing}) \\
&& \ac{RTK} base and survey station setup (\ref{sec:mission_overview}) \\
mav\_gtsam\_estimator & \url{github.com/ethz-asl/mav_gtsam_estimator}   & \ac{GTSAM} \ac{GNSS} and \ac{IMU} fusion (\ref{sec:localization})\\
versavis & \url{github.com/rikba/versavis/tree/feature/gnss_sync}   & \textit{VersaVIS} firmware with \ac{PPS} sync (\ref{sec:timing})\\
\bottomrule
\end{tabular}}
\caption{List of open-source contributions.}
\label{tab:opensource}
\end{table}

%Future work
\reviewchanges{
\section{Outlook}
% Radar future work
We believe that airborne radar imaging on \acp{MAV} is a unique opportunity to reveal hidden objects in inaccessible places and should be pursued further.
Despite the recent progress in the field and the promising results from the presented system, several open research questions remain.
\subsection*{Navigation}
We think that the proportional gain control structure utilized in this work is sufficient to operate in a multitude of terrains.
A simple way to immediately improve flight accuracy and precision is to replace the DJI pose feedback with our online, sliding window state estimation.
A more challenging, but promising task would be the development of radar survey trajectories in undulating and cluttered terrain.
The path planning would have to avoid collisions while simultaneously sampling the terrain uniformly.
In a first step, the survey surface could be approximated as piece-wise linear such that the presented geometric motion primitives are still applicable.
In the long run the detailed terrain and admissible flight space has to be considered to generate informative surveys~\cite{bruggenwirth2019robotic,zhang2018p}.
Additionally, it is important to continue developing radar subsampling strategies to filter informative measurements given the flight path and terrain~\cite{garcia20203d}.
\subsection*{Localization}
Both sensor timing and localization quality showed sufficient performance margins for \ac{GPSAR} imaging.
However, we reached the point where the downstream imaging task did not discriminate between different localization qualities.
In order to continue high-precision state estimation research a precise outdoor ground truth such as a 6DOF total station would be required~\cite{leicaat960xr}.
This would allow investigating the \ac{GNSS} error noise model, and \ac{IMU} vibration damping~\cite{braun2016high}.
Furthermore, quantifiable integration of \ac{LiDAR}, vision or radar to enable \ac{GPSAR} imaging in \ac{GNSS} denied environments would be possible~\cite{zhang2014loam,bahnemann2019eth,kramer2020radar}.
Additionally, an automatic calibration routine to determine the frequency-dependent radar antenna phase centers with respect to the \ac{IMU} could be developed~\cite{doer2020radar}.
At the moment this is still determined from \ac{CAD} and probably one of the largest error sources in the localization pipeline.
\subsection*{Airborne \ac{GPSAR} in Humanitarian Demining}
This and the related work show great potential of using \acp{MAV} for demining.
From a development viewpoint it would be important to collaborate with research groups from affected regions in order to transition the technique into a realistic use case.
Knowledge of the target environment, such as terrain, typical flight obstacles, vegetation, and landmine types, would help focus the development of the individual components.
This includes the navigation and localization, but also the development of the radar imaging.
The radar back projection algorithm still assumes a simplified soil model and the surface model is limited by the quality of the \ac{DSM}.
Techniques that deal with heterogeneous environments, clutter, and vegetation still need to be developed.
More advanced radar imaging capabilities will automatically demand even better navigation and localization capabilities.
%One possible region are the Golan Heights in \reffig{fig:golan}, as they are well accessible and relatively sparsely vegetated which would reduce the domain adaptation required.  
%\begin{figure}
%    \centering
%    \includegraphics{fig/golan.jpg}
%    \caption{\reviewchanges{A minefield in the Golan Heights 2017. To guide the future development of \ac{GPSAR} on \acp{MAV} for humanitarian demining, research teams from affected regions should be involved.}}
%    \label{fig:golan}
%\end{figure}
}

Nevertheless, the results obtained in this work are already applicable across multiple domains, especially in cases where automatic data acquisition and/or accurate outdoor localization is required. Examples include direct photogrammetry, \ac{LiDAR} mapping and time-series data acquisition.

\subsubsection*{Acknowledgments}
This system was developed within project FindMine, a project started by the Urs Endress Foundation to develop new technologies for humanitarian demining~\cite{findmine2021}. The authors would thank Urs Endress for his personal engagement. Furthermore, the authors would like to thank Sportamt Z\"urich and FIFA Z\"urich for providing the test grounds.
Last but not least the authors thank Michael Riner-Kuhn and Matthias M\"uller for their \acl{PCB} designs, Ralf Burr for the radar hardware development and Markus Schartel for his initial work on the radar back projection algorithm.

% \listoftodos

\bibliographystyle{apalike-place}
\bibliography{references}

\begin{thebibliography}{}

\bibitem[Albrektsen and Johansen, 2018]{albrektsen2018user}
Albrektsen, S.~M. and Johansen, T.~A. (2018).
\newblock {User-Configurable Timing and Navigation for {UAV}s}.
\newblock {\em Sensors}, 18(8):2468.

\bibitem[B{\"a}hnemann et~al., 2021]{bahnemann2021revisiting}
B{\"a}hnemann, R., Lawrance, N., Chung, J.~J., Pantic, M., Siegwart, R., and
  Nieto, J. (2021).
\newblock Revisiting {B}oustrophedon {C}overage {P}ath {P}lanning as a
  {G}eneralized {T}raveling {S}alesman {P}roblem.
\newblock In {\em Field and Service Robotics}, pages 277--290. Springer.

\bibitem[B{\"a}hnemann et~al., 2019]{bahnemann2019eth}
B{\"a}hnemann, R., Pantic, M., Popovi{\'c}, M., Schindler, D., Tranzatto, M.,
  Kamel, M., Grimm, M., Widauer, J., Siegwart, R., and Nieto, J. (2019).
\newblock {T}he {ETH-MAV} {T}eam in the {MBZ} {I}nternational {R}obotics
  {C}hallenge.
\newblock {\em Journal of Field Robotics}, 36(1):78--103.

\bibitem[Bekar et~al., 2021]{bekar2021low}
Bekar, A., Antoniou, M., and Baker, C.~J. (2021).
\newblock {L}ow-{C}ost, {H}igh-{R}esolution, {D}rone-{B}orne {SAR} {I}maging.
\newblock {\em IEEE Transactions on Geoscience and Remote Sensing}.

\bibitem[Black, 1964]{black1964passive}
Black, H.~D. (1964).
\newblock {A Passive System for Determining the Attitude of a Satellite}.
\newblock {\em AIAA journal}, 2(7):1350--1351.

\bibitem[Braun, 2016]{braun2016high}
Braun, B. (2016).
\newblock {\em {High Performance Kalman Filter Tuning for Integrated Navigation
  Systems}}.
\newblock PhD thesis, Technische Universit{\"a}t M{\"u}nchen.

\bibitem[Br{\"u}ggenwirth and Rial, 2019]{bruggenwirth2019robotic}
Br{\"u}ggenwirth, S. and Rial, F. (2019).
\newblock {Robotic Control for Cognitive UWB Radar}.
\newblock In {\em Robotic Intelligence}, pages 145--151. World Scientific.

\bibitem[Burr et~al., 2021]{burr2021uav}
Burr, R., Schartel, M., Grathwohl, A., Mayer, W., Walter, T., and Waldschmidt,
  C. (2021).
\newblock {UAV-Borne FMCW InSAR for Focusing Buried Objects}.
\newblock {\em IEEE Geoscience and Remote Sensing Letters}.

\bibitem[Burr et~al., 2018]{burr2018design}
Burr, R., Schartel, M., Schmidt, P., Mayer, W., Walter, T., and Waldschmidt, C.
  (2018).
\newblock {Design and Implementation of a FMCW GPR for UAV-based Mine
  Detection}.
\newblock In {\em 2018 IEEE MTT-S International Conference on Microwaves for
  Intelligent Mobility (ICMIM)}, pages 1--4. IEEE.

\bibitem[Burri et~al., 2015]{burri2015real}
Burri, M., Oleynikova, H., Achtelik, M.~W., and Siegwart, R. (2015).
\newblock {Real-Time Visual-Inertial Mapping, Re-localization and Planning
  Onboard MAVs in Unknown Environments}.
\newblock In {\em 2015 IEEE/RSJ International Conference on Intelligent Robots
  and Systems (IROS)}, pages 1872--1878. IEEE.

\bibitem[Cao et~al., 2021]{cao2021gvins}
Cao, S., Lu, X., and Shen, S. (2021).
\newblock {GVINS: Tightly Coupled GNSS-Visual-Inertial for Smooth and
  Consistent State Estimation}.
\newblock {\em arXiv preprint arXiv:2103.07899}.

\bibitem[Cioffi and Scaramuzza, 2020]{cioffi2020tightly}
Cioffi, G. and Scaramuzza, D. (2020).
\newblock {Tightly-coupled Fusion of Global Positional Measurements in
  Optimization-based Visual-Inertial Odometry}.
\newblock In {\em 2020 IEEE/RSJ International Conference on Intelligent Robots
  and Systems (IROS)}, pages 5089--5095. IEEE.

\bibitem[Colorado et~al., 2017]{colorado2017integrated}
Colorado, J., Perez, M., Mondragon, I., Mendez, D., Parra, C., Devia, C.,
  Martinez-Moritz, J., and Neira, L. (2017).
\newblock An integrated aerial system for landmine detection: {SDR}-based
  {G}round {P}enetrating {R}adar onboard an autonomous drone.
\newblock {\em Advanced Robotics}, 31(15):791--808.

\bibitem[Cramer, 2001]{cramer2001}
Cramer, E.~A. (2001).
\newblock {The Mineseeker Airship: supporting the {UN}}.
\newblock In {\em The Journal of Mine Action}, volume~5, pages 108--113.

\bibitem[Cristoforato et~al., 2000]{cristoforato2000feasibility}
Cristoforato, S., Bishop, P., and Thornhill, C. (2000).
\newblock {The feasibility of operating an Ultra Wideband Synthetic Aperture
  Radar (UWB SAR) from an airship for the detection of mined areas in a
  humanitarian role}.
\newblock {\em 3rd International Airship Convention and Exhibition}.

\bibitem[Curnow and Lichvar, 1997]{chrony1997}
Curnow, R. and Lichvar, M. (1997).
\newblock {Chrony - a versatile implementation of the Network Time Protocol
  (NTP)}.
\newblock \url{https://chrony.tuxfamily.org/}.
\newblock Last checked on June~07, 2021.

\bibitem[Daniels, 2009]{daniels2009ground}
Daniels, D.~J. (2009).
\newblock {{G}round Penetrating Radar for Buried Landmine and {IED} Detection}.
\newblock In {\em Unexploded Ordnance Detection and Mitigation}, pages 89--111.
  Springer.

\bibitem[Dellaert, 2020]{dellaert2020}
Dellaert, F. (2020).
\newblock {Derivatives and Differentials}.
\newblock \url{https://github.com/borglab/gtsam/blob/4.0.3/doc/math.pdf}.
\newblock Last checked on November~04, 2021.

\bibitem[Dellaert et~al., 2017]{dellaert2017factor}
Dellaert, F., Kaess, M., et~al. (2017).
\newblock {Factor Graphs for Robot Perception}.
\newblock {\em Foundations and Trends{\textregistered} in Robotics},
  6(1-2):1--139.

\bibitem[Ding et~al., 2008]{ding2008time}
Ding, W., Wang, J., Li, Y., Mumford, P., and Rizos, C. (2008).
\newblock {Time Synchronization Error and Calibration in Integrated GPS/INS
  Systems}.
\newblock {\em ETRI journal}, 30(1):59--67.

\bibitem[Doer and Trommer, 2020]{doer2020radar}
Doer, C. and Trommer, G.~F. (2020).
\newblock {R}adar {I}nertial {O}dometry {W}ith {O}nline {C}alibration.
\newblock In {\em 2020 European Navigation Conference (ENC)}, pages 1--10.
  IEEE.

\bibitem[{Dronecode}, 2021]{qgroundcontrol}
{Dronecode} (2021).
\newblock {P}lan - {QGroundControl} {U}ser {G}uide.
\newblock
  \url{https://docs.qgroundcontrol.com/master/en/PlanView/PlanView.html\#flight-speed}.
\newblock Last checked on Nov~28, 2021.

\bibitem[Eidson et~al., 2002]{eidson2002ieee}
Eidson, J.~C., Fischer, M., and White, J. (2002).
\newblock {IEEE-1588™ Standard for a Precision Clock Synchronization Protocol
  for Networked Measurement and Control Systems}.
\newblock In {\em Proceedings of the 34th Annual Precise Time and Time Interval
  Systems and Applications Meeting}, pages 243--254.

\bibitem[Endress, 2021]{findmine2021}
Endress, U. (2021).
\newblock {FindMine and Urs Endress Foundation Against Mines and Bombs}.
\newblock \url{https://www.ue-stiftung.org/findmine}.
\newblock Last checked on June~16, 2021.

\bibitem[Esposito et~al., 2020]{esposito2020uav}
Esposito, G., Noviello, C., Soldovieri, F., Catapano, I., Fasano, G.,
  Gagliarde, G., Luisi, G., and Saccoccio, F. (2020).
\newblock The {UAV} radar imaging prototype developed in the frame of the
  {VESTA} project.
\newblock In {\em 2020 IEEE Radar Conference (RadarConf20)}, pages 1--5. IEEE.

\bibitem[Faizullin et~al., 2021a]{faizullin2021synchronized}
Faizullin, M., Kornilova, A., Akhmetyanov, A., Pakulev, K., Sadkov, A., and
  Ferrer, G. (2021a).
\newblock {S}ynchronized {S}martphone {V}ideo {R}ecording {S}ystem of {D}epth
  and {RGB} {I}mage {F}rames with {S}ub-millisecond {P}recision.
\newblock {\em arXiv preprint arXiv:2111.03552}.

\bibitem[Faizullin et~al., 2021b]{faizullin2021open}
Faizullin, M., Kornilova, A., and Ferrer, G. (2021b).
\newblock {O}pen-{S}ource {LiDAR} {T}ime {S}ynchronization {S}ystem by
  {M}imicking {GPS}-clock.
\newblock {\em arXiv preprint arXiv:2107.02625}.

\bibitem[Fang et~al., 2017]{fang2017robust}
Fang, Z., Yang, S., Jain, S., Dubey, G., Roth, S., Maeta, S., Nuske, S., Zhang,
  Y., and Scherer, S. (2017).
\newblock {Robust Autonomous Flight in Constrained and Visually Degraded
  Shipboard Environments}.
\newblock {\em Journal of Field Robotics}, 34(1):25--52.

\bibitem[Fasano et~al., 2017]{fasano2017proof}
Fasano, G., Renga, A., Vetrella, A.~R., Ludeno, G., Catapano, I., and
  Soldovieri, F. (2017).
\newblock {Proof of concept of micro-UAV-based radar imaging}.
\newblock In {\em 2017 International Conference on Unmanned Aircraft Systems
  (ICUAS)}, pages 1316--1323. IEEE.

\bibitem[Forster et~al., 2015]{forster2015imu}
Forster, C., Carlone, L., Dellaert, F., and Scaramuzza, D. (2015).
\newblock {IMU Preintegration on Manifold for Efficient Visual-Inertial
  Maximum-a-Posteriori Estimation}.
\newblock In {\em 2015 IEEE International Conference on Robotics and Automation
  (ICRA)}. Georgia Institute of Technology.

\bibitem[Frey et~al., 2021]{frey2021measurement}
Frey, O., Werner, C.~L., Manconi, A., and Coscione, R. (2021).
\newblock {Measurement of surface displacements with a UAV-borne/car-borne
  L-band DInSAR system: system performance and use cases}.
\newblock In {\em 2021 IEEE International Geoscience and Remote Sensing
  Symposium IGARSS}, pages 628--631. IEEE.

\bibitem[Furgale et~al., 2012]{furgale2012continuous}
Furgale, P., Barfoot, T.~D., and Sibley, G. (2012).
\newblock {Continuous-Time Batch Estimation using Temporal Basis Functions}.
\newblock In {\em 2012 IEEE International Conference on Robotics and
  Automation}, pages 2088--2095. IEEE.

\bibitem[Furgale et~al., 2013]{furgale2013unified}
Furgale, P., Rehder, J., and Siegwart, R. (2013).
\newblock {Unified Temporal and Spatial Calibration for Multi-sensor Systems}.
\newblock In {\em International Conference on Intelligent Robots and Systems
  (IROS)}. IEEE.

\bibitem[Garc{\'\i}a-Fern{\'a}ndez et~al., 2019]{garcia2019autonomous}
Garc{\'\i}a-Fern{\'a}ndez, M., L{\'o}pez, Y.~{\'A}., and Andr{\'e}s, F. L.-H.
  (2019).
\newblock {Autonomous Airborne 3D SAR Imaging System for Subsurface Sensing:
  UWB-GPR on Board a UAV for Landmine and IED Detection}.
\newblock {\em Remote Sensing}, 11(20):2357.

\bibitem[Garc{\'\i}a-Fern{\'a}ndez et~al., 2020a]{garcia20203d}
Garc{\'\i}a-Fern{\'a}ndez, M., L{\'o}pez, Y.~{\'A}., and Andr{\'e}s, F. L.-H.
  (2020a).
\newblock {3D-SAR Processing of UAV-mounted GPR Measurements: Dealing with
  Non-Uniform Sampling}.
\newblock In {\em 2020 14th European Conference on Antennas and Propagation
  (EuCAP)}, pages 1--5. IEEE.

\bibitem[Garc{\'\i}a-Fern{\'a}ndez et~al., 2020b]{garcia2020airborne}
Garc{\'\i}a-Fern{\'a}ndez, M., L{\'o}pez, Y.~{\'A}., and Andr{\'e}s, F. L.-H.
  (2020b).
\newblock {Airborne Multi-Channel Ground Penetrating Radar for Improvised
  Explosive Devices and Landmine Detection}.
\newblock {\em IEEE Access}, 8:165927--165943.

\bibitem[Garc{\'\i}a-Fern{\'a}ndez et~al., 2018]{fernandez2018synthetic}
Garc{\'\i}a-Fern{\'a}ndez, M., L{\'o}pez, Y.~{\'A}., Arboleya, A.~A.,
  Vald{\'e}s, B.~G., Vaqueiro, Y.~R., Andr{\'e}s, F. L.-H., and Garc{\'\i}a,
  A.~P. (2018).
\newblock {Synthetic Aperture Radar Imaging System for Landmine Detection Using
  a Ground Penetrating Radar on Board a Unmanned Aerial Vehicle}.
\newblock {\em IEEE Access}, 6:45100--45112.

\bibitem[Grathwohl et~al., 2021]{grathwohl2021experimental}
Grathwohl, A., Hinz, P., Burr, R., Steiner, M., and Waldschmidt, C. (2021).
\newblock {Experimental Study on the Detection of Avalanche Victims using an
  Airborne Ground Penetrating Synthetic Aperture Radar}.
\newblock In {\em 2021 IEEE Radar Conference (RadarConf21)}, pages 1--6. IEEE.

\bibitem[Heinzel et~al., 2019]{heinzel2019comparison}
Heinzel, A., Schartel, M., Burr, R., B{\"a}hnemann, R., Schreiber, E., Peichl,
  M., and Waldschmidt, C. (2019).
\newblock A comparison of ground-based and airborne {SAR} systems for the
  detection of landmines, {UXO}, and {IED}s.
\newblock In {\em Radar Sensor Technology XXIII}, volume 11003, page 1100304.
  International Society for Optics and Photonics.

\bibitem[Hexagon, 2021]{leicaat960xr}
Hexagon (2021).
\newblock {Leica Absolute Tracker AT960}.
\newblock
  \url{https://www.creativeinfocom.com/pdfs/leica-absolute-tracker-at930-brochure.pdf}.
\newblock Last checked on October~22, 2021.

\bibitem[Huck et~al., 2011]{huck2011precise}
Huck, T., Westenberger, A., Fritzsche, M., Schwarz, T., and Dietmayer, K.
  (2011).
\newblock {Precise Timestamping and Temporal Synchronization in Multi-Sensor
  Fusion}.
\newblock In {\em 2011 IEEE intelligent vehicles symposium (IV)}, pages
  242--247. IEEE.

\bibitem[Indelman et~al., 2013]{indelman2013information}
Indelman, V., Williams, S., Kaess, M., and Dellaert, F. (2013).
\newblock Information fusion in navigation systems via factor graph based
  incremental smoothing.
\newblock {\em Robotics and Autonomous Systems}, 61(8):721--738.

\bibitem[{International Campaign to Ban Landmines}, 2020]{landminemonitor2020}
{International Campaign to Ban Landmines} (2020).
\newblock {Landmine Monitor 2020}.
\newblock Technical report, Cluster Munition Coalition (ICBL-CMC).

\bibitem[Kaess et~al., 2012]{kaess2012isam2}
Kaess, M., Johannsson, H., Roberts, R., Ila, V., Leonard, J.~J., and Dellaert,
  F. (2012).
\newblock {iSAM2: Incremental smoothing and mapping using the Bayes tree}.
\newblock {\em The International Journal of Robotics Research}, 31(2):216--235.

\bibitem[Kais et~al., 2006]{kais2006multi}
Kais, M., Millescamps, D., B{\'e}taille, D., Lusetti, B., and Chapelon, A.
  (2006).
\newblock {A Multi-Sensor Acquisition Architecture and Real-Time Reference for
  Sensor and Fusion Methods Benchmarking}.
\newblock In {\em 2006 IEEE Intelligent Vehicles Symposium}, pages 418--423.
  IEEE.

\bibitem[Kaplan and Hegarty, 2005]{kaplan2005understanding}
Kaplan, E. and Hegarty, C. (2005).
\newblock {\em {UNDERSTANDING GPS: PRINCIPLES AND APPLICATIONS}}.
\newblock Artech house.

\bibitem[Kelly et~al., 2021]{kelly2021question}
Kelly, J., Grebe, C., and Giamou, M. (2021).
\newblock A {Q}uestion of {T}ime: {R}evisiting the {U}se of {R}ecursive
  {F}iltering for {T}emporal {C}alibration of {M}ultisensor {S}ystems.
\newblock {\em arXiv preprint arXiv:2106.00391}.

\bibitem[Kelly and Sukhatme, 2014]{kelly2014general}
Kelly, J. and Sukhatme, G.~S. (2014).
\newblock {A General Framework for Temporal Calibration of Multiple
  Proprioceptive and Exteroceptive Sensors}.
\newblock In {\em Experimental Robotics}, pages 195--209. Springer.

\bibitem[Kramer et~al., 2020]{kramer2020radar}
Kramer, A., Stahoviak, C., Santamaria-Navarro, A., Agha-mohammadi, A.-a., and
  Heckman, C. (2020).
\newblock {Radar-Inertial Ego-Velocity Estimation for Visually Degraded
  Environments}.
\newblock In {\em 2020 IEEE International Conference on Robotics and Automation
  (ICRA)}, pages 5739--5746. IEEE.

\bibitem[Li and Mourikis, 2014]{li2014online}
Li, M. and Mourikis, A.~I. (2014).
\newblock Online temporal calibration for camera--{IMU} systems: {T}heory and
  algorithms.
\newblock {\em The International Journal of Robotics Research}, 33(7):947--964.

\bibitem[Li et~al., 2015]{li2015precise}
Li, X., Zhang, X., Ren, X., Fritsche, M., Wickert, J., and Schuh, H. (2015).
\newblock {Precise positioning with current multi-constellation Global
  Navigation Satellite Systems: GPS, GLONASS, Galileo and BeiDou}.
\newblock {\em Scientific reports}, 5:8328.

\bibitem[Looney, 2018]{adis2018}
Looney, M. (2018).
\newblock {ADIS16448 Data Sampling}.
\newblock
  \url{https://ez.analog.com/mems/w/documents/4122/adis16448-data-sampling}.
\newblock Last checked on June~07, 2021.

\bibitem[Lynen et~al., 2013]{lynen2013robust}
Lynen, S., Achtelik, M.~W., Weiss, S., Chli, M., and Siegwart, R. (2013).
\newblock {A Robust and Modular Multi-Sensor Fusion Approach Applied to MAV
  Navigation}.
\newblock In {\em 2013 IEEE/RSJ international conference on intelligent robots
  and systems}, pages 3923--3929. IEEE.

\bibitem[Mian et~al., 2015]{mian2015direct}
Mian, O., Lutes, J., Lipa, G., Hutton, J., Gavelle, E., and Borghini, S.
  (2015).
\newblock {DIRECT GEOREFERENCING ON SMALL UNMANNED AERIAL PLATFORMS FOR
  IMPROVED RELIABILITY AND ACCURACY OF MAPPING WITHOUT THE NEED FOR GROUND
  CONTROL POINTS}.
\newblock {\em The international archives of photogrammetry, remote sensing and
  spatial information sciences}, 40(1):397.

\bibitem[{Mirage~Systems~Inc.}, 1984]{miragesystems}
{Mirage~Systems~Inc.} (1984).
\newblock {Mirage Systems - a subsurface imaging, technology development and
  systems company}.
\newblock \url{https://miragesystems.com/}.
\newblock Last checked on Nov~24, 2021.

\bibitem[Moreira et~al., 2013]{moreira2013tutorial}
Moreira, A., Prats-Iraola, P., Younis, M., Krieger, G., Hajnsek, I., and
  Papathanassiou, K.~P. (2013).
\newblock {A Tutorial on Synthetic Aperture Radar}.
\newblock {\em IEEE Geoscience and remote sensing magazine}, 1(1):6--43.

\bibitem[Moussally et~al., 2004]{moussally2004ground}
Moussally, G.~J., Fries, R.~W., and Bortins, R. (2004).
\newblock Ground-penetrating synthetic-aperture radar for wide-area airborne
  minefield detection.
\newblock In {\em Detection and Remediation Technologies for Mines and Minelike
  Targets IX}, volume 5415, pages 1042--1052. International Society for Optics
  and Photonics.

\bibitem[Mueller et~al., 2015]{mueller2015computationally}
Mueller, M.~W., Hehn, M., and D'Andrea, R. (2015).
\newblock {A Computationally Efficient Motion Primitive for Quadrocopter
  Trajectory Generation}.
\newblock {\em IEEE Transactions on Robotics}, 31(6):1294--1310.

\bibitem[Nikolic et~al., 2014]{nikolic2014synchronized}
Nikolic, J., Rehder, J., Burri, M., Gohl, P., Leutenegger, S., Furgale, P.~T.,
  and Siegwart, R. (2014).
\newblock {A Synchronized Visual-Inertial Sensor System with FPGA
  Pre-Processing for Accurate Real-Time SLAM}.
\newblock In {\em 2014 IEEE international conference on robotics and automation
  (ICRA)}, pages 431--437. IEEE.

\bibitem[Osadcuks et~al., 2020]{osadcuks2020clock}
Osadcuks, V., Pudzs, M., Zujevs, A., Pecka, A., and Ardavs, A. (2020).
\newblock Clock-based time synchronization for an event-based camera dataset
  acquisition platform.
\newblock In {\em 2020 IEEE International Conference on Robotics and Automation
  (ICRA)}, pages 4695--4701. IEEE.

\bibitem[Peichl et~al., 2014]{peichl2014tirami}
Peichl, M., Schreiber, E., Heinzel, A., and Kempf, T. (2014).
\newblock {TIRAMI}-{SAR}-a synthetic aperture radar approach for efficient
  detection of landmines and {UXO}.
\newblock In {\em EUSAR 2014; 10th European Conference on Synthetic Aperture
  Radar}, pages 1--4. VDE.

\bibitem[Pix4D, 2021]{pix4d2021}
Pix4D (2021).
\newblock {PIX4Dmapper photogrammetry software}.
\newblock
  \url{https://www.pix4d.com/product/pix4dmapper-photogrammetry-software}.
\newblock Last checked on June~07, 2021.

\bibitem[Pradalier, 2017]{pradalier2017task}
Pradalier, C. (2017).
\newblock {A task scheduler for ROS}.
\newblock Technical Report UMI 2958, GeorgiaTech-CNRS.
\newblock Last checked on May~25, 2021.

\bibitem[Qin and Shen, 2018]{qin2018online}
Qin, T. and Shen, S. (2018).
\newblock {Online Temporal Calibration for Monocular Visual-Inertial Systems}.
\newblock In {\em 2018 IEEE/RSJ International Conference on Intelligent Robots
  and Systems (IROS)}, pages 3662--3669. IEEE.

\bibitem[Richter et~al., 2016]{richter2016polynomial}
Richter, C., Bry, A., and Roy, N. (2016).
\newblock {Polynomial Trajectory Planning for Aggressive Quadrotor Flight in
  Dense Indoor Environments}.
\newblock In {\em Robotics research}, pages 649--666. Springer.

\bibitem[Roumeliotis and Burdick, 2002]{roumeliotis2002stochastic}
Roumeliotis, S.~I. and Burdick, J.~W. (2002).
\newblock {S}tochastic {C}loning: {A} generalized framework for processing
  relative state measurements.
\newblock In {\em Proceedings 2002 IEEE International Conference on Robotics
  and Automation (Cat. No. 02CH37292)}, volume~2, pages 1788--1795. IEEE.

\bibitem[Schartel, 2021]{schartel2021signalverarbeitungskonzepte}
Schartel, M. (2021).
\newblock {\em Signalverarbeitungskonzepte zur Minendetektion mittels
  drohnengest{\"u}tztem Ground Penetrating Synthetic Aperture Radar}.
\newblock PhD thesis, Universit{\"a}t Ulm.

\bibitem[Schartel et~al., 2020a]{schartel2020experimental}
Schartel, M., Burr, R., B{\"a}hnemann, R., Mayer, W., and Waldschmidt, C.
  (2020a).
\newblock An {E}xperimental {S}tudy on {A}irborne {L}andmine {D}etection
  {U}sing a {C}ircular {S}ynthetic {A}perture {R}adar.
\newblock {\em arXiv preprint arXiv:2005.02600}.

\bibitem[Schartel et~al., 2018]{schartel2018uav}
Schartel, M., Burr, R., Mayer, W., Docci, N., and Waldschmidt, C. (2018).
\newblock {UAV-Based Ground Penetrating Synthetic Aperture Radar}.
\newblock In {\em IEEE MTT-S International Conference on Microwaves for
  Intelligent Mobility}, pages 1--4. IEEE.

\bibitem[Schartel et~al., 2020b]{schartel2020tripwire}
Schartel, M., Grathwohl, A., Schmid, C., Burr, R., and Waldschmidt, C. (2020b).
\newblock {TRIPWIRE DETECTION IN SAR IMAGES USING A MODIFIED RADON TRANSFORM}.
\newblock In {\em IGARSS 2020-2020 IEEE International Geoscience and Remote
  Sensing Symposium}, pages 746--749. IEEE.

\bibitem[{Schiebel~Corporation}, 1997]{camcopter}
{Schiebel~Corporation} (1997).
\newblock {CAMCOPTER S-100 Technical Specifications}.
\newblock \url{https://schiebel.net/products/camcopter-s-100-system-2/}.
\newblock Last checked on Nov~24, 2021.

\bibitem[Schneider et~al., 2016]{schneider2016fast}
Schneider, J., Eling, C., Klingbeil, L., Kuhlmann, H., F{\"o}rstner, W., and
  Stachniss, C. (2016).
\newblock {Fast and Effective Online Pose Estimation and Mapping for UAVs}.
\newblock In {\em 2016 IEEE International Conference on Robotics and Automation
  (ICRA)}, pages 4784--4791. IEEE.

\bibitem[Schreiber et~al., 2019]{schreiber2019advanced}
Schreiber, E., Heinzel, A., Peichl, M., Engel, M., and Wiesbeck, W. (2019).
\newblock {Advanced Buried Object Detection by Multichannel, UAV/Drone Carried
  Synthetic Aperture Radar}.
\newblock In {\em 2019 13th European Conference on Antennas and Propagation
  (EuCAP)}, pages 1--5. IEEE.

\bibitem[{\v{S}}ipo{\v{s}} and Gleich, 2020]{vsipovs2020lightweight}
{\v{S}}ipo{\v{s}}, D. and Gleich, D. (2020).
\newblock A lightweight and low-power {UAV}-borne ground penetrating radar
  design for landmine detection.
\newblock {\em Sensors}, 20(8):2234.

\bibitem[Skog and Handel, 2011]{skog2011time}
Skog, I. and Handel, P. (2011).
\newblock {Time Synchronization Errors in Loosely Coupled GPS-Aided Inertial
  Navigation Systems}.
\newblock {\em IEEE transactions on intelligent transportation systems},
  12(4):1014--1023.

\bibitem[Sommer et~al., 2017]{sommer2017low}
Sommer, H., Khanna, R., Gilitschenski, I., Taylor, Z., Siegwart, R., and Nieto,
  J. (2017).
\newblock A low-cost system for high-rate, high-accuracy temporal calibration
  for lidars and cameras.
\newblock In {\em 2017 IEEE/RSJ International Conference on Intelligent Robots
  and Systems (IROS)}, pages 2219--2226. IEEE.

\bibitem[Strode and Groves, 2016]{strode2016gnss}
Strode, P.~R. and Groves, P.~D. (2016).
\newblock {GNSS} multipath detection using three-frequency signal-to-noise
  measurements.
\newblock {\em GPS solutions}, 20(3):399--412.

\bibitem[Svedin et~al., 2021]{svedin2021small}
Svedin, J., Bernland, A., and Gustafsson, A. (2021).
\newblock {S}mall {UAV}-based high resolution {SAR} using low-cost radar,
  {GNSS}/{RTK} and {IMU} sensors.
\newblock In {\em 2020 17th European Radar Conference (EuRAD)}, pages 186--189.
  IEEE.

\bibitem[{Swift Navigation}, 2019]{piksi2019}
{Swift Navigation} (2019).
\newblock {Piksi Multi - GNSS Module Hardware Specification}.
\newblock
  \url{https://www.swiftnav.com/resource-files/Piksi\%20Multi/v2.2/Specification/Piksi\%20Multi\%20HW\%20Specification\%20v2.2\%20\%5B000-534-02-02\%5D.pdf}.
\newblock Last checked on June~07, 2021.

\bibitem[Tschopp et~al., 2020]{tschopp2020versavis}
Tschopp, F., Riner, M., Fehr, M., Bernreiter, L., Furrer, F., Novkovic, T.,
  Pfrunder, A., Cadena, C., Siegwart, R., and Nieto, J. (2020).
\newblock {VersaVIS—An Open Versatile Multi-Camera Visual-Inertial Sensor
  Suite}.
\newblock {\em Sensors}, 20(5):1439.

\bibitem[{United Nations Mine Action Service (UNMAS)}, 2019]{gichd2021imas0820}
{United Nations Mine Action Service (UNMAS)} (2019).
\newblock {IMAS 08.10 - Non-technical survey}.
\newblock
  \url{https://www.mineactionstandards.org/fileadmin/MAS/documents/standards/IMAS-08.10-Ed1-Am4-Non-Technical_Survey.pdf}.
\newblock Last checked on September~27, 2021.

\bibitem[{Vicon Motion Systems}, 2014]{vicon2014}
{Vicon Motion Systems} (2014).
\newblock {Bonita Optical Motion Tracking}.
\newblock \url{https://est-kl.com/images/PDF/Vicon/Bonita_Optical.pdf}.
\newblock Last checked on June~07, 2021.

\bibitem[Wang et~al., 2008]{wang2008image}
Wang, J., Li, Y., Zhou, Z., Jin, T., Yang, Y., and Wang, Y. (2008).
\newblock {Image Formation Techniques for Vehicle-Mounted Forward-Looking
  Ground Penetrating SAR}.
\newblock In {\em 2008 International Conference on Information and Automation},
  pages 667--671. IEEE.

\bibitem[Yanghuan et~al., 2011]{yanghuan2011lever}
Yanghuan, L., Fulai, L., Qian, S., and Zhimin, Z. (2011).
\newblock {Lever Arm Rotation Compensation Method for UAV Mounted SAR}.
\newblock In {\em 2011 3rd International Asia-Pacific Conference on Synthetic
  Aperture Radar (APSAR)}, pages 1--3. IEEE.

\bibitem[Zaugg and Long, 2015]{zaugg2015generalized}
Zaugg, E.~C. and Long, D.~G. (2015).
\newblock {Generalized Frequency Scaling and Backprojection for LFM-CW SAR
  Processing}.
\newblock {\em IEEE Transactions on Geoscience and Remote Sensing},
  53(7):3600--3614.

\bibitem[Zhang et~al., 2018]{zhang2018p}
Zhang, J., Chadha, R.~G., Velivela, V., and Singh, S. (2018).
\newblock {P-CAP: Pre-computed Alternative Paths to Enable Aggressive Aerial
  Maneuvers in Cluttered Environments}.
\newblock In {\em 2018 IEEE/RSJ International Conference on Intelligent Robots
  and Systems (IROS)}, pages 8456--8463. IEEE.

\bibitem[Zhang and Singh, 2014]{zhang2014loam}
Zhang, J. and Singh, S. (2014).
\newblock {LOAM}: {L}idar {O}dometry and {M}apping in {R}eal-time.
\newblock In {\em Robotics: Science \& Systems (RSS)}. IEEE.

\end{thebibliography}

\end{document}